\documentclass[10pt,twocolumn,letterpaper]{article}

\usepackage[pagenumbers]{iccv} 

\newcommand{\hide}[1]{}

\newcommand{\dquote}[1]{``#1''}

\usepackage{color}
\usepackage{xcolor,colortbl}
\usepackage{mathtools}
\usepackage{multirow}
\usepackage{bm} 
\usepackage{pifont}
\usepackage{subcaption} 
\usepackage{enumitem} 
\usepackage{xspace}
\usepackage{amsmath}
\usepackage{amsfonts}
\usepackage{booktabs}
\definecolor{purple}{rgb}{0.65,0,0.65}
\definecolor{dark_green}{rgb}{0, 0.5, 0}
\definecolor{blueish}{rgb}{0.0, 0.3, .6}
\definecolor{LightCyan}{rgb}{0.88,0.95,1}
\definecolor{LightGrey}{rgb}{0.56,0.56,0.56}
\definecolor{tabhighlight}{HTML}{f2f0f0} 
\newcolumntype{h}{>{\columncolor{tabhighlight}}c}
\definecolor{positivetabledelta}{rgb}{0.172, 0.750, 0.098} 
\definecolor{negativetabledelta}{rgb}{0.851, 0.2, 0.078}

\makeatletter
\newcommand*\wt[2][0.15ex]{%
        \begingroup
        \mathchoice{\wt@helper{#1}{#2}{\displaystyle}{\textfont}}
                   {\wt@helper{#1}{#2}{\textstyle}{\textfont}}
                   {\wt@helper{#1}{#2}{\scriptstyle}{\scriptfont}}
                   {\wt@helper{#1}{#2}{\scriptscriptstyle}{\scriptscriptfont}}%
        \endgroup
        #2%
}
\newcommand*\wt@helper[4]{%
        \def\currentfont{\the#41}%
        \def\currentskewchar{\char\the\skewchar\currentfont}%
        \setbox\tw@\hbox{\currentfont#2\currentskewchar}%
        \dimen@ii\wd\tw@
        \setbox\tw@\hbox{\currentfont#2{}\currentskewchar}%
        \advance\dimen@ii-\wd\tw@
        \rlap{\raisebox{-#1}{$\m@th#3\kern\dimen@ii\widetilde{\phantom{#2}}$}}%
}
\makeatother

\usepackage[export]{adjustbox} 

\newlength{\gridimagewidth}
\usepackage{array}
\newcolumntype{C}[1]{>{\centering\let\newline\\\arraybackslash\hspace{0pt}}m{#1}}

\newcommand{\tit}[1]{\smallbreak\noindent\textbf{#1}}

\definecolor{ForestGreen}{RGB}{34, 139, 34}
\newcommand{\hgreen}[1]{\textcolor{ForestGreen}{\textbf{#1}}} 
\newcommand{\hred}[1]{\textcolor{red}{\textbf{#1}}} 

\makeatletter
\DeclareRobustCommand\onedot{\futurelet\@let@token\@onedot}
\def\@onedot{\ifx\@let@token.\else.\null\fi\xspace}

\def\eg{\emph{e.g}\onedot} 
\def\ie{\emph{i.e}\onedot}

\makeatother

\newcommand{\method}{QualiCLIP\xspace}
\newcommand{\methodplus}{QualiCLIP$^{+}$\xspace}
\newcommand{\methodextended}{QualiCLIP (Quality-aware CLIP)\xspace}
\newcommand{\methodextension}{Quality-aware CLIP\xspace}
\newcommand{\Lone}{\mathcal{L}_{cons}\xspace}
\newcommand{\Ltwo}{\mathcal{L}_{pos}\xspace}
\newcommand{\Lthree}{\mathcal{L}_{neg}\xspace}
\newcommand{\lambdaone}{\lambda_{cons}\xspace}
\newcommand{\lambdatwo}{\lambda_{pos}\xspace}
\newcommand{\lambdathree}{\lambda_{neg}\xspace}
\newcommand{\marginone}{m_{cons}\xspace}
\newcommand{\margintwo}{m_{rank}\xspace}
\newcommand{\live}{LIVE\xspace}
\newcommand{\clive}{CLIVE\xspace}
\newcommand{\csiq}{CSIQ\xspace}
\newcommand{\tid}{TID2013\xspace}
\newcommand{\kadid}{KADID-10k\xspace}

\newcommand{\agiqaonek}{AGIQA-1K\xspace}
\newcommand{\agiqathreek}{AGIQA-3K\xspace}
\newcommand{\pipal}{PIPAL\xspace}
\newcommand{\cviu}{CVIU\xspace}
\newcommand{\koniq}{KonIQ-10k\xspace}
\newcommand{\spaq}{SPAQ\xspace}
\newcommand{\flive}{FLIVE\xspace}
\newcommand{\srcc}{SRCC\xspace}
\newcommand{\plcc}{PLCC\xspace}

\newcommand{\cmark}{\ding{51}}%
\newcommand{\xmark}{\ding{55}}%

\definecolor{iccvblue}{rgb}{0.21,0.49,0.74}
\usepackage[pagebackref,breaklinks,colorlinks,allcolors=iccvblue]{hyperref}

\def\paperID{} 
\def\confName{ICCV}
\def\confYear{2025}

\title{Quality-Aware Image-Text Alignment\\for Opinion-Unaware Image Quality Assessment}

\author{Lorenzo Agnolucci \quad Leonardo Galteri \quad Marco Bertini \vspace{5pt} \\
University of Florence \\
{\tt\small [name.surname]@unifi.it}
}

\begin{document}
\maketitle

\begin{abstract}
No-Reference Image Quality Assessment (NR-IQA) focuses on designing methods to measure image quality in alignment with human perception when a high-quality reference image is unavailable. Most state-of-the-art NR-IQA approaches are opinion-aware, i.e. they require human annotations for training. This dependency limits their scalability and broad applicability. To overcome this limitation, we propose \methodextended, a CLIP-based self-supervised opinion-unaware approach that does not require human opinions. In particular, we introduce a quality-aware image-text alignment strategy to make CLIP generate quality-aware image representations. Starting from pristine images, we synthetically degrade them with increasing levels of intensity. Then, we train CLIP to rank these degraded images based on their similarity to quality-related antonym text prompts. At the same time, we force CLIP to generate consistent representations for images with similar content and the same level of degradation. Our experiments show that the proposed method improves over existing opinion-unaware approaches across multiple datasets with diverse distortion types. Moreover, despite not requiring human annotations, \method achieves excellent performance against supervised opinion-aware methods in cross-dataset experiments, thus demonstrating remarkable generalization capabilities. 
The code and the model are publicly available at \small{\href{https://github.com/miccunifi/QualiCLIP}{\url{https://github.com/miccunifi/QualiCLIP}}}. 
\end{abstract}

\section{Introduction} \label{sec:intro}

\begin{figure}[t]
    \centering
    \includegraphics[width=0.495\columnwidth]{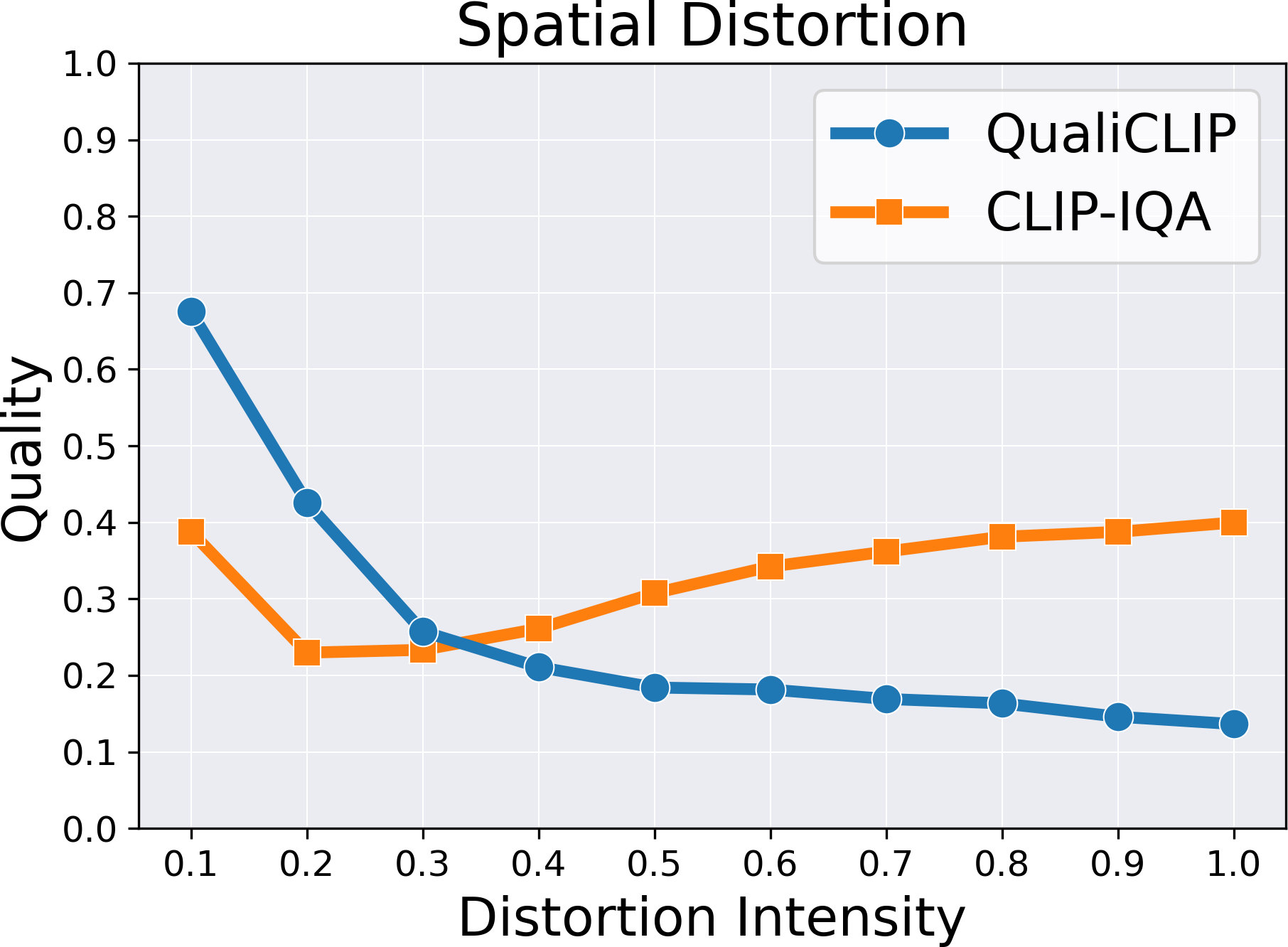}
    \includegraphics[width=0.495\columnwidth]{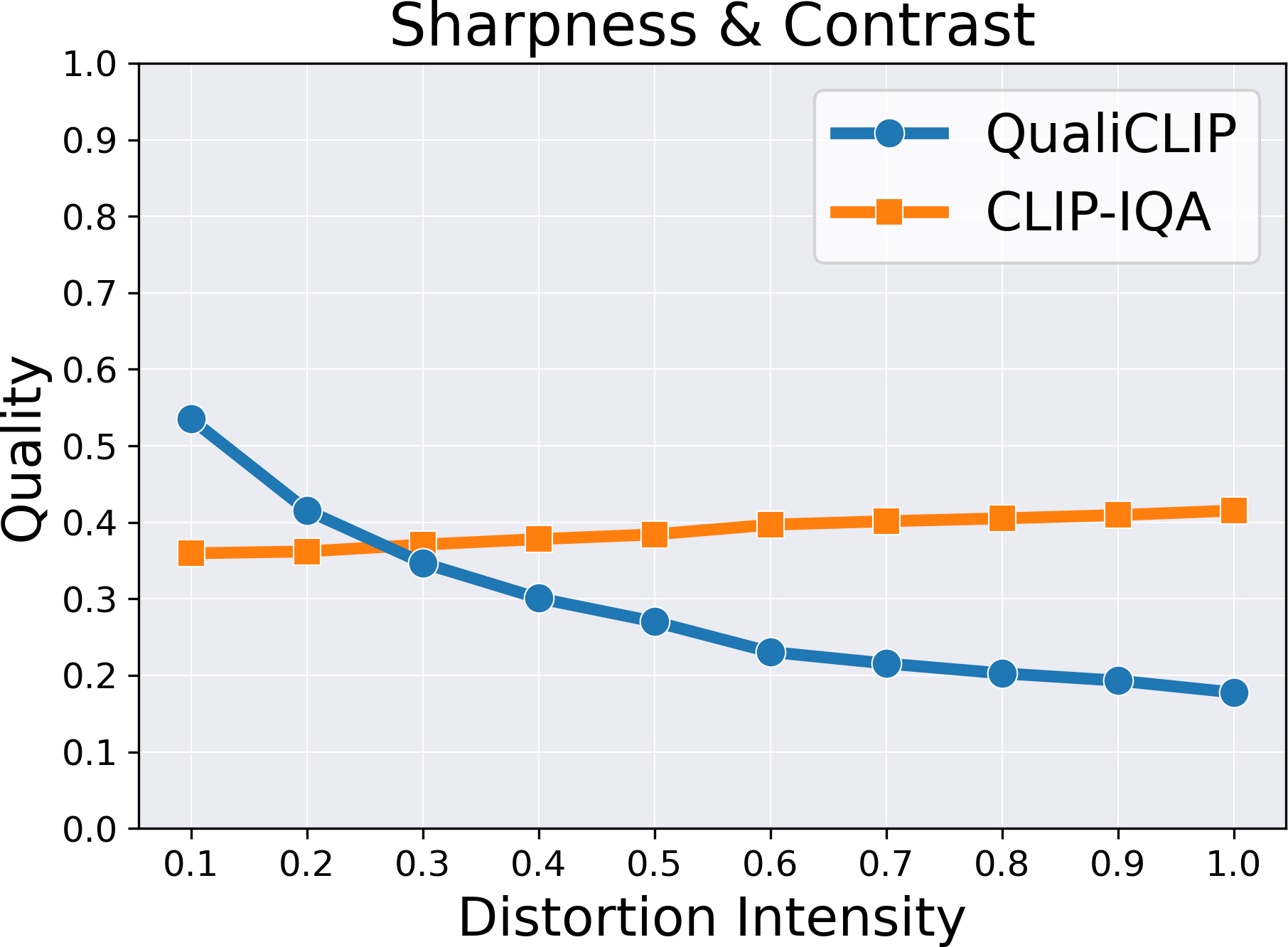}
    \caption{Comparison between the image quality scores predicted by CLIP-IQA \cite{wang2023exploring} and the proposed \method for increasing distortion intensities of different types of synthetic degradation. We average the results of 1000 randomly sampled images from the \koniq \cite{hosu2020koniq10k} dataset. Our method corresponds to a stronger inverse correlation between the predicted quality scores and the severity of the degradation. The distortion intensities are scaled between 0 and 1 for clearer visualization.}
    \label{fig:teaser_degradation_correlation}
\end{figure}

Image Quality Assessment (IQA) aims to automatically evaluate the quality of images in accordance with human judgments represented by a Mean Opinion Score (MOS). Specifically, No-Reference IQA (NR-IQA) focuses on developing methods that do not require a high-quality reference image and that are consequently more easily applicable in real-world scenarios. NR-IQA plays a critical role in diverse industries and research domains. For example, given the large number of photos that are captured and shared daily on social media platforms, it is imperative to design approaches that can measure image quality objectively to store and process these images effectively. 

Most NR-IQA methods are opinion-aware, \ie they require human-labeled MOS as supervision during the training process \cite{su2020blindly, saha2023re, zhang2023blind, zhao2023quality}. Some approaches, such as HyperIQA \cite{su2020blindly} or LIQE \cite{zhang2023blind}, directly train the model parameters on IQA datasets. Other methods, namely QPT \cite{zhao2023quality} or Re-IQA \cite{saha2023re}, pre-train an encoder on unlabeled data via self-supervised learning and then either fine-tune the encoder weights or train a linear regressor using MOS. However, annotating IQA datasets is expensive and resource-intensive, as several human ratings are needed for each image for its MOS to be reliable.
For example, the FLIVE dataset \cite{ying2020patches}, which contains 40K real-world images, required about 4M ratings, up to 50 for a single image.
The need for human annotations significantly hinders the scalability of opinion-aware approaches. In addition, these methods show limited generalization capabilities and thus applicability to scenarios where labeled data is unavailable, as their performance considerably deteriorates on unseen datasets. To remove the requirement for expensive MOS, several opinion-unaware approaches have been proposed \cite{ma2017dipiq, babu2023no, gu2019no}. 
For instance, CL-MI \cite{babu2023no} introduces a two-stage self-supervised approach that employs two different training strategies for synthetically and authentically degraded images. 
Nevertheless, existing opinion-unaware methods achieve significantly lower performance than opinion-aware approaches in cross-dataset experiments, thus exhibiting limited applicability.

In this context, we propose to leverage recent advancements in Vision-Language Models (VLMs) by presenting a self-supervised opinion-unaware approach based on CLIP \cite{radford2021learning}. Recently, CLIP-based methods achieved promising performance in NR-IQA \cite{wang2023exploring, zhang2023blind, srinath2024learning}. For example, CLIP-IQA \cite{wang2023exploring} proposes to compute the quality score by measuring the similarity between an image and two quality-related antonym prompts without any task-specific training. However, off-the-shelf CLIP models struggle to generate quality-aware image representations \cite{zhang2023blind, liang2023iterative}, as they focus more on high-level semantics than low-level image characteristics, such as noise and blur. To highlight this issue, we randomly sample 1000 images from the \koniq dataset \cite{hosu2020koniq10k} and synthetically degrade them with several distortions using increasing levels of intensity. Then, we compute the quality score of each image through CLIP-IQA and average the results. We expect the more degraded versions of the images to correspond to lower quality scores. However, \cref{fig:teaser_degradation_correlation} shows that CLIP-IQA exhibits a low correlation between the predicted quality and the degree of distortion. This finding indicates that CLIP is \textit{not} intrinsically quality-aware.

To address this issue, we propose a quality-aware image-text alignment strategy that relies on self-supervised learning to remove the need for human annotations. 
We start by synthetically degrading pairs of pristine images using increasing levels of intensity. Then, we measure the similarity between each image and \textit{antonym prompts} related to image quality, such as ``\textit{Good photo}'' and ``\textit{Bad photo}''. We refer to these prompts as \textit{positive} and \textit{negative} prompts, respectively. Finally, we employ a training strategy based on a margin ranking loss \cite{liang2023iterative, liu2017rankiqa} that allows us to achieve two objectives. First, we want CLIP to generate consistent representations for images having similar content and comparable quality, \ie exhibiting the same amount of distortion. Second, the similarity between the positive (negative) prompt and the increasingly degraded versions of the images must correlate inversely (directly) with the intensity of the distortion. Our approach, named \method (\methodextension), is both self-supervised and opinion-unaware, as we do not rely on any form of supervision -- especially MOS -- at any step of the training process. Thanks to our training strategy, the image-text alignment in CLIP's embedding space prioritizes low-level image characteristics over high-level semantics. Consequently, \method generates image representations whose similarity to the antonym prompts correlates with the inherent quality of the images, as illustrated in \cref{fig:teaser_degradation_correlation}.

The experiments demonstrate that the proposed approach improves over existing opinion-unaware methods across multiple datasets encompassing various degradations. Furthermore, \method is the only opinion-unaware approach that consistently obtains remarkable results even when compared against supervised opinion-aware techniques in the cross-dataset setting. The strong and robust performance of our model across different datasets highlights its commendable generalization capabilities and suitability for real-world applications. 

We summarize our contributions as follows:
\begin{itemize}
    \item We propose \method, a CLIP-based self-supervised opinion-unaware approach for NR-IQA that does not require any type of supervision, especially MOS;
    \item We introduce a quality-aware image-text alignment strategy based on ranking increasingly degraded pairs of images according to their similarity to quality-related antonym prompts. After training, \method generates quality-aware image representations;
    \item Our method improves on existing opinion-unaware approaches across multiple datasets and achieves excellent results even when compared to supervised opinion-aware techniques in cross-dataset experiments. 
\end{itemize}

\section{Related Work}

\tit{No-Reference Image Quality Assessment}
Due to its wide range of applications in real-world scenarios, in recent years research on NR-IQA has gained significant momentum \cite{mittal2012making, su2020blindly, madhusudana2022image, agnolucci2024arniqa, babu2023no}. Traditional methods \cite{mittal2012making, mittal2012no, zhang2015feature} rely on extracting hand-crafted image features to derive quality scores. Such approaches achieve promising performance on synthetic datasets but struggle on images with authentic distortions. More recently, several methods relying on supervised learning have been introduced \cite{su2020blindly, golestaneh2022no, madhusudana2022image, agnolucci2024arniqa, xu2024boosting, saha2023re, shi2024transformer}. Some approaches directly employ MOS during model training \cite{su2020blindly, golestaneh2022no, xu2024boosting}. For example, HyperIQA \cite{su2020blindly} proposes a self-adaptive hypernetwork that separates content understanding from quality prediction. Another research direction involves pre-training an encoder on unlabeled images via self-supervised learning. Then, the image representations are mapped to quality scores by fine-tuning the encoder weights \cite{zhao2023quality} or training a linear regressor \cite{madhusudana2022image, saha2023re, agnolucci2024arniqa} on human annotations. For instance, QPT \cite{zhao2023quality} and Re-IQA \cite{saha2023re} use a contrastive loss to train the encoder to separate between images degraded with different types and degrees of distortion. Despite their impressive performance, the scalability and applicability of supervised methods are limited by their need for costly human annotations. This requirement is removed by opinion-unaware approaches \cite{mittal2012making, zhang2015feature, ma2017dipiq, babu2023no, shukla2024opinion, ni2024opinion}. Some of them, such as NIQE \cite{mittal2012making}, are based on natural scene statistics \cite{mittal2012making, zhang2015feature}, while others employ self-supervised learning \cite{ma2017dipiq, babu2023no, shukla2024opinion, gu2019no}. For example, CL-MI \cite{babu2023no} pre-trains an encoder on synthetic data and then fine-tunes it on authentic images via a mutual information-based loss. Nevertheless, existing opinion-unaware approaches fall behind supervised methods in cross-dataset experiments. In contrast, despite not requiring MOS, our method achieves remarkable performance on unseen datasets even when considering opinion-aware techniques.

\tit{Vision-Language Models for NR-IQA}
VLMs, such as CLIP \cite{radford2021learning}, have achieved impressive performance in several low-level vision tasks, including image and video restoration \cite{liang2023iterative, luo2023controlling, agnolucci2024reference} and quality assessment \cite{wang2023exploring, zhang2023blind, srinath2024learning, wu2023towardsrobust, wu2023q, wu2024qalign, wu2023exploring}. CLIP-IQA \cite{wang2023exploring} studies the capabilities of CLIP in assessing the quality and abstract perception of images without task-specific training. In addition, the authors train a model named CLIP-IQA$^{+}$ based on learning two antonym prompts using MOS. LIQE \cite{zhang2023blind} fine-tunes CLIP in a supervised way with a multi-task learning approach exploiting scene and distortion-type information. Recently, several methods based on Multimodal Large Language Models (MLLMs) have been proposed \cite{you2023depicting, wu2023q, wu2024qalign}. While these approaches achieve impressive results, they require significant computational resources due to the high demands of MLLMs. Among VLM-based methods for NR-IQA, the most similar to our work is GRepQ \cite{srinath2024learning}, which trains a low-level and a high-level CLIP-based encoder via self-supervised learning. CLIP is fine-tuned by separating higher and lower-quality groups of images within the same batch with a contrastive loss depending on their predicted quality, obtained by measuring their similarity to antonym prompts. GRepQ predicts the final quality score by combining the features of the two encoders and feeding them as input to a linear regressor, which is trained on IQA datasets using MOS. In contrast, we present a CLIP-only self-supervised approach that removes the need for a low-level encoder. We propose to synthetically degrade pairs of images with increasing levels of intensity and make our model learn to rank them through a ranking loss according to their degree of distortion. The ranking is based directly on the similarity between the text features and each of the antonym prompts, instead of relying on the predicted quality as GRepQ. Also, differently from GRepQ, we do not require any form of supervision at any step of our approach.

\tit{Learning to rank}
Learning to rank images has proven to be an effective technique for image quality and aesthetics assessment \cite{liu2017rankiqa, golestaneh2022no, ke2023vila, thong2022content, ma2017dipiq, roy2023test}. For instance, VILA \cite{ke2023vila} tackles image aesthetics assessment by training a learnable residual projection on top of CLIP to rank the quality of a single pair of images according to their MOS. Another example is RankIQA \cite{liu2017rankiqa}, which involves synthetically degrading images with varying degrees using dataset-specific distortions. Then, for each IQA dataset, the authors first pre-train a Siamese network by ranking the images based on their level of degradation and then fine-tune it with the MOS.
In our work, we employ a given set of distortions to degrade pairs of crops with increasing levels of intensity. Then, we leverage the information provided by their implicit quality ranking to train a model to rank them according to their similarity to antonym prompts. In this way, our method does not require fine-tuning on ground-truth labels.


\section{Proposed Approach}
We propose a quality-aware image-text alignment strategy to make CLIP generate quality-aware image representations. First, we synthetically degrade pairs of crops with increasing levels of intensity. Then, we fine-tune CLIP's image encoder by ranking the similarity between two antonym prompts and the progressively distorted image pairs based on their degree of degradation, while guaranteeing consistent representations for each pair of crops. We keep CLIP's text encoder fixed. We use a ResNet50 \cite{he2016deep} as the backbone for CLIP. We do not employ any supervision -- particularly MOS -- at any step of the training process. Due to space limitations, we provide the implementation details in the supplementary material.

\subsection{CLIP Preliminaries}
CLIP (Contrastive Language-Image Pre-training) \cite{radford2021learning} is a vision-language model trained on a large-scale dataset to semantically align images and corresponding text captions in a shared embedding space. The authors employ a contrastive loss that maximizes the similarity between paired image-text samples while minimizing the similarity with all the other samples within a batch. CLIP comprises an image encoder $\psi_{I}$ and a text encoder $\psi_{T}$. Given an image $I$, the image encoder extracts its feature representation $x = \psi_{I}(I) \in \mathbb{R}^{d}$, where $d$ is the dimension of CLIP's embedding space. For a given text caption $T$, each tokenized word is mapped to the token embedding space $\mathcal{W}$ through a word embedding layer $E_w$. Then, the text encoder $\psi_{T}$ is used to generate the textual feature representation $y=\psi_{T}(E_w(T)) \in \mathbb{R}^{d}$ from the token embeddings. Thanks to its training strategy, CLIP generates similar representations within the common embedding space for images and text expressing the same concepts.

\subsection{Synthetic Degradation with Increasing Levels of Intensity} \label{sec:image_degradation_model} 
\begin{figure}[t]
    \centering
    \includegraphics[width=\columnwidth]{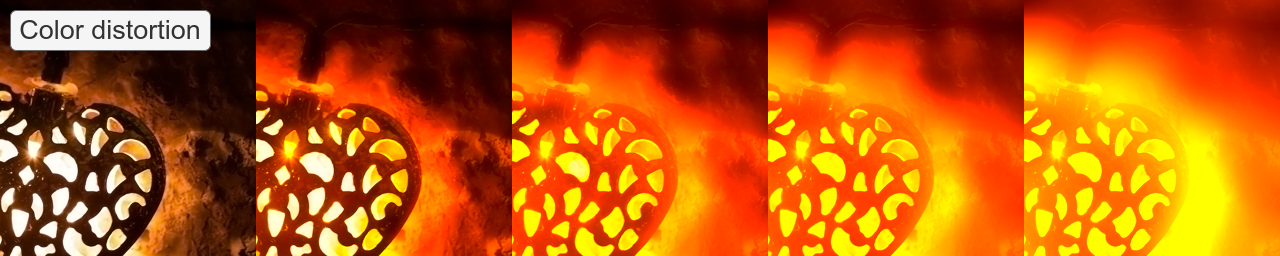}
    \includegraphics[width=\columnwidth]{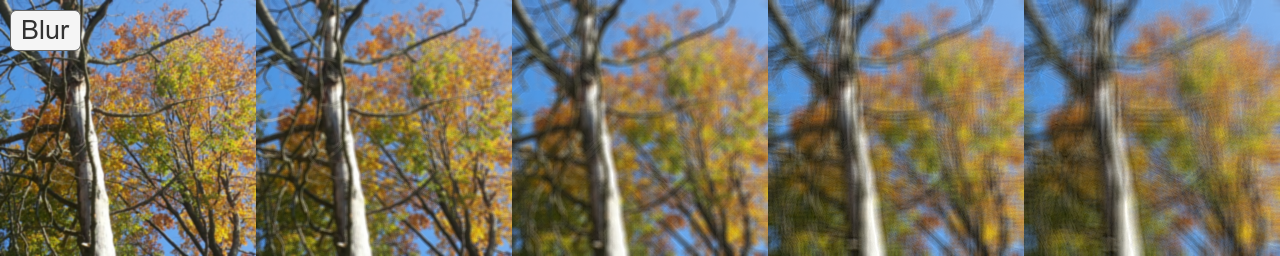}
    \includegraphics[width=\columnwidth]{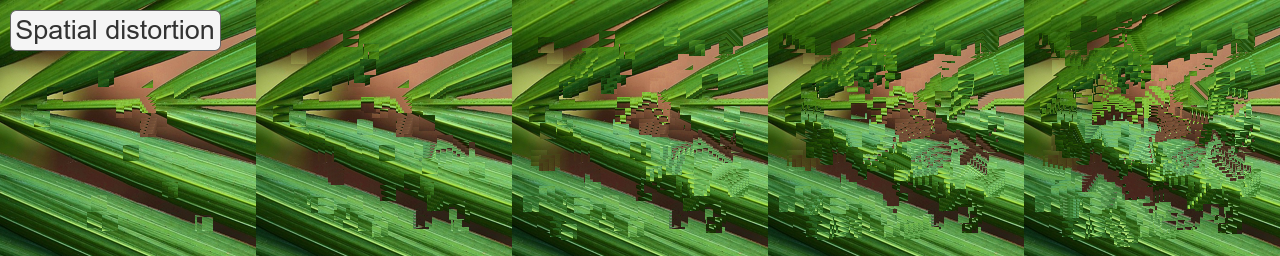}
    \caption{Examples of synthetic degradations for five increasing levels of intensity.}
    \label{fig:example_degraded_crops}
\end{figure}

Although authentic distortions cannot be perfectly replicated synthetically, prior studies have shown that synthetic distortions remain effective for training self-supervised NR-IQA models that generalize well to real-world images \cite{madhusudana2022image, zhao2023quality, saha2023re, agnolucci2024arniqa}. Following these works, we synthetically degrade unlabeled pristine images using progressively increasing levels of intensity. In this way, we can train our model in a self-supervised way to rank the different versions of each image according to the severity of their degradation. Following \cite{agnolucci2024arniqa}, we consider 24 distinct degradation types spanning the 7 distortion groups defined by the \kadid dataset \cite{kadid10k}: 1) Brightness change; 2) Blur; 3) Spatial distortions; 4) Noise; 5) Color distortions; 6) Compression; 7) Sharpness \& contrast. Each distortion has $L \!=\! 5$ levels of intensity. See the supplementary material for more details on the specific degradation types. \cref{fig:example_degraded_crops} shows some examples of degraded images for varying degrees of intensity.

We start from the 140K pristine images of the KADIS-700k dataset \cite{kadid10k}. For each image, we extract a pair of random overlapping crops. Then, we randomly sample $D\!=\!1$ distortion groups and a degradation within each group. We apply the $D$ distortions to both crops using $L\!=\!5$ distinct levels of intensity, resulting in $L$ pairs of equally degraded crops, one for each level. Contrary to RankIQA \cite{liu2017rankiqa}, we obtain two images for each degree of distortion, as depicted in the leftmost part of \cref{fig:method}. Given two such pairs of crops, we can infer which has a higher quality based on the corresponding level of degradation. We leverage this information to train our model with a ranking loss.

\subsection{Quality-Aware Image-Text Alignment} \label{sec:training_strategy}

As \cref{fig:teaser_degradation_correlation} shows, CLIP struggles to generate accurate quality-aware image representations that reflect the severity of the degradation. To address this issue, we propose a quality-aware image-text alignment strategy to fine-tune CLIP's image encoder. The idea of our approach is that given two degraded versions of the same image, a positive prompt referring to high image quality -- such as ``\textit{Good photo}'' -- should be more similar to the less degraded version. The opposite consideration applies considering a negative prompt related to low image quality, such as ``\textit{Bad photo}''. At the same time, two images with overlapping content and equal degree of degradation should have comparable similarities to such a pair of quality-related antonym prompts. Note that, given two unlabeled images with completely different content, we can not make any assumptions about their relative quality \cite{chiu2020assessing}, or, in other words, their similarity to the prompts. Our training strategy leverages multiple pairs of increasingly degraded images to achieve two objectives: \textit{O1)}: we want CLIP to generate consistent representations for images with similar content and comparable quality, \ie showing the same amount of distortion; \textit{O2)}: the similarity between the positive (negative) prompt and the distinct versions of the images must correlate inversely (directly) with the corresponding level of degradation.

\begin{figure*}[t]
    \centering
    \includegraphics[width=\linewidth]{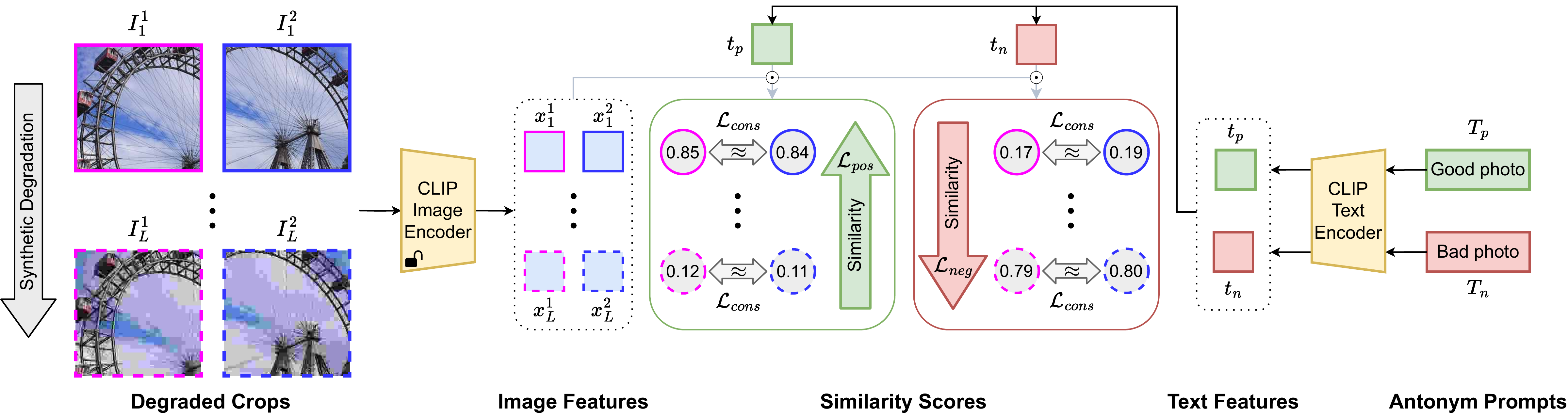}
    \vspace{-15pt}
    \caption{Overview of the proposed quality-aware image-text alignment strategy. Starting from a pair of two random overlapping crops from a pristine image, we synthetically degrade them with $L$ increasing levels of intensity, resulting in $L$ pairs. Then, given two quality-related antonym prompts $T_p$ and $T_n$, we fine-tune CLIP's image encoder with three margin ranking losses ($\Lone$, $\Ltwo$, $\Lthree$) by considering the similarity between the prompts and the degraded crops. Specifically, we use $\Lone$ to force CLIP to generate consistent representations for the crops belonging to each pair, since they exhibit similar content and the same degree of distortion. At the same time, we make the similarity between the prompt $T_p$ (or $T_n$) and the increasingly degraded versions of the crops correlate inversely (or directly) with the intensity of the distortion through $\Ltwo$ (or $\Lthree$).
    }
    \vspace{-5pt}
    \label{fig:method}
\end{figure*}

Let $I_i^1$ and $I_i^2$ be the $i$-th pair of increasingly degraded crops obtained as detailed in \cref{sec:image_degradation_model}, where $i \in \{ 1, \ldots, L\}$ and $L\!=\!5$ is the number of considered distortion levels. For $i, j \in \{ 1, \ldots, L\}$ with $j > i$, the $j$-th pair of crops is more degraded than the $i$-th one. Given each pair of crops, we extract the corresponding features through CLIP's image encoder $\psi_{I}$, resulting in $x_i^1 = \psi_{I}(I_i^1)$ and $x_i^2 = \psi_{I}(I_i^2)$. Similarly to \cite{wang2023exploring}, we remove the positional embedding to relax CLIP's requirement of fixed-size inputs. Let $T_p$ and $T_n$ be a pair of antonym prompts related to image quality, such as ``\textit{Good photo}'' and ``\textit{Bad photo}''. We refer to $T_p$ and $T_n$ as \textit{positive} and \textit{negative} prompts, respectively. In practice, we use multiple pairs of antonym prompts, similar to \cite{wu2023towardsrobust, wu2023exploring}. We use CLIP's text encoder $\psi_{T}$ to extract the text features associated with the prompts, obtaining $t_p = \psi_{T}(T_p)$ and $t_n = \psi_{T}(T_n)$. We normalize both the image and text features to have a unit $L_2$-norm.

To achieve objective \textit{O1}, we propose to employ a consistency loss term to guarantee that the similarity between the features of the prompts and those of each of the two images composing each degraded pair is comparable.
We assume that two overlapping crops extracted from the same image have a comparable quality, analogously to \cite{saha2023re, zhao2023quality}.
We rely on a margin ranking loss \cite{liang2023iterative, liu2017rankiqa, golestaneh2022no} defined as: 
\begin{align}
    \Lone &= \sum_{i=1}^{L} \left [ \max(0, \left |  c(x_i^1, t_p) - c(x_i^2, t_p) \right | - \marginone) \right. \nonumber \\
    &+ \max(0, \left |  c(x_i^1, t_n) - c(x_i^2, t_n) \right | - \marginone) \left. \right ],
    \label{eq:loss_one}
\end{align}
where $c(\cdot)$ stands for the cosine similarity and the margin $\marginone$ is a hyperparameter. Intuitively, $\marginone$ must be small enough to force the similarities between the prompts and each of the two crops to be comparable. With CLIP, the cosine similarity between each image-prompt pair is in $[0, 1]$.

Given the $i$-th level of synthetic degradation, with $i \in \{ 1, \ldots, L\}$, we assume that the quality of the two distorted crops of the $i$-th pair is higher than that of the two images composing the $(i+1)$-th one, analogously to \cite{liu2017rankiqa, roy2023test}. Thus, we enforce that the similarity between the features of the \textit{positive} prompt and those of two crops is \textit{higher} than when considering more degraded versions of the two crops. Specifically, we define a margin ranking loss as:
\begin{equation}
    \resizebox{\columnwidth}{!}{$ \begin{aligned}
    \Ltwo &= \sum_{i=1}^{L-1} \sum_{j=i+1}^{L} \sum_{k=1}^{2} \left [ \max(0, c(x_j^k, t_p) - c(x_i^1, t_p) + \margintwo) \right. \\
    &+ \max(0, c(x_j^k, t_p) - c(x_i^2, t_p) + \margintwo) \left. \right ],
    \end{aligned}$}
    \label{eq:loss_two}
\end{equation}
where the margin $\margintwo$ is a hyperparameter. The opposite of the consideration made above applies when we take into account the negative prompt. Therefore, we add a loss term to impose that the similarity between the features of the \textit{negative} prompt and those of two crops is \textit{lower} than when considering more degraded versions of the two crops:
\begin{equation}
    \resizebox{\columnwidth}{!}{$ \begin{aligned}
    \Lthree &= \sum_{i=1}^{L-1} \sum_{j=i+1}^{L} \sum_{k=1}^{2} \left [ \max(0, c(x_i^1, t_n) - c(x_j^k, t_n) + \margintwo) \right. \\
    &+ \max(0, c(x_i^2, t_n) - c(x_j^k, t_n) + \margintwo) \left. \right ].
    \end{aligned}$}
    \label{eq:loss_three}
\end{equation}
Intuitively, $\margintwo$ must be large enough to make the similarities between the prompts and the increasingly degraded versions of the two crops noticeably different. Using the combination of $\Ltwo$ and $\Lthree$ achieves objective \textit{O2}.

The final training loss is given by:
\begin{equation}
    \mathcal{L} = \lambdaone\Lone + \lambdatwo\Ltwo + \lambdathree\Lthree ,
    \label{eq:loss_final}
\end{equation}
where $\lambdaone$, $\lambdatwo$, and $\lambdathree$ represent the loss weights. \cref{fig:method} shows an overview of our training strategy. Given that we do not employ any MOS, our approach is both self-supervised and opinion-unaware. Thanks to the proposed training strategy, CLIP learns to align images and texts based more on low-level characteristics, such as noise and blur, rather than high-level semantics. As a result, the similarity between the antonym prompts and the image representations obtained by \method correlates with the inherent quality of the images, as shown in \cref{fig:teaser_degradation_correlation}.

At inference time, given an image $I$, we extract its features $x$ using CLIP's image encoder. Then, we compute the cosine similarity between $x$ and the features $t_p$ and $t_n$ of the antonym prompts, resulting in $s_p$ and $s_n$. Finally, we obtain the final quality score $q\!\in\![0, 1]$ as:
\begin{equation}
    q = \frac{e^{(s_{p}/\tau)}}{e^{(s_{p}/\tau)} + e^{(s_{n}/\tau)}}, \label{eq:inference}
\end{equation}
where $\tau$ is a temperature hyperparameter. 
Note that, since we keep CLIP's text encoder weights frozen, we need to compute the text features of the antonym prompts only once, and we can use them both for training and inference. Therefore, at inference time, the computational cost of our method is the same as that of an image-encoder-only model with the same backbone.

\begin{table*}[t]
  \centering
  \setlength{\tabcolsep}{4.25pt}
  \resizebox{\linewidth}{!}{ 
  \begin{tabular}{lccccccccccccccccc} 
  \toprule
  \multicolumn{1}{c}{} & \multicolumn{1}{c}{} & \multicolumn{8}{c}{\textit{Authentic Datasets}} & \multicolumn{4}{c}{\textit{Image Restoration Datasets}} & \multicolumn{4}{c}{\textit{AIGC Datasets}} \\ 
  \cmidrule(lr){3-10} \cmidrule(lr){11-14} \cmidrule(lr){15-18}
  \multicolumn{1}{c}{} & \multicolumn{1}{c}{} & \multicolumn{2}{c}{\koniq} & \multicolumn{2}{c}{\clive} & \multicolumn{2}{c}{\flive} & \multicolumn{2}{c}{\spaq} & \multicolumn{2}{c}{\cviu} & \multicolumn{2}{c}{\pipal} & \multicolumn{2}{c}{\agiqaonek} & \multicolumn{2}{c}{\agiqathreek}\\
  \multicolumn{1}{l}{Method} & OU & \srcc & \plcc & \srcc & \plcc & \srcc & \plcc & \srcc & \plcc & \srcc & \plcc & \srcc & \plcc & \srcc & \plcc & \srcc & \plcc \\ 
  \midrule
  CONTRIQUE \cite{madhusudana2022image} & \hred{\xmark} & 0.874 & 0.882 & 0.806 & 0.838 & 0.596 & 0.648 & 0.910 & 0.912 & 0.912 & 0.916 & 0.581 & 0.611 & 0.799 & 0.855 & 0.817 & 0.879 \\
  Re-IQA \cite{saha2023re} & \hred{\xmark} & 0.883 & 0.887 & 0.783 & 0.797 & 0.623 & 0.685 & 0.909 & 0.912 & 0.887 & 0.895 & 0.568 & 0.587 & 0.783 & 0.84 & 0.811 & 0.874 \\
  ARNIQA \cite{agnolucci2024arniqa} & \hred{\xmark} & 0.869 & 0.883 & 0.797 & 0.823 & 0.595 & 0.670 & 0.904 & 0.909 & 0.919 & 0.926 & 0.634 & 0.666 & 0.768 & 0.849 & 0.803 & 0.881 \\
  CLIP-IQA$^{+}$ \cite{wang2023exploring} & \hred{\xmark} & 0.873 & 0.890 & 0.815 & 0.849 & 0.602 & 0.662 & 0.901 & 0.904 & 0.925 & 0.918 & 0.552 & 0.558 & 0.817 & 0.855 & 0.844 & 0.894 \\ 
  GRepQ \cite{srinath2024learning} & \hred{\xmark} & 0.882 & 0.883 & 0.793 & 0.813 & 0.576 & 0.611 & 0.902 & 0.903 & 0.882 & 0.857 & 0.554 & 0.568 & 0.740 & 0.797 & 0.807 & 0.858 \\ \midrule
  NIQE \cite{mittal2012making} & \hgreen{\cmark} & 0.527 & 0.532 & 0.446 & 0.469 & 0.203 & 0.262 & 0.700 & 0.712 & 0.685 & 0.696 & 0.167 & 0.181 & 0.623 & 0.721 & 0.510 & 0.526 \\
  IL-NIQE \cite{zhang2015feature} & \hgreen{\cmark} & 0.511 & 0.532 & 0.427 & 0.481 & 0.234 & 0.286 & 0.713 & 0.724 & 0.595 & 0.668 & 0.231 & 0.220 & 0.645 & 0.757 & 0.528 & 0.544 \\
  CL-MI \cite{babu2023no} & \hgreen{\cmark} & 0.664 & 0.653 & 0.494 & 0.480 & 0.290 & 0.322 & 0.701 & 0.701 & 0.433 & 0.481 & 0.281 & 0.282 & 0.474 & 0.621 & 0.591 & 0.665 \\
  MDFS \cite{ni2024opinion} & \hgreen{\cmark} & 0.739 & 0.746 & 0.427 & 0.467 & 0.360 & 0.407 & \underline{0.805} & \underline{0.812} & 0.702 & 0.742 & 0.255 & 0.262 & \underline{0.734} & \textbf{0.818} & \textbf{0.697} & 0.723 \\
  CONTRIQUE-OU \cite{madhusudana2022image} & \hgreen{\cmark} & 0.651 & 0.637 & 0.348 & 0.364 & 0.321 & 0.341 & 0.677 & 0.685 & 0.610 & 0.715 & 0.225 & 0.218 & 0.644 & 0.717 & 0.465 & 0.489 \\
  Re-IQA-OU \cite{saha2023re} & \hgreen{\cmark} & 0.580 & 0.568 & 0.379 & 0.388 & 0.350 & 0.341 & 0.613 & 0.616 & 0.648 & 0.662 & 0.152 & 0.151 & 0.644 & 0.764 & 0.366 & 0.430 \\
  ARNIQA-OU \cite{agnolucci2024arniqa} & \hgreen{\cmark} & 0.746 & 0.762 & 0.468 & 0.557 & \underline{0.428} & \underline{0.468} & 0.788 & 0.797 & 0.673 & 0.712 & 0.327 & 0.331 & 0.667 & 0.786 & 0.614 & 0.679 \\
  CLIP-IQA \cite{wang2023exploring} & \hgreen{\cmark} & 0.695 & 0.730 & 0.656 & 0.670 & 0.344 & 0.473 & 0.733 & 0.734 & 0.613 & 0.642 & 0.274 & 0.279 & 0.498 & 0.651 & 0.638 & 0.711 \\
  GRepQ-OU \cite{srinath2024learning} & \hgreen{\cmark} & \underline{0.776} & \underline{0.798} & \underline{0.722} & \underline{0.766} & 0.289 & 0.423 & 0.802 & 0.805 & \underline{0.748} & \underline{0.785} & \textbf{0.434} & \underline{0.419} & 0.161 & 0.530 & 0.613 & \underline{0.734} \\ \midrule
  \rowcolor{LightCyan} \textbf{\method} & \hgreen{\cmark} & \textbf{0.817} & \textbf{0.838} & \textbf{0.725} & \textbf{0.802} & \textbf{0.442} & \textbf{0.556} & \textbf{0.841} & \textbf{0.851} & \textbf{0.840} & \textbf{0.852} & \underline{0.410} & \textbf{0.424} & \textbf{0.736} & \underline{0.805} & \underline{0.667} & \textbf{0.735} \\
  \bottomrule
  \end{tabular}}
  \vspace{-5pt}
  \caption{Quantitative results of the \textit{zero-shot} evaluation setting. OU stands for Opinion-Unaware. Best and second-best scores for OU methods are highlighted in bold and underlined, respectively. The suffix ``-OU'' indicates approaches modified to be opinion-unaware (see \cref{sec:evaluation_protocol}). For reference, we report the performance of supervised methods (\ie OU$=\!\hred{\xmark}$) trained on the training split of each testing dataset.}
  \vspace{-5pt}
  \label{tab:zeroshot}
\end{table*}

\tit{Discussion} 
Our quality-aware image-text alignment strategy stems from the inherent limitations of applying common self-supervised NR-IQA training techniques to CLIP. Prior methods, such as QPT \cite{zhao2023quality} and Re-IQA \cite{saha2023re}, train an encoder using a contrastive loss that maximizes the similarity between the representations of crops from the same degraded image, while minimizing the similarity with the representations of crops coming from different images within the same batch. While this strategy proved its effectiveness for image-encoder-only models, applying it to CLIP's image encoder would introduce a significant mismatch with CLIP's training process \cite{radford2021learning}. 
Indeed, CLIP is trained using an inter-modal (\ie image-text) objective, aligning the features -- extracted with the image and text encoders -- of corresponding images and texts within a common embedding space. Consequently, fine-tuning CLIP's image encoder by considering only intra-modal (\ie image-image) similarities without exploiting its alignment with the text encoder contradicts its design \cite{udandarao2023sus, mistretta2025cross}. For this reason, we propose to train our model by employing image-text similarities to leverage CLIP's inherent inter-modal alignment.
Additionally, using a contrastive loss to maximize (or minimize) the similarity of the antonym prompts with multiple different training samples within the same batch would correspond to making assumptions on the relative quality of unlabeled images with completely different content, which is unfeasible \cite{zhao2023quality}. Instead, by relying on a ranking loss that only considers progressively degraded versions of the same image, we can leverage their inherent quality ranking as supervision to train our model in an effective way.

\begin{table*}[t]
  \centering
  \setlength{\tabcolsep}{4.5pt}
  \resizebox{\linewidth}{!}{ 
  \begin{tabular}{lccccccccccccccc} 
  \toprule
  \multicolumn{2}{c}{} & \multicolumn{6}{c}{\textit{Authentic Datasets}} & \multicolumn{4}{c}{\textit{Image Restoration Datasets}} & \multicolumn{4}{c}{\textit{AIGC Datasets}} \\ 
  \cmidrule(lr){3-8} \cmidrule(lr){9-12} \cmidrule(lr){13-16}
  \multicolumn{1}{c}{} & \multicolumn{1}{c}{} & \multicolumn{2}{c}{\koniq} & \multicolumn{2}{c}{\clive} & \multicolumn{2}{c}{\spaq} & \multicolumn{2}{c}{\cviu} & \multicolumn{2}{c}{\pipal} & \multicolumn{2}{c}{\agiqaonek} & \multicolumn{2}{c}{\agiqathreek} \\
  \multicolumn{1}{l}{Method} & \multicolumn{1}{l}{OU} & \srcc & \plcc & \srcc & \plcc & \srcc & \plcc & \srcc & \plcc & \srcc & \plcc & \srcc & \plcc & \srcc & \plcc \\ 
  \midrule
  HyperIQA \cite{su2020blindly} & \hred{\xmark} & 0.738 & 0.735 & 0.699 & 0.722 & 0.649 & 0.650 & 0.361 & 0.400 & 0.289 & 0.95 & 0.403 & 0.582 & 0.490 & 0.570 \\
  DBCNN \cite{zhang2018blind} & \hred{\xmark} & 0.804 & 0.823 & 0.660 & 0.698 & 0.808 & 0.817 & 0.786 & 0.803 & 0.356 & 0.374 & \underline{0.525} & \underline{0.726} & 0.510 & 0.657 \\
  TReS \cite{golestaneh2022no} & \hred{\xmark} & 0.759 & 0.758 & 0.696 & 0.727 & 0.739 & 0.735 & 0.466 & 0.510 & 0.326 & 0.328 & 0.392 & 0.585 & 0.515 & 0.569 \\
  LIQE \cite{zhang2023blind} & \hred{\xmark} & \underline{0.813} & 0.801 & \underline{0.743} & 0.756 & 0.713 & 0.712 & 0.560 & 0.625 & 0.349 & 0.393 & 0.450 & 0.663 & 0.563 & 0.633 \\
  QCN \cite{shin2024blind} & \hred{\xmark} & 0.728 & 0.789 & 0.653 & 0.731 & 0.815 & 0.819 & 0.753 & 0.757 & 0.370 & 0.382 & 0.458 & 0.705 & 0.518 & 0.595 \\
  CONTRIQUE \cite{madhusudana2022image} & \hred{\xmark} & 0.777 & 0.777 & 0.734 & 0.766 & 0.820 & 0.827 & 0.616 & 0.681 & 0.357 & 0.390 & 0.485 & 0.663 & 0.464 & 0.540 \\
  Re-IQA \cite{saha2023re} & \hred{\xmark} & 0.796 & \underline{0.827} & 0.690 & 0.761 & 0.825 & 0.830 & 0.714 & 0.741 & 0.306 & 0.302 & 0.406 & 0.675 & 0.554 & 0.625 \\
  ARNIQA \cite{agnolucci2024arniqa} & \hred{\xmark} & 0.798 & 0.810 & 0.699 & 0.765 & \underline{0.837} & \underline{0.846} & 0.632 & 0.673 & 0.351 & 0.361 & 0.437 & 0.682 & 0.487 & 0.585 \\
  CLIP-IQA$^{+}$ \cite{wang2023exploring} & \hred{\xmark} & 0.783 & 0.798 & 0.694 & 0.741 & 0.755 & 0.754 & 0.747 & 0.770 & 0.319 & 0.332 & 0.502 & 0.595 & 0.505 & 0.518 \\
  GRepQ \cite{srinath2024learning} & \hred{\xmark} & 0.810 & 0.818 & \textbf{0.751} & \underline{0.772} & 0.829 & 0.838 & \underline{0.799} & \underline{0.816} & \textbf{0.448} & \textbf{0.460} & 0.321 & 0.610 & \underline{0.581} & \underline{0.697} \\ \midrule
  \rowcolor{LightCyan} \textbf{\method} & \hgreen{\cmark} & \textbf{0.817} & \textbf{0.838} & 0.725 & \textbf{0.802} & \textbf{0.841} & \textbf{0.851} & \textbf{0.840} & \textbf{0.852} & \underline{0.410} & \underline{0.424} & \textbf{0.736} & \textbf{0.805} & \textbf{0.667} & \textbf{0.735} \\
  \bottomrule
  \end{tabular}}
  \vspace{-5pt}
  \caption{Quantitative results of the \textit{cross-dataset} evaluation setting. We employ the FLIVE \cite{ying2020patches} dataset to train the supervised methods. OU stands for Opinion-Unaware. Best and second-best scores are highlighted in bold and underlined, respectively.}
  \vspace{-5pt}
  \label{tab:cross_dataset}
\end{table*}

\section{Experimental Results} \label{sec:experimental_results}

\subsection{Quantitative Results} \label{sec:quantitative_results}
We conduct several experiments to compare the performance of \method with existing opinion-unaware and opinion-aware methods. In the supplementary material, we also study the robustness and explainability of our model.

\tit{Evaluation protocol} \label{sec:evaluation_protocol}
We evaluate the performance using Spearman's Rank-order Correlation Coefficient (\srcc) and Pearson's Linear Correlation Coefficient (\plcc), which measure prediction monotonicity and accuracy, respectively. Higher values of \srcc and \plcc correspond to better results. Following \cite{antkowiak2000final}, we pass the quality predictions through a four-parameter logistic non-linearity before computing \plcc.

We evaluate our method on several IQA datasets, each containing images annotated with human judgments of picture quality in the form of MOS. These datasets feature images with various types of distortions, including authentic degradations, artifacts from image restoration methods, and AI-generated content. Specifically, we consider four authentic datasets: \koniq \cite{hosu2020koniq10k}, \clive \cite{ghadiyaram2015massive}, \flive \cite{ying2020patches}, and \spaq \cite{fang2020perceptual}; two image restoration datasets: \cviu \cite{ma2017learning} and \pipal \cite{jinjin2020pipal}; and two AIGC datasets: \agiqaonek \cite{zhang2023perceptual} and \agiqathreek \cite{li2023agiqa}. Additional details on the datasets are provided in the supplementary material, where we also report experiments on images with synthetic distortions. Following \cite{madhusudana2022image, agnolucci2024arniqa, saha2023re}, we randomly split the datasets into 70\% for training, 10\% for validation, and 20\% for testing. For datasets that include reference images, namely the image restoration and synthetic datasets, we ensure that splits are made based on reference images to prevent content overlap. To mitigate selection bias in the training set, we repeat the training/testing process 10 times and report the median results. Due to its large size, for FLIVE, we follow \cite{madhusudana2022image, agnolucci2024arniqa, saha2023re} and use only the official train-validation-test split \cite{ying2020patches}.

We compare our approach to state-of-the-art methods in two settings: \textit{zero-shot} and \textit{cross-dataset}. Our method remains consistent across both settings; the only variation lies in the competing methods. For a fair comparison, we compute the results of the baselines using our evaluation protocol. For each baseline, we employ the official pre-trained model when available, or, otherwise, train the model following the procedure described in the original paper. In the \textit{zero-shot} setting, we compare with existing opinion-unaware methods. In addition, following \cite{babu2023no}, we consider opinion-aware approaches that can be modified to function without requiring MOS (indicated with the suffix ``-OU''). In particular, for GRepQ \cite{srinath2024learning}, we follow the zero-shot strategy detailed in the original paper. For methods based on an image encoder and a linear regressor, such as CONTRIQUE \cite{madhusudana2022image}, we extract the image features via the pre-trained encoder and then employ a NIQE-style framework to predict the quality scores, similar to \cite{babu2023no}. 
In the \textit{cross-dataset} setting, we evaluate the generalization capabilities of our model by comparing it with supervised opinion-aware methods on testing datasets different from the training one. Due to its large scale, we employ FLIVE as the training dataset for the baselines. Additionally, we report results using CLIVE and PIPAL in the supplementary material. For a fair comparison, we train LIQE \cite{zhang2023blind} using a ResNet50 backbone and restrict our analysis to methods that do not rely on an MLLM, as these models entail substantial computational demands.

\tit{Zero-shot setting}
We report the results for the \textit{zero-shot} setting in \cref{tab:zeroshot}. 
Our approach achieves the best performance on 13 out of 16 evaluation metrics and ranks second on the remaining 3, with SRCC improvements over the best-performing baseline of up to 9.2\%, observed on the CVIU dataset. Notably, \method sets the new state of the art for opinion-unaware approaches on authentic and image restoration datasets, proving the effectiveness of our training strategy. The improvement over CLIP-IQA highlights that our model generates more accurate quality-aware image representations than off-the-shelf CLIP models. Compared to GRepQ-OU, which is the strongest existing approach in most scenarios, the proposed method obtains better results on all the testing datasets excluding PIPAL. Moreover, while GRepQ-OU combines a low-level encoder with a high-level fine-tuned CLIP-based encoder, \method relies solely on CLIP, making it more straightforward and efficient. For reference, \cref{tab:zeroshot} also includes the performance of supervised opinion-aware methods trained on the training split of each testing dataset. We observe that all opinion-unaware methods fall behind supervised opinion-aware approaches when a training set is available, showing that there is still room for improving the performance of opinion-unaware models.

\tit{Cross-dataset setting}
\cref{tab:cross_dataset} shows the results for the \textit{cross-dataset} setting. Despite not requiring MOS, \method outperforms the baselines on 11 out of 14 evaluation metrics. Specifically, our method achieves excellent performance across datasets with various types of distortions, demonstrating its robustness. This makes our model well-suited for real-world applications where a training set is unavailable. Moreover, comparing \cref{tab:zeroshot,tab:cross_dataset} reveals two key observations. First, the performance of supervised opinion-aware models significantly decreases when tested on unseen datasets (\eg GRepQ on AIGC datasets), highlighting their limited generalization capabilities. Second, \method stands out as the only opinion-unaware approach to consistently obtain remarkable results even against supervised opinion-aware methods.

\subsection{Ablation Studies}
We conduct ablation studies to analyze the impact of different components of our training strategy, the importance of each loss term, and the contribution of each of the antonym prompts in the quality score computation. For simplicity, we only report the SRCC results on the authentic datasets.

\begin{table}
    \centering
    \setlength{\tabcolsep}{4.5pt}
    \begin{tabular}{lcccc} 
    \toprule
    Ablation & \koniq & \clive & \flive & \spaq \\ 
    \midrule
    D = 2 & \underline{0.815} & 0.718 & \underline{0.437} & \underline{0.834} \\
    L = 3 & 0.808 & \underline{0.719} & 0.382 & 0.788 \\
    quality-based & 0.745 & 0.708 & 0.396 & 0.756 \\ \midrule
    \rowcolor{LightCyan} \textbf{\method} & \textbf{0.817} & \textbf{0.725} & \textbf{0.442} & \textbf{0.841} \\
    \bottomrule
    \end{tabular}
    \vspace{-5pt}
    \caption{Ablation study on the training strategy. Best and second-best scores are highlighted in bold and underlined, respectively.}
    \label{tab:ablation_training_strategy}
    \vspace{-10pt}
\end{table}

\tit{Training strategy} We evaluate the performance achieved by modified versions of our approach: 1) \textit{$D\!=\!2$}: we apply two sequential degradations to each crop in \cref{sec:image_degradation_model} instead of just one; 2) \textit{$L\!=\!3$}: we consider only three levels of degradation in \cref{sec:image_degradation_model,sec:training_strategy} instead of five; 3) we compute the ranking loss using the predicted quality scores -- obtained with \cref{eq:inference} -- associated to each degraded crop instead of its similarity to the antonym prompts;
\cref{tab:ablation_training_strategy} shows the results. First, we notice that employing more than one distortion leads to slightly worse performance. We argue that this result stems from the synthetic degradation becoming too severe independently of the level of intensity, making it overly challenging for the model to rank the crops effectively. Moreover, considering only $L\!=\!3$ levels of degradation provides less information to the model during training compared to using five different levels and thus significantly worsens the results. Then, we observe that directly employing the predicted quality scores in the ranking loss instead of the similarity to the prompts achieves poor performance. We attribute this outcome to an increased discrepancy between CLIP's training and fine-tuning process. Indeed, while the predicted quality scores originate from two prompts (see \cref{eq:inference}), the proposed strategy considers multiple pairs of single images and texts, which we argue is more similar to the technique used for training CLIP \cite{radford2021learning}. 

\begin{table}[]
    \centering
    \setlength{\tabcolsep}{4.5pt}
    \resizebox{\linewidth}{!}{ 
    \begin{tabular}{ccccccc} 
    \toprule
    $\Lone$ & $\Ltwo$ & $\Lthree$ & \koniq & \clive & \flive & \spaq \\ 
    \midrule
    \hgreen{\cmark} & \hred{\xmark} & \hred{\xmark} & 0.733 & 0.668 & 0.368 & 0.703 \\
    \hred{\xmark} & \hgreen{\cmark} & \hred{\xmark} & 0.764 & 0.532 & 0.410 & 0.769 \\
    \hred{\xmark} & \hred{\xmark} & \hgreen{\cmark} & 0.805 & 0.714 & 0.420 & 0.812 \\
    \hgreen{\cmark} & \hgreen{\cmark} & \hred{\xmark} & 0.771 & 0.550 & 0.404 & 0.791 \\
    \hgreen{\cmark} & \hred{\xmark} & \hgreen{\cmark} & 0.800 & \underline{0.715} & 0.427 & 0.820 \\
    \hred{\xmark} & \hgreen{\cmark} & \hgreen{\cmark} & \underline{0.811} & 0.710 & \underline{0.434} & \underline{0.837} \\ \midrule
    \rowcolor{LightCyan} \hgreen{\cmark} & \hgreen{\cmark} & \hgreen{\cmark} & \textbf{0.817} & \textbf{0.725} & \textbf{0.442} & \textbf{0.841} \\
    \bottomrule
    \end{tabular}}
    \vspace{-5pt}
    \caption{Ablation study on the loss terms. Best and second-best scores are highlighted in bold and underlined, respectively.}
    \label{tab:ablation_loss}
    \vspace{-5pt}
\end{table}

\tit{Loss terms}
We study the importance of each loss term in \cref{eq:loss_final} and report the results in \cref{tab:ablation_loss}. First, we notice that using only $\Lone$ leads to a significant performance decrease, as $\Lone$ does not exploit the information provided by the intrinsic ranking of the increasingly degraded crops. Nevertheless, $\Lone$ consistently yields a positive impact when combined with any of the other loss terms. Then, we observe that, while $\Ltwo$ and $\Lthree$ differ only for the type of prompt they consider, $\Lthree$ proves to be significantly more critical for the training process. 
Nevertheless, \cref{tab:ablation_loss} shows that combining the three loss terms achieves the best results, proving that they are all crucial for training CLIP to generate accurate quality-aware image representations. 

\tit{Individual prompt contributions}
The results of the ablation studies on the training loss terms show that $\Lthree$ is more critical than $\Ltwo$ for the training process. We recall that $\Ltwo$ and $\Lthree$ involve the alignment between the images and the positive and negative prompts, respectively. This suggests that the similarity between the image and the negative prompt has a greater influence than that with the positive prompt on the quality score computation (as in \cref{eq:inference}). To support this hypothesis, we study the individual prompt contributions in obtaining the final quality scores.

We conduct an experiment where we directly use the similarity between the image and each of the antonym prompts as the quality score. This is possible because both the similarities and the quality scores range between 0 and 1. \cref{tab:individual_prompt_contributions} shows the results. We observe that the similarity between the negative prompt and the image provides significantly more information about its inherent quality than the positive prompt. This result supports our hypothesis and is consistent with the greater importance of $\Lthree$ in our training strategy. Nonetheless, \cref{tab:individual_prompt_contributions} also indicates that both prompts are essential for the quality score computation, as their combination results in the best performance.

We carry out an additional experiment to determine whether the discrepancy in the contributions of the positive and negative prompts arises from our training strategy or is inherent to CLIP itself. Specifically, we follow the experimental setting described above to evaluate the individual contributions of the prompts in the quality score computation of CLIP-IQA \cite{wang2023exploring}. We recall that CLIP-IQA employs an off-the-shelf CLIP model and computes the final quality scores using a strategy similar to \cref{eq:inference}. Our experiment reveals that using $T_p$ and $T_n$ individually results in a \srcc of $0.010$ and $0.571$, respectively, on the \koniq dataset. This outcome leads us to conclude that the similarity with the negative prompt inherently provides more meaningful information about image quality compared to using the positive prompt. We will investigate this finding more thoroughly in future work.

\begin{table}[t]
  \centering
  \setlength{\tabcolsep}{4.5pt}
  \begin{tabular}{cccccc} 
  \toprule
  $T_p$ & $T_n$ & \koniq & \clive & \flive & \spaq \\ 
  \midrule
  \hgreen{\cmark} & \hred{\xmark} & 0.230 & 0.063 & 0.064 & 0.059 \\
  \hred{\xmark} & \hgreen{\cmark} & \underline{0.661} & \underline{0.686} & \underline{0.332} & \underline{0.818} \\ \midrule
  \rowcolor{LightCyan} \hgreen{\cmark} & \hgreen{\cmark} & \textbf{0.817} & \textbf{0.725} & \textbf{0.442} & \textbf{0.841} \\
  \bottomrule
  \end{tabular}
  \vspace{-5pt}
  \caption{Analysis of the individual prompt contributions in the quality score computation. Best and second-best scores are highlighted in bold and underlined, respectively.}
  \label{tab:individual_prompt_contributions}
  \vspace{-5pt}
\end{table}

\section{Conclusion}
In this work, we propose QualiCLIP, a self-supervised opinion-unaware approach that enhances CLIP's ability to produce accurate quality-aware image representations. In particular, we design a quality-aware image-text alignment strategy that trains CLIP to rank increasingly synthetically degraded images based on their similarity with antonym prompts, while ensuring consistent representations for images with similar content and comparable quality. Compared to existing opinion-unaware methods, \method shows significant performance improvements across several datasets. Moreover, it is the only opinion-unaware approach that, in most cases, outperforms opinion-aware methods in cross-dataset experiments. Thus, we believe that \method could serve as a strong baseline for evaluating the generalization capabilities of future NR-IQA approaches.

\tit{Acknowledgments} \\ This work was partially supported by the European Commission under European Horizon 2020 Programme, grant number 951911 - AI4Media.

{
    \small
    \bibliographystyle{ieeenat_fullname}
    \bibliography{main}

\begin{thebibliography}{60}
\providecommand{\natexlab}[1]{#1}
\providecommand{\url}[1]{\texttt{#1}}
\expandafter\ifx\csname urlstyle\endcsname\relax
  \providecommand{\doi}[1]{doi: #1}\else
  \providecommand{\doi}{doi: \begingroup \urlstyle{rm}\Url}\fi

\bibitem[Agnolucci et~al.(2024{\natexlab{a}})Agnolucci, Galteri, Bertini, and Del~Bimbo]{agnolucci2024arniqa}
Lorenzo Agnolucci, Leonardo Galteri, Marco Bertini, and Alberto Del~Bimbo.
\newblock {ARNIQA: Learning Distortion Manifold for Image Quality Assessment}.
\newblock In \emph{Proceedings of the IEEE/CVF Winter Conference on Applications of Computer Vision}, pages 189--198, 2024{\natexlab{a}}.

\bibitem[Agnolucci et~al.(2024{\natexlab{b}})Agnolucci, Galteri, Bertini, and Del~Bimbo]{agnolucci2024reference}
Lorenzo Agnolucci, Leonardo Galteri, Marco Bertini, and Alberto Del~Bimbo.
\newblock {Reference-Based Restoration of Digitized Analog Videotapes}.
\newblock In \emph{Proceedings of the IEEE/CVF Winter Conference on Applications of Computer Vision}, pages 1659--1668, 2024{\natexlab{b}}.

\bibitem[Babu et~al.(2023)Babu, Kannan, and Soundararajan]{babu2023no}
Nithin~C Babu, Vignesh Kannan, and Rajiv Soundararajan.
\newblock {No Reference Opinion Unaware Quality Assessment of Authentically Distorted Images}.
\newblock In \emph{Proceedings of the IEEE/CVF Winter Conference on Applications of Computer Vision}, pages 2459--2468, 2023.

\bibitem[Chiu et~al.(2020)Chiu, Zhao, and Gurari]{chiu2020assessing}
Tai-Yin Chiu, Yinan Zhao, and Danna Gurari.
\newblock {Assessing Image Quality Issues for Real-World Problems}.
\newblock In \emph{proceedings of the IEEE/CVF conference on computer vision and pattern recognition}, pages 3646--3656, 2020.

\bibitem[Fang et~al.(2020)Fang, Zhu, Zeng, Ma, and Wang]{fang2020perceptual}
Yuming Fang, Hanwei Zhu, Yan Zeng, Kede Ma, and Zhou Wang.
\newblock {Perceptual Quality Assessment of Smartphone Photography}.
\newblock In \emph{Proceedings of the IEEE/CVF Conference on Computer Vision and Pattern Recognition}, pages 3677--3686, 2020.

\bibitem[Ghadiyaram and Bovik(2015)]{ghadiyaram2015massive}
Deepti Ghadiyaram and Alan~C Bovik.
\newblock {Massive Online Crowdsourced Study of Subjective and Objective Picture Quality}.
\newblock \emph{IEEE Transactions on Image Processing}, 25\penalty0 (1):\penalty0 372--387, 2015.

\bibitem[Golestaneh et~al.(2022)Golestaneh, Dadsetan, and Kitani]{golestaneh2022no}
S~Alireza Golestaneh, Saba Dadsetan, and Kris~M Kitani.
\newblock {No-Reference Image Quality Assessment via Transformers, Relative Ranking, and Self-Consistency}.
\newblock In \emph{Proceedings of the IEEE/CVF Winter Conference on Applications of Computer Vision}, pages 1220--1230, 2022.

\bibitem[Group(2000)]{antkowiak2000final}
Video Quality~Experts Group.
\newblock {Final Report from the Video Quality Experts Group on the Validation of Objective Models of Video Quality Assessment}.
\newblock 2000.

\bibitem[Gu et~al.(2019)Gu, Meng, Da, Xiang, and Pan]{gu2019no}
Jie Gu, Gaofeng Meng, Cheng Da, Shiming Xiang, and Chunhong Pan.
\newblock {No-Reference Image Quality Assessment with Reinforcement Recursive List-Wise Ranking}.
\newblock In \emph{Proceedings of the AAAI Conference on Artificial Intelligence}, pages 8336--8343, 2019.

\bibitem[He et~al.(2016)He, Zhang, Ren, and Sun]{he2016deep}
Kaiming He, Xiangyu Zhang, Shaoqing Ren, and Jian Sun.
\newblock {Deep Residual Learning for Image Recognition}.
\newblock In \emph{Proceedings of the IEEE conference on computer vision and pattern recognition}, pages 770--778, 2016.

\bibitem[{Hosu} et~al.(2020){Hosu}, {Lin}, {Sziranyi}, and {Saupe}]{hosu2020koniq10k}
V. {Hosu}, H. {Lin}, T. {Sziranyi}, and D. {Saupe}.
\newblock {KonIQ-10k: An Ecologically Valid Database for Deep Learning of Blind Image Quality Assessment}.
\newblock \emph{IEEE Transactions on Image Processing}, 29:\penalty0 4041--4056, 2020.

\bibitem[Jinjin et~al.(2020)Jinjin, Haoming, Haoyu, Xiaoxing, Ren, and Chao]{jinjin2020pipal}
Gu Jinjin, Cai Haoming, Chen Haoyu, Ye Xiaoxing, Jimmy~S Ren, and Dong Chao.
\newblock {PIPAL: a Large-Scale Image Quality Assessment Dataset for Perceptual Image Restoration}.
\newblock In \emph{European Conference on Computer Vision}, pages 633--651. Springer, 2020.

\bibitem[Ke et~al.(2023)Ke, Ye, Yu, Wu, Milanfar, and Yang]{ke2023vila}
Junjie Ke, Keren Ye, Jiahui Yu, Yonghui Wu, Peyman Milanfar, and Feng Yang.
\newblock {VILA: Learning Image Aesthetics from User Comments with Vision-Language Pretraining}.
\newblock In \emph{Proceedings of the IEEE/CVF Conference on Computer Vision and Pattern Recognition}, pages 10041--10051, 2023.

\bibitem[Larson and Chandler(2010)]{larson2010most}
Eric~C Larson and Damon~M Chandler.
\newblock {Most Apparent Distortion: Full-Reference Image Quality Assessment and the Role of Strategy}.
\newblock \emph{Journal of Electronic Imaging}, 19\penalty0 (1):\penalty0 011006--011006, 2010.

\bibitem[Li et~al.(2023)Li, Zhang, Wu, Sun, Min, Liu, Zhai, and Lin]{li2023agiqa}
Chunyi Li, Zicheng Zhang, Haoning Wu, Wei Sun, Xiongkuo Min, Xiaohong Liu, Guangtao Zhai, and Weisi Lin.
\newblock {AGIQA-3K: an Open Database for AI-Generated Image Quality Assessment}.
\newblock \emph{{IEEE Transactions on Circuits and Systems for Video Technology}}, 34\penalty0 (8):\penalty0 6833--6846, 2023.

\bibitem[Liang et~al.(2023)Liang, Li, Zhou, Feng, and Loy]{liang2023iterative}
Zhexin Liang, Chongyi Li, Shangchen Zhou, Ruicheng Feng, and Chen~Change Loy.
\newblock {Iterative Prompt Learning for Unsupervised Backlit Image Enhancement}.
\newblock In \emph{Proceedings of the IEEE/CVF International Conference on Computer Vision}, pages 8094--8103, 2023.

\bibitem[Liao et~al.(2001)Liao, Chen, Chung, et~al.]{liao2001fast}
Ping-Sung Liao, Tse-Sheng Chen, Pau-Choo Chung, et~al.
\newblock {A Fast Algorithm for Multilevel Thresholding}.
\newblock \emph{J. Inf. Sci. Eng.}, 17\penalty0 (5):\penalty0 713--727, 2001.

\bibitem[Lin et~al.(2019)Lin, Hosu, and Saupe]{kadid10k}
Hanhe Lin, Vlad Hosu, and Dietmar Saupe.
\newblock {KADID-10k: A Large-scale Artificially Distorted IQA Database}.
\newblock In \emph{2019 Tenth International Conference on Quality of Multimedia Experience (QoMEX)}, pages 1--3. IEEE, 2019.

\bibitem[Liu et~al.(2017)Liu, Van De~Weijer, and Bagdanov]{liu2017rankiqa}
Xialei Liu, Joost Van De~Weijer, and Andrew~D Bagdanov.
\newblock {RankIQA: Learning from Rankings for No-Reference Image Quality Assessment}.
\newblock In \emph{Proceedings of the IEEE International Conference on Computer Vision}, pages 1040--1049, 2017.

\bibitem[Loshchilov and Hutter(2017)]{loshchilov2017decoupled}
Ilya Loshchilov and Frank Hutter.
\newblock {Decoupled Weight Decay Regularization}.
\newblock \emph{arXiv preprint arXiv:1711.05101}, 2017.

\bibitem[Luo et~al.(2023)Luo, Gustafsson, Zhao, Sj{\"o}lund, and Sch{\"o}n]{luo2023controlling}
Ziwei Luo, Fredrik~K Gustafsson, Zheng Zhao, Jens Sj{\"o}lund, and Thomas~B Sch{\"o}n.
\newblock {Controlling Vision-Language Models for Universal Image Restoration}.
\newblock \emph{arXiv preprint arXiv:2310.01018}, 2023.

\bibitem[Ma et~al.(2017{\natexlab{a}})Ma, Yang, Yang, and Yang]{ma2017learning}
Chao Ma, Chih-Yuan Yang, Xiaokang Yang, and Ming-Hsuan Yang.
\newblock {Learning a No-Reference Quality Metric for Single-Image Super-Resolution}.
\newblock \emph{Computer Vision and Image Understanding}, 158:\penalty0 1--16, 2017{\natexlab{a}}.

\bibitem[Ma et~al.(2016{\natexlab{a}})Ma, Duanmu, Wu, Wang, Yong, Li, and Zhang]{ma2016waterloo}
Kede Ma, Zhengfang Duanmu, Qingbo Wu, Zhou Wang, Hongwei Yong, Hongliang Li, and Lei Zhang.
\newblock {Waterloo Exploration Database: New Challenges for Image Quality Assessment Models}.
\newblock \emph{IEEE Transactions on Image Processing}, 26\penalty0 (2):\penalty0 1004--1016, 2016{\natexlab{a}}.

\bibitem[Ma et~al.(2016{\natexlab{b}})Ma, Wu, Wang, Duanmu, Yong, Li, and Zhang]{ma2016group}
Kede Ma, Qingbo Wu, Zhou Wang, Zhengfang Duanmu, Hongwei Yong, Hongliang Li, and Lei Zhang.
\newblock {Group MAD Competition - A New Methodology to Compare Objective Image Quality Models}.
\newblock In \emph{Proceedings of the IEEE Conference on Computer Vision and Pattern Recognition}, pages 1664--1673, 2016{\natexlab{b}}.

\bibitem[Ma et~al.(2017{\natexlab{b}})Ma, Liu, Liu, Wang, and Tao]{ma2017dipiq}
Kede Ma, Wentao Liu, Tongliang Liu, Zhou Wang, and Dacheng Tao.
\newblock {dipIQ: Blind Image Quality Assessment by Learning-to-Rank Discriminable Image Pairs}.
\newblock \emph{IEEE Transactions on Image Processing}, 26\penalty0 (8):\penalty0 3951--3964, 2017{\natexlab{b}}.

\bibitem[Madhusudana et~al.(2022)Madhusudana, Birkbeck, Wang, Adsumilli, and Bovik]{madhusudana2022image}
Pavan~C Madhusudana, Neil Birkbeck, Yilin Wang, Balu Adsumilli, and Alan~C Bovik.
\newblock {Image Quality Assessment Using Contrastive Learning}.
\newblock \emph{IEEE Transactions on Image Processing}, 31:\penalty0 4149--4161, 2022.

\bibitem[Mistretta et~al.(2025)Mistretta, Baldrati, Agnolucci, Bertini, and Bagdanov]{mistretta2025cross}
Marco Mistretta, Alberto Baldrati, Lorenzo Agnolucci, Marco Bertini, and Andrew~D. Bagdanov.
\newblock {Cross the Gap: Exposing the Intra-modal Misalignment in {CLIP} via Modality Inversion}.
\newblock In \emph{The Thirteenth International Conference on Learning Representations}, 2025.

\bibitem[Mittal et~al.(2012{\natexlab{a}})Mittal, Moorthy, and Bovik]{mittal2012no}
Anish Mittal, Anush~Krishna Moorthy, and Alan~Conrad Bovik.
\newblock {No-Reference Image Quality Assessment in the Spatial Domain}.
\newblock \emph{IEEE Transactions on Image Processing}, 21\penalty0 (12):\penalty0 4695--4708, 2012{\natexlab{a}}.

\bibitem[Mittal et~al.(2012{\natexlab{b}})Mittal, Soundararajan, and Bovik]{mittal2012making}
Anish Mittal, Rajiv Soundararajan, and Alan~C Bovik.
\newblock {Making a ``Completely Blind” Image Quality Analyzer}.
\newblock \emph{IEEE Signal Processing Letters}, 20\penalty0 (3):\penalty0 209--212, 2012{\natexlab{b}}.

\bibitem[Ni et~al.(2024)Ni, Liu, Ding, Yang, Wang, and Wang]{ni2024opinion}
Zhangkai Ni, Yue Liu, Keyan Ding, Wenhan Yang, Hanli Wang, and Shiqi Wang.
\newblock {Opinion-Unaware Blind Image Quality Assessment using Multi-Scale Deep Feature Statistics}.
\newblock \emph{IEEE Transactions on Multimedia}, 2024.

\bibitem[Ponomarenko et~al.(2015)Ponomarenko, Jin, Ieremeiev, Lukin, Egiazarian, Astola, Vozel, Chehdi, Carli, Battisti, et~al.]{ponomarenko2015image}
Nikolay Ponomarenko, Lina Jin, Oleg Ieremeiev, Vladimir Lukin, Karen Egiazarian, Jaakko Astola, Benoit Vozel, Kacem Chehdi, Marco Carli, Federica Battisti, et~al.
\newblock {Image Database TID2013: Peculiarities, Results and Perspectives}.
\newblock \emph{Signal Processing: Image Communication}, 30:\penalty0 57--77, 2015.

\bibitem[Radford et~al.(2021)Radford, Kim, Hallacy, Ramesh, Goh, Agarwal, Sastry, Askell, Mishkin, Clark, et~al.]{radford2021learning}
Alec Radford, Jong~Wook Kim, Chris Hallacy, Aditya Ramesh, Gabriel Goh, Sandhini Agarwal, Girish Sastry, Amanda Askell, Pamela Mishkin, Jack Clark, et~al.
\newblock {Learning Transferable Visual Models from Natural Language Supervision}.
\newblock In \emph{International Conference on Machine Learning}, pages 8748--8763. PMLR, 2021.

\bibitem[Rombach et~al.(2022)Rombach, Blattmann, Lorenz, Esser, and Ommer]{rombach2022high}
Robin Rombach, Andreas Blattmann, Dominik Lorenz, Patrick Esser, and Bj{\"o}rn Ommer.
\newblock {High-Resolution Image Synthesis with Latent Diffusion Models}.
\newblock In \emph{Proceedings of the IEEE/CVF Conference on Computer Vision and Pattern Recognition}, pages 10684--10695, 2022.

\bibitem[Roy et~al.(2023)Roy, Mitra, Biswas, and Soundararajan]{roy2023test}
Subhadeep Roy, Shankhanil Mitra, Soma Biswas, and Rajiv Soundararajan.
\newblock {Test Time Adaptation for Blind Image Quality Assessment}.
\newblock In \emph{Proceedings of the IEEE/CVF International Conference on Computer Vision}, pages 16742--16751, 2023.

\bibitem[Saha et~al.(2023)Saha, Mishra, and Bovik]{saha2023re}
Avinab Saha, Sandeep Mishra, and Alan~C Bovik.
\newblock {Re-IQA: Unsupervised Learning for Image Quality Assessment in the Wild}.
\newblock In \emph{Proceedings of the IEEE/CVF Conference on Computer Vision and Pattern Recognition}, pages 5846--5855, 2023.

\bibitem[Selvaraju et~al.(2017)Selvaraju, Cogswell, Das, Vedantam, Parikh, and Batra]{selvaraju2017grad}
Ramprasaath~R Selvaraju, Michael Cogswell, Abhishek Das, Ramakrishna Vedantam, Devi Parikh, and Dhruv Batra.
\newblock {Grad-CAM: Visual Explanations from Deep Networks via Gradient-Based Localization}.
\newblock In \emph{Proceedings of the IEEE International Conference on Computer Vision}, pages 618--626, 2017.

\bibitem[Sheikh et~al.(2006)Sheikh, Sabir, and Bovik]{sheikh2006statistical}
Hamid~R Sheikh, Muhammad~F Sabir, and Alan~C Bovik.
\newblock {A Statistical Evaluation of Recent Full Reference Image Quality Assessment Algorithms}.
\newblock \emph{IEEE Transactions on Image Processing}, 15\penalty0 (11):\penalty0 3440--3451, 2006.

\bibitem[Shi et~al.(2024)Shi, Gao, and Qin]{shi2024transformer}
Jinsong Shi, Pan Gao, and Jie Qin.
\newblock {Transformer-based No-Reference Image Quality Assessment via Supervised Contrastive Learning}.
\newblock In \emph{Proceedings of the AAAI Conference on Artificial Intelligence}, pages 4829--4837, 2024.

\bibitem[Shin et~al.(2024)Shin, Lee, and Kim]{shin2024blind}
Nyeong-Ho Shin, Seon-Ho Lee, and Chang-Su Kim.
\newblock {Blind Image Quality Assessment Based on Geometric Order Learning}.
\newblock In \emph{Proceedings of the IEEE/CVF Conference on Computer Vision and Pattern Recognition}, pages 12799--12808, 2024.

\bibitem[Shukla et~al.(2024)Shukla, Upadhyay, Bhugra, and Sharma]{shukla2024opinion}
Ankit Shukla, Avinash Upadhyay, Swati Bhugra, and Manoj Sharma.
\newblock {Opinion Unaware Image Quality Assessment via Adversarial Convolutional Variational Autoencoder}.
\newblock In \emph{Proceedings of the IEEE/CVF Winter Conference on Applications of Computer Vision}, pages 2153--2163, 2024.

\bibitem[Srinath et~al.(2024)Srinath, Mitra, Rao, and Soundararajan]{srinath2024learning}
Suhas Srinath, Shankhanil Mitra, Shika Rao, and Rajiv Soundararajan.
\newblock {Learning Generalizable Perceptual Representations for Data-Efficient No-Reference Image Quality Assessment}.
\newblock In \emph{Proceedings of the IEEE/CVF Winter Conference on Applications of Computer Vision}, pages 22--31, 2024.

\bibitem[Su et~al.(2020)Su, Yan, Zhu, Zhang, Ge, Sun, and Zhang]{su2020blindly}
Shaolin Su, Qingsen Yan, Yu Zhu, Cheng Zhang, Xin Ge, Jinqiu Sun, and Yanning Zhang.
\newblock {Blindly Assess Image Quality in the Wild Guided by a Self-Adaptive Hyper Network}.
\newblock In \emph{Proceedings of the IEEE/CVF Conference on Computer Vision and Pattern Recognition}, pages 3667--3676, 2020.

\bibitem[Thomee et~al.(2016)Thomee, Shamma, Friedland, Elizalde, Ni, Poland, Borth, and Li]{thomee2016yfcc100m}
Bart Thomee, David~A Shamma, Gerald Friedland, Benjamin Elizalde, Karl Ni, Douglas Poland, Damian Borth, and Li-Jia Li.
\newblock {YFCC100M: The New Data in Multimedia Research}.
\newblock \emph{Communications of the ACM}, 59\penalty0 (2):\penalty0 64--73, 2016.

\bibitem[Thong et~al.(2022)Thong, Pereira, Parisot, Leonardis, and McDonagh]{thong2022content}
William Thong, Jose~Costa Pereira, Sarah Parisot, Ales Leonardis, and Steven McDonagh.
\newblock {Content-Diverse Comparisons Improve IQA}.
\newblock \emph{arXiv preprint arXiv:2211.05215}, 2022.

\bibitem[Udandarao et~al.(2023)Udandarao, Gupta, and Albanie]{udandarao2023sus}
Vishaal Udandarao, Ankush Gupta, and Samuel Albanie.
\newblock {SuS-X: Training-Free Name-Only Transfer of Vision-Language Models}.
\newblock In \emph{Proceedings of the IEEE/CVF International Conference on Computer Vision}, pages 2725--2736, 2023.

\bibitem[Van~der Maaten and Hinton(2008)]{van2008visualizing}
Laurens Van~der Maaten and Geoffrey Hinton.
\newblock {Visualizing Data Using t-SNE}.
\newblock \emph{Journal of Machine Learning Research}, 9\penalty0 (11), 2008.

\bibitem[Wang et~al.(2023)Wang, Chan, and Loy]{wang2023exploring}
Jianyi Wang, Kelvin~CK Chan, and Chen~Change Loy.
\newblock {Exploring CLIP for Assessing the Look and Feel of Images}.
\newblock In \emph{Proceedings of the AAAI Conference on Artificial Intelligence}, pages 2555--2563, 2023.

\bibitem[Wu et~al.(2023{\natexlab{a}})Wu, Liao, Hou, Chen, Zhang, Wang, Sun, Yan, and Lin]{wu2023exploring}
Haoning Wu, Liang Liao, Jingwen Hou, Chaofeng Chen, Erli Zhang, Annan Wang, Wenxiu Sun, Qiong Yan, and Weisi Lin.
\newblock {Exploring Opinion-Unaware Video Quality Assessment with Semantic Affinity Criterion}.
\newblock \emph{arXiv preprint arXiv:2302.13269}, 2023{\natexlab{a}}.

\bibitem[Wu et~al.(2023{\natexlab{b}})Wu, Liao, Wang, Chen, Hou, Sun, Yan, and Lin]{wu2023towardsrobust}
Haoning Wu, Liang Liao, Annan Wang, Chaofeng Chen, Jingwen Hou, Wenxiu Sun, Qiong Yan, and Weisi Lin.
\newblock {Towards Robust Text-Prompted Semantic Criterion for In-the-Wild Video Quality Assessment}.
\newblock \emph{arXiv preprint arXiv:2304.14672}, 2023{\natexlab{b}}.

\bibitem[Wu et~al.(2024{\natexlab{a}})Wu, Zhang, Zhang, Chen, Liao, Wang, Li, Sun, Yan, Zhai, et~al.]{wu2023q}
Haoning Wu, Zicheng Zhang, Erli Zhang, Chaofeng Chen, Liang Liao, Annan Wang, Chunyi Li, Wenxiu Sun, Qiong Yan, Guangtao Zhai, et~al.
\newblock {Q-Bench: A Benchmark for General-Purpose Foundation Models on Low-level Vision}.
\newblock In \emph{The Twelfth International Conference on Learning Representations}, 2024{\natexlab{a}}.

\bibitem[Wu et~al.(2024{\natexlab{b}})Wu, Zhang, Zhang, Chen, Liao, Li, Gao, Wang, Zhang, Sun, Yan, Min, Zhai, and Lin]{wu2024qalign}
Haoning Wu, Zicheng Zhang, Weixia Zhang, Chaofeng Chen, Liang Liao, Chunyi Li, Yixuan Gao, Annan Wang, Erli Zhang, Wenxiu Sun, Qiong Yan, Xiongkuo Min, Guangtao Zhai, and Weisi Lin.
\newblock {Q-Align: Teaching LMMs for Visual Scoring via Discrete Text-Defined Levels}.
\newblock In \emph{Proceedings of the 41st International Conference on Machine Learning}, pages 54015--54029. PMLR, 2024{\natexlab{b}}.

\bibitem[Xu et~al.(2024)Xu, Liao, Xiao, Chen, Wu, Yan, and Lin]{xu2024boosting}
Kangmin Xu, Liang Liao, Jing Xiao, Chaofeng Chen, Haoning Wu, Qiong Yan, and Weisi Lin.
\newblock {Boosting Image Quality Assessment through Efficient Transformer Adaptation with Local Feature Enhancement}.
\newblock In \emph{Proceedings of the IEEE/CVF Conference on Computer Vision and Pattern Recognition}, pages 2662--2672, 2024.

\bibitem[Ying et~al.(2020)Ying, Niu, Gupta, Mahajan, Ghadiyaram, and Bovik]{ying2020patches}
Zhenqiang Ying, Haoran Niu, Praful Gupta, Dhruv Mahajan, Deepti Ghadiyaram, and Alan Bovik.
\newblock {From Patches to Pictures (PaQ-2-PiQ): Mapping the Perceptual Space of Picture Quality}.
\newblock In \emph{Proceedings of the IEEE/CVF Conference on Computer Vision and Pattern Recognition}, pages 3575--3585, 2020.

\bibitem[You et~al.(2023)You, Li, Gu, Yin, Xue, and Dong]{you2023depicting}
Zhiyuan You, Zheyuan Li, Jinjin Gu, Zhenfei Yin, Tianfan Xue, and Chao Dong.
\newblock {Depicting Beyond Scores: Advancing Image Quality Assessment Through Multi-Modal Language Models}.
\newblock \emph{arXiv preprint arXiv:2312.08962}, 2023.

\bibitem[Zhang et~al.(2015)Zhang, Zhang, and Bovik]{zhang2015feature}
Lin Zhang, Lei Zhang, and Alan~C Bovik.
\newblock {A Feature-Enriched Completely Blind Image Quality Evaluator}.
\newblock \emph{IEEE Transactions on Image Processing}, 24\penalty0 (8):\penalty0 2579--2591, 2015.

\bibitem[Zhang et~al.(2023{\natexlab{a}})Zhang, Rao, and Agrawala]{zhang2023adding}
Lvmin Zhang, Anyi Rao, and Maneesh Agrawala.
\newblock {Adding Conditional Control to Text-to-Image Diffusion Models}.
\newblock In \emph{Proceedings of the IEEE/CVF International Conference on Computer Vision}, pages 3836--3847, 2023{\natexlab{a}}.

\bibitem[Zhang et~al.(2018)Zhang, Ma, Yan, Deng, and Wang]{zhang2018blind}
Weixia Zhang, Kede Ma, Jia Yan, Dexiang Deng, and Zhou Wang.
\newblock {Blind Image Quality Assessment using a Deep Bilinear Convolutional Neural Network}.
\newblock \emph{IEEE Transactions on Circuits and Systems for Video Technology}, 30\penalty0 (1):\penalty0 36--47, 2018.

\bibitem[Zhang et~al.(2023{\natexlab{b}})Zhang, Zhai, Wei, Yang, and Ma]{zhang2023blind}
Weixia Zhang, Guangtao Zhai, Ying Wei, Xiaokang Yang, and Kede Ma.
\newblock {Blind Image Quality Assessment via Vision-Language Correspondence: A Multitask Learning Perspective}.
\newblock In \emph{Proceedings of the IEEE/CVF Conference on Computer Vision and Pattern Recognition}, pages 14071--14081, 2023{\natexlab{b}}.

\bibitem[Zhang et~al.(2023{\natexlab{c}})Zhang, Li, Sun, Liu, Min, and Zhai]{zhang2023perceptual}
Zicheng Zhang, Chunyi Li, Wei Sun, Xiaohong Liu, Xiongkuo Min, and Guangtao Zhai.
\newblock {A Perceptual Quality Assessment Exploration for AIGC Images}.
\newblock In \emph{IEEE International Conference on Multimedia and Expo Workshops}, pages 440--445. IEEE, 2023{\natexlab{c}}.

\bibitem[Zhao et~al.(2023)Zhao, Yuan, Sun, Li, and Wen]{zhao2023quality}
Kai Zhao, Kun Yuan, Ming Sun, Mading Li, and Xing Wen.
\newblock {Quality-Aware Pre-Trained Models for Blind Image Quality Assessment}.
\newblock In \emph{Proceedings of the IEEE/CVF Conference on Computer Vision and Pattern Recognition}, pages 22302--22313, 2023.

\end{thebibliography}
}

\appendix

\clearpage
\maketitlesupplementary

\renewcommand{\thesection}{S\arabic{section}}
\renewcommand{\thefigure}{S\arabic{figure}}
\renewcommand{\thetable}{S\arabic{table}}
\renewcommand{\theequation}{S\arabic{equation}}

\setcounter{table}{0}
\setcounter{figure}{0}

\section*{Overview}
In this document, we present additional details and extended experimental analyses to complement the main paper. The supplementary material is organized as follows:
\begin{enumerate}[label=S\arabic*.]
    \item Datasets: we report the details of the datasets employed in the experiments;
    \item Additional Experimental Results:
        \begin{enumerate}[label=\theenumi\arabic*.]
            \item Quantitative Results: we provide the zero-shot results on synthetic datasets and the cross-dataset performance using the CLIVE and PIPAL datasets for training the baselines;
            \item Ablation Studies: we analyze the impact on the performance of the backbone of CLIP's image encoder;
            \item gMAD Competition: we conduct the gMAD competition against GRepQ and CLIP-IQA;
            \item t-SNE Visualization: we visualize the image representations of \method with t-SNE;
            \item Supervised \method: we extend our approach to leverage human annotations; 
            \item Inference Time: we evaluate the inference time of our model;
        \end{enumerate}
    \item Implementation Details: we provide the implementation details of the training strategy, the prompts, and the synthetic distortions.
    \item Limitations: we discuss the limitations of the proposed approach.
\end{enumerate}

\section{Datasets}
To carry out the experiments, we employ several types of datasets, namely authentic, image restoration, AIGC, and synthetic datasets. 
We rely on four authentic datasets: \koniq \cite{hosu2020koniq10k}, \clive \cite{ghadiyaram2015massive}, \flive \cite{ying2020patches}, and \spaq \cite{fang2020perceptual}. \koniq contains 10K images sampled from the YFCC100M \cite{thomee2016yfcc100m} database. \clive consists of 1162 images captured with a wide range of mobile devices. \flive is the largest existing dataset for NR-IQA and is composed of about 40K real-world images. \spaq comprises 11K high-resolution photos taken with several smartphones. Following \cite{fang2020perceptual}, we resize the \spaq images so that the shorter side is 512 pixels. 
We use two image restoration datasets: \cviu \cite{ma2017learning} and \pipal \cite{jinjin2020pipal}. \cviu stems from 30 reference images distorted with 9 super-resolution methods, resulting in 1620 images. \pipal comprises 23200 images degraded with 40 distortion types, including GAN-based super-resolution methods, and originates from 250 reference images.
We employ two AIGC datasets: \agiqaonek \cite{zhang2023perceptual} and \agiqathreek \cite{li2023agiqa}. \agiqaonek includes 1080 images generated with 2 diffusion models. \agiqathreek consists of 2982 images generated via 6 generative models, including auto-regressive and diffusion ones.
We consider four synthetic datasets: \live \cite{sheikh2006statistical}, \csiq \cite{larson2010most}, \tid \cite{ponomarenko2015image}, and \kadid \cite{kadid10k}. \live comprises 779 images degraded with 5 different distortion types at 5 levels of intensity, with 29 reference images as the base. \csiq originates from 30 reference images, each distorted with 6 distinct degradations at 5 intensity levels, resulting in 866 images. \tid and \kadid comprise 3000 and 10125 images degraded using 24 and 25 types of distortion across 5 different degrees of intensity, originating from 25 and 81 reference images, respectively.

\section{Additional Experimental Results}

\begin{table*}[t]
  \centering
  \begin{tabular}{lccccccccc} 
  \toprule
  \multicolumn{1}{c}{} & \multicolumn{1}{c}{} & \multicolumn{2}{c}{\live} & \multicolumn{2}{c}{\csiq} & \multicolumn{2}{c}{\tid} & \multicolumn{2}{c}{\kadid} \\
  \multicolumn{1}{l}{Method} & OU & \srcc & \plcc & \srcc & \plcc & \srcc & \plcc & \srcc & \plcc \\ 
  \midrule
  CONTRIQUE \cite{madhusudana2022image} & \hred{\xmark} & 0.962 & 0.967 & 0.946 & 0.960 & 0.821 & 0.843 & 0.927 & 0.928 \\
  Re-IQA \cite{saha2023re} & \hred{\xmark} & 0.954 & 0.956 & 0.946 & 0.954 & 0.873 & 0.895 & 0.867 & 0.865 \\
  ARNIQA \cite{agnolucci2024arniqa} & \hred{\xmark} & 0.965 & 0.970 & 0.961 & 0.969 & 0.874 & 0.896 & 0.905 & 0.908 \\
  CLIP-IQA$^{+}$ \cite{wang2023exploring} & \hred{\xmark} & 0.950 & 0.946 & 0.862 & 0.882 & 0.804 & 0.834 & 0.818 & 0.822 \\ 
  GRepQ \cite{srinath2024learning} & \hred{\xmark} & 0.926 & 0.931 & 0.844 & 0.871 & 0.668 & 0.681 & 0.808 & 0.797 \\ \midrule
  NIQE \cite{mittal2012making} & \hgreen{\cmark} & \underline{0.903} & \underline{0.906} & 0.642 & 0.727 & 0.308 & 0.419 & 0.380 & 0.442 \\
  IL-NIQE \cite{zhang2015feature} & \hgreen{\cmark} & 0.878 & 0.884 & \textbf{0.848} & \textbf{0.883} & 0.504 & 0.636 & 0.557 & 0.599 \\
  CL-MI \cite{babu2023no} & \hgreen{\cmark} & 0.749 & 0.731 & 0.619 & 0.615 & 0.249 & 0.317 & 0.518 & 0.529 \\
  MDFS \cite{ni2024opinion} & \hgreen{\cmark} & \textbf{0.926} & \textbf{0.930} & 0.797 & 0.829 & \underline{0.554} & \underline{0.657} & 0.608 & 0.634 \\
  CONTRIQUE-OU \cite{madhusudana2022image} & \hgreen{\cmark} & 0.841 & 0.840 & 0.684 & 0.700 & 0.321 & 0.344 & 0.552 & 0.564 \\
  Re-IQA-OU \cite{saha2023re} & \hgreen{\cmark} & 0.781 & 0.779 & 0.713 & 0.725 & 0.284 & 0.339 & 0.516 & 0.534 \\
  ARNIQA-OU \cite{agnolucci2024arniqa} & \hgreen{\cmark} & 0.853 & 0.852 & \underline{0.832} & 0.810 & 0.467 & 0.527 & \underline{0.632} & \underline{0.638} \\
  CLIP-IQA \cite{wang2023exploring} & \hgreen{\cmark} & 0.626 & 0.651 & 0.748 & 0.805 & 0.525 & 0.632 & 0.465 & 0.473 \\
  GRepQ-OU \cite{srinath2024learning} & \hgreen{\cmark} & 0.714 & 0.698 & 0.740 & 0.742 & 0.423 & 0.565 & 0.416 & 0.463 \\ \midrule
  \rowcolor{LightCyan} \textbf{\method} & \hgreen{\cmark} & 0.898 & 0.885 & 0.804 & \underline{0.842} & \textbf{0.651} & \textbf{0.732} & \textbf{0.654} & \textbf{0.664} \\
  \bottomrule
  \end{tabular}
  \vspace{-5pt}
  \caption{Quantitative results of the \textit{zero-shot} evaluation setting using synthetic datasets. OU stands for Opinion-Unaware. Best and second-best scores for OU methods are highlighted in bold and underlined, respectively. The suffix ``-OU'' indicates approaches modified to be opinion-unaware (see Sec. \textcolor{iccvblue}{4.1}). For reference, we report the performance of supervised methods (\ie OU$=\!\hred{\xmark}$) trained on the training split of each testing dataset.}
  \label{tab:zeroshot_synthetic}
\end{table*}

\begin{table*}[t]
  \centering
  \setlength{\tabcolsep}{4.5pt}
  \resizebox{\linewidth}{!}{ 
  \begin{tabular}{lccccccccccccccc} 
  \toprule
  \multicolumn{2}{c}{} & \multicolumn{6}{c}{\textit{Authentic Datasets}} & \multicolumn{4}{c}{\textit{Image Restoration Datasets}} & \multicolumn{4}{c}{\textit{AIGC Datasets}} \\ 
  \cmidrule(lr){3-8} \cmidrule(lr){9-12} \cmidrule(lr){13-16}
  \multicolumn{1}{c}{} & \multicolumn{1}{c}{} & \multicolumn{2}{c}{\koniq} & \multicolumn{2}{c}{\flive} & \multicolumn{2}{c}{\spaq} & \multicolumn{2}{c}{\cviu} & \multicolumn{2}{c}{\pipal} & \multicolumn{2}{c}{\agiqaonek} & \multicolumn{2}{c}{\agiqathreek} \\
  \multicolumn{1}{l}{Method} & \multicolumn{1}{l}{OU} & \srcc & \plcc & \srcc & \plcc & \srcc & \plcc & \srcc & \plcc & \srcc & \plcc & \srcc & \plcc & \srcc & \plcc \\ 
  \midrule
  HyperIQA \cite{su2020blindly} & \hred{\xmark} & 0.734 & 0.773 & 0.328 & 0.483 & 0.771 & 0.796 & 0.632 & 0.642 & 0.361 & 0.384 & 0.183 & 0.351 & 0.594 & 0.686 \\
  DBCNN \cite{zhang2018blind} & \hred{\xmark} & 0.703 & 0.755 & 0.299 & 0.435 & 0.806 & 0.816 & 0.728 & 0.748 & 0.332 & 0.341 & \underline{0.445} & \underline{0.693} & 0.448 & 0.558 \\
  TReS \cite{golestaneh2022no} & \hred{\xmark} & 0.727 & 0.756 & 0.359 & 0.494 & 0.858 & 0.863 & 0.713 & 0.739 & \underline{0.397} & \textbf{0.453} & 0.364 & 0.643 & 0.549 & 0.666 \\
  LIQE \cite{zhang2023blind} & \hred{\xmark} & \underline{0.810} & 0.787 & 0.277 & 0.271 & 0.830 & 0.840 & 0.704 & 0.717 & \textbf{0.446} & \underline{0.445} & 0.417 & 0.609 & \underline{0.653} & \underline{0.732} \\
  QCN \cite{shin2024blind} & \hred{\xmark} & 0.762 & \underline{0.835} & 0.428 & 0.537 & \underline{0.860} & \underline{0.867} & 0.765 & 0.770 & 0.393 & 0.413 & 0.337 & 0.677 & 0.584 & 0.729 \\
  CONTRIQUE \cite{madhusudana2022image} & \hred{\xmark} & 0.736 & 0.749 & 0.399 & 0.527 & 0.834 & 0.842 & 0.583 & 0.601 & 0.310 & 0.370 & 0.256 & 0.393 & 0.503 & 0.597 \\
  Re-IQA \cite{saha2023re} & \hred{\xmark} & 0.600 & 0.618 & 0.252 & 0.341 & 0.820 & 0.829 & 0.576 & 0.685 & 0.360 & 0.386 & 0.330 & 0.548 & 0.431 & 0.516 \\
  ARNIQA \cite{agnolucci2024arniqa} & \hred{\xmark} & 0.745 & 0.775 & \underline{0.434} & 0.526 & \textbf{0.861} & \textbf{0.871} & 0.610 & 0.643 & 0.369 & 0.382 & 0.320 & 0.629 & 0.526 & 0.649 \\
  CLIP-IQA$^{+}$ \cite{wang2023exploring} & \hred{\xmark} & 0.769 & 0.808 & 0.410 & \underline{0.547} & 0.846 & 0.854 & \underline{0.837} & \underline{0.851} & 0.369 & 0.385 & 0.395 & 0.584 & 0.637 & 0.712 \\
  GRepQ \cite{srinath2024learning} & \hred{\xmark} & 0.742 & 0.750 & 0.391 & 0.464 & 0.855 & 0.865 & 0.701 & 0.717 & 0.368 & 0.385 & 0.346 & 0.589 & 0.616 & 0.730 \\ \midrule
  \rowcolor{LightCyan} \textbf{\method} & \hgreen{\cmark} & \textbf{0.817} & \textbf{0.838} & \textbf{0.442} & \textbf{0.556} & 0.841 & 0.851 & \textbf{0.840} & \textbf{0.852} & 0.412 & 0.422 & \textbf{0.736} & \textbf{0.805} & \textbf{0.667} & \textbf{0.735} \\
  \bottomrule
  \end{tabular}}
  \vspace{-5pt}
  \caption{Quantitative results of the \textit{cross-dataset} evaluation setting. We employ the CLIVE \cite{ghadiyaram2015massive} dataset to train the supervised methods. OU stands for Opinion-Unaware. Best and second-best scores are highlighted in bold and underlined, respectively.}
  \label{tab:cross_dataset_clive}
\end{table*}

\begin{table*}[t]
  \centering
  \setlength{\tabcolsep}{4.5pt}
  \resizebox{\linewidth}{!}{ 
  \begin{tabular}{lccccccccccccccc} 
  \toprule
  \multicolumn{2}{c}{} & \multicolumn{8}{c}{\textit{Authentic Datasets}} & \multicolumn{2}{c}{\textit{IR Dataset}} & \multicolumn{4}{c}{\textit{AIGC Datasets}} \\ 
  \cmidrule(lr){3-10} \cmidrule(lr){11-12} \cmidrule(lr){13-16}
  \multicolumn{1}{c}{} & \multicolumn{1}{c}{} & \multicolumn{2}{c}{\koniq} & \multicolumn{2}{c}{\clive} & \multicolumn{2}{c}{\flive} & \multicolumn{2}{c}{\spaq} & \multicolumn{2}{c}{\cviu} & \multicolumn{2}{c}{\agiqaonek} & \multicolumn{2}{c}{\agiqathreek} \\
  \multicolumn{1}{l}{Method} & \multicolumn{1}{l}{OU} & \srcc & \plcc & \srcc & \plcc & \srcc & \plcc & \srcc & \plcc & \srcc & \plcc & \srcc & \plcc & \srcc & \plcc \\ 
  \midrule
  HyperIQA \cite{su2020blindly} & \hred{\xmark} & 0.570 & 0.564 & 0.560 & 0.642 & 0.246 & 0.310 & 0.752 & 0.762 & \underline{0.821} & \underline{0.835} & \underline{0.493} & 0.486 & 0.497 & 0.535 \\
  DBCNN \cite{zhang2018blind} & \hred{\xmark} & \underline{0.718} & \underline{0.744} & 0.570 & 0.638 & 0.267 & 0.378 & 0.779 & 0.795 & 0.803 & 0.814 & 0.427 & 0.651 & 0.516 & 0.601 \\
  LIQE \cite{zhang2023blind} & \hred{\xmark} & 0.681 & 0.696 & \underline{0.657} & \underline{0.691} & 0.138 & 0.178 & 0.692 & 0.690 & 0.797 & 0.810 & 0.413 & 0.465 & 0.560 & 0.621 \\
  QCN \cite{shin2024blind} & \hred{\xmark} & 0.584 & 0.663 & 0.463 & 0.559 & \underline{0.292} & \underline{0.408} & \underline{0.785} & \underline{0.797} & 0.766 & 0.767 & 0.484 & \underline{0.679} & 0.563 & 0.650 \\
  CONTRIQUE \cite{madhusudana2022image} & \hred{\xmark} & 0.445 & 0.460 & 0.342 & 0.408 & 0.228 & 0.308 & 0.596 & 0.600 & 0.719 & 0.723 & 0.311 & 0.535 & 0.344 & 0.392 \\
  Re-IQA \cite{saha2023re} & \hred{\xmark} & 0.351 & 0.376 & 0.297 & 0.358 & 0.206 & 0.273 & 0.633 & 0.641 & 0.670 & 0.692 & 0.291 & 0.410 & 0.317 & 0.375 \\
  ARNIQA \cite{agnolucci2024arniqa} & \hred{\xmark} & 0.246 & 0.325 & 0.389 & 0.486 & 0.224 & 0.252 & 0.614 & 0.644 & 0.757 & 0.768 & 0.173 & 0.564 & 0.557 & 0.629 \\
  CLIP-IQA$^{+}$ \cite{wang2023exploring} & \hred{\xmark} & 0.568 & 0.586 & 0.551 & 0.597 & 0.231 & 0.313 & 0.758 & 0.766 & 0.748 & 0.764 & 0.442 & 0.560 & \textbf{0.679} & \underline{0.720} \\
  GRepQ \cite{srinath2024learning} & \hred{\xmark} & 0.679 & 0.682 & 0.536 & 0.583 & 0.285 & 0.345 & 0.754 & 0.765 & 0.792 & 0.820 & 0.174 & 0.591 & 0.573 & 0.679 \\ \midrule
  \rowcolor{LightCyan} \textbf{\method} & \hgreen{\cmark} & \textbf{0.817} & \textbf{0.838} & \textbf{0.725} & \textbf{0.802} & \textbf{0.442} & \textbf{0.556} & \textbf{0.841} & \textbf{0.851} & \textbf{0.840} & \textbf{0.852} & \textbf{0.736} & \textbf{0.805} & \underline{0.667} & \textbf{0.735} \\
  \bottomrule
  \end{tabular}}
  \vspace{-5pt}
  \caption{Quantitative results of the \textit{cross-dataset} evaluation setting. We employ the PIPAL \cite{jinjin2020pipal} dataset to train the supervised methods. OU stands for Opinion-Unaware. IR signifies Image Restoration. Best and second-best scores are highlighted in bold and underlined, respectively.}
  \label{tab:cross_dataset_pipal}
\end{table*}

\subsection{Quantitative Results}

\tit{Zero-shot setting}
In \cref{tab:zeroshot_synthetic}, we compare the performance of our model with existing opinion-unaware approaches on synthetic datasets. We observe that our method achieves competitive performance, obtaining the most consistent results among the considered approaches. However, as also observed in Sec. \textcolor{iccvblue}{4.1}, supervised opinion-aware methods achieve better performance than opinion-unaware ones when a training set is available.

\tit{Cross-dataset setting}
\cref{tab:cross_dataset_clive,tab:cross_dataset_pipal} present the results of the \textit{cross-dataset} setting when employing CLIVE \cite{ghadiyaram2015massive} and PIPAL \cite{jinjin2020pipal}, respectively, to train the supervised baselines.
Note that we do not report the performance of TReS \cite{golestaneh2022no} with PIPAL as the training dataset as there is no public pre-trained model available. The results show that \method achieves excellent performance regardless of the dataset used to train the baselines. Moreover, we observe that the opinion-aware approaches generally perform worse when trained on PIPAL compared to CLIVE. We attribute this outcome to the nature of the degradation types included in PIPAL, which are different from those in the other testing datasets. This result highlights the sensitivity of the supervised opinion-aware methods to the training data, and further confirms their limited generalization capabilities.

\subsection{Ablation Studies}

\tit{Backbone of CLIP's image encoder}
Following \cite{wang2023exploring}, we evaluate the impact of the backbone architecture of CLIP's image encoder on performance. Specifically, we examine the ResNet50 and ViT-B/32 backbones. In \cref{tab:backbone} we compare the results of \method with CLIP-IQA \cite{wang2023exploring}, which leverages an off-the-shelf CLIP model. As also observed by \citet{wang2023exploring}, relying on the ViT-B/32 significantly hinders the performance over the ResNet50. This outcome stems from the stronger inductive bias of convolutional networks compared to transformers, which are more sensitive to the removal of the positional embedding. Nevertheless, for both backbones, we observe that \method outperforms CLIP-IQA.

\begin{table}[t]
  \centering
  \setlength{\tabcolsep}{5pt}
  \resizebox{\linewidth}{!}{ 
  \begin{tabular}{clcccc} 
  \toprule
   & Method & \koniq & \clive & \flive & \spaq \\ 
  \midrule
   \multirow{2}{*}{\rotatebox[origin=c]{90}{RN50}} & CLIP-IQA & 0.695 & 0.656 & 0.344 & 0.733 \\
    & \cellcolor{LightCyan}\textbf{\method} & \cellcolor{LightCyan}\textbf{0.817} & \cellcolor{LightCyan}\textbf{0.725} & \cellcolor{LightCyan}\textbf{0.442} & \cellcolor{LightCyan}\textbf{0.841} \\ \midrule
   \multirow{2}{*}{\rotatebox[origin=c]{90}{B/32}} & CLIP-IQA & 0.390 & 0.328 & 0.211 & 0.608 \\
    & \cellcolor{LightCyan}\textbf{\method} & \cellcolor{LightCyan}\textbf{0.626} & \cellcolor{LightCyan}\textbf{0.334} & \cellcolor{LightCyan}\textbf{0.217} & \cellcolor{LightCyan}\textbf{0.782} \\
  \bottomrule
  \end{tabular}}
  \vspace{-5pt}
  \caption{Ablation study on the CLIP backbone. RN50 and B/32 stand for ResNet50 and ViT-B/32, respectively. Best SRCC scores for each backbone are highlighted in bold.}
  \label{tab:backbone}
\end{table}

\subsection{gMAD Competition}

\begin{figure}[t]
    \centering
    \begin{subfigure}{0.485\columnwidth}
        \includegraphics[width=\linewidth]{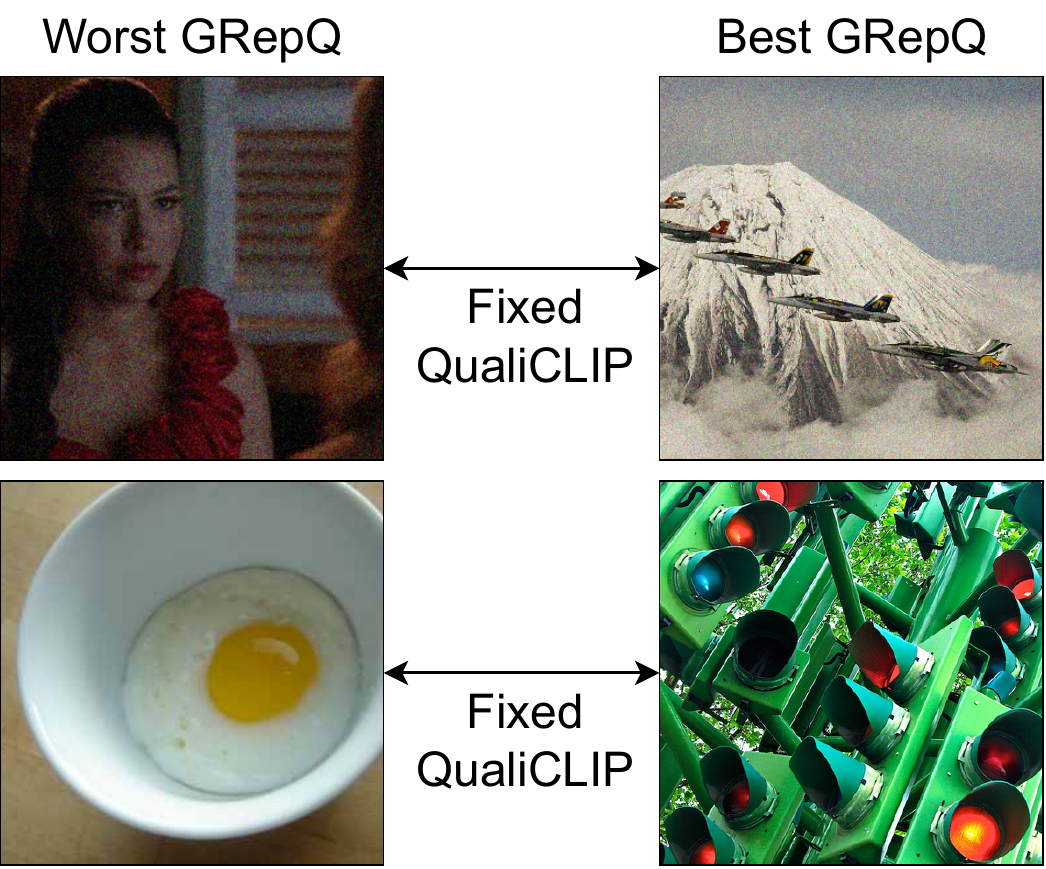}
        \caption{}
        \label{fig:ours_def}
    \end{subfigure}
    \hfill
    \begin{subfigure}{0.485\columnwidth}
        \includegraphics[width=\linewidth]{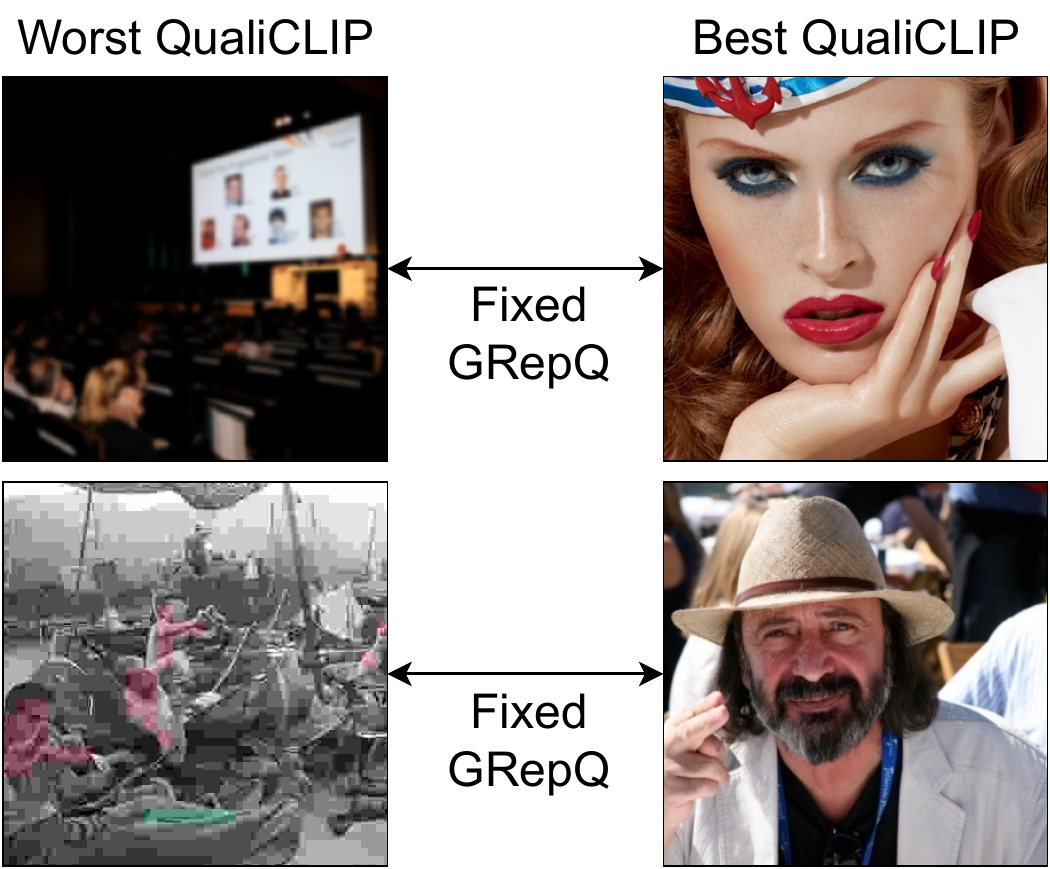}
        \caption{}
        \label{fig:ours_att}
    \end{subfigure}
    \vspace{-5pt}
    \caption{gMAD competition results between \method and GRepQ \cite{srinath2024learning}. (a): Fixed \method at a low- (top) and high-quality (bottom) level, respectively. (b): Fixed GRepQ at a low- (top) and high-quality (bottom) level, respectively.}
    \label{fig:gmad_grepq}
\end{figure}

\begin{figure}[t]
    \centering
    \begin{subfigure}{0.485\columnwidth}
        \includegraphics[width=\linewidth]{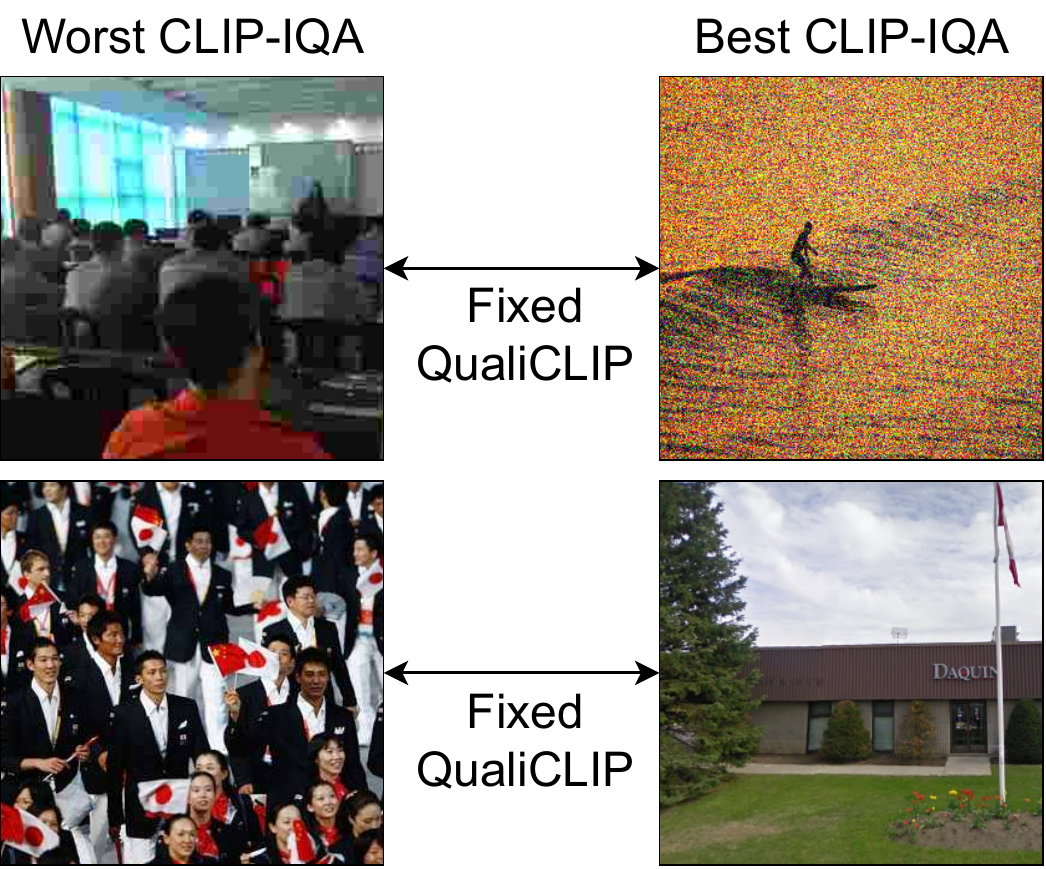}
        \caption{}
        \label{fig:ours_def_clipiqa}
    \end{subfigure}
    \hfill
    \begin{subfigure}{0.485\columnwidth}
        \includegraphics[width=\linewidth]{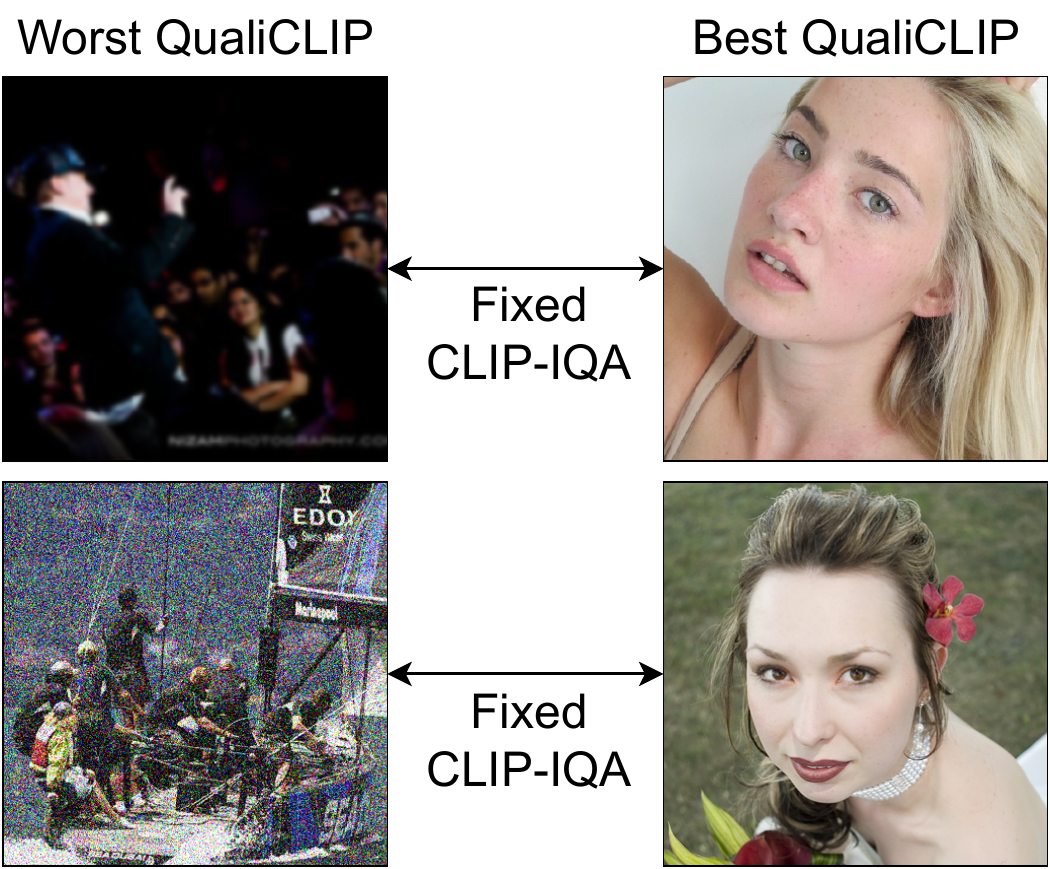}
        \caption{}
        \label{fig:ours_att_clipiqa}
    \end{subfigure}
    \vspace{-5pt}
    \caption{gMAD \cite{ma2016group} competition results between \method and CLIP-IQA \cite{wang2023exploring}. (a): Fixed \method at a low- (top) and high-quality (bottom) level, respectively. (b): Fixed CLIP-IQA at a low- (top) and high-quality (bottom) level, respectively.}
    \label{fig:gmad_clipiqa}
\end{figure}

To assess the robustness of our model, we carry out the group maximum differentiation (gMAD) competition \cite{ma2016group}. In particular, we compare \method against GRepQ and CLIP-IQA using the Waterloo Exploration Database \cite{ma2016waterloo} dataset, which comprises 95K synthetically degraded images without MOS annotations. In this evaluation, one model is fixed to function as a defender, and its quality predictions are grouped into two distinct levels. The other model assumes the role of the attacker, tasked with identifying image pairs within each level that exhibit the greatest quality difference. For a model to demonstrate robustness, the selected image pairs should show comparable quality when acting as the defender while exhibiting a notable quality disparity when assuming the role of the attacker. We observe that when we fix \method at a low-quality level (\cref{fig:ours_def} top), GRepQ fails to find picture pairs with an obvious quality difference. When considering a high-quality level (\cref{fig:ours_def} bottom), the image pair identified by GRepQ shows a slight quality gap. However, when assuming the role of the attacker (\cref{fig:ours_att}), \method successfully exposes the failures of GRepQ, as it pinpoints image pairs displaying a significant quality disparity. \cref{fig:gmad_clipiqa} shows that the same considerations can be drawn when analyzing the results of the gMAD competition between \method and CLIP-IQA. Hence, our approach demonstrates greater robustness than GRepQ and CLIP-IQA.

\begin{figure}[t]
    \centering
    \begin{subfigure}{\columnwidth}
        \centering
        \includegraphics[width=0.9\columnwidth]{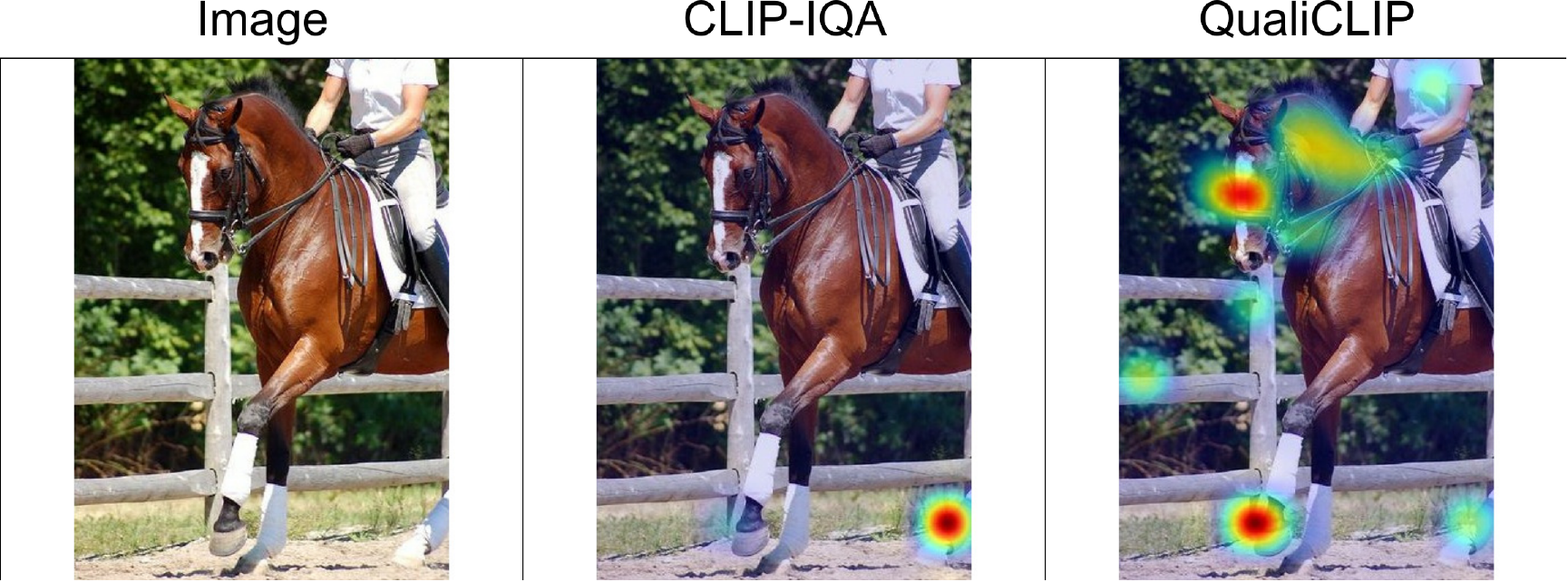}
        \caption{Positive prompt}
        \label{fig:gradcam_positive}
    \end{subfigure}
    \hfill
    \begin{subfigure}{\columnwidth}
        \centering
        \includegraphics[width=0.9\columnwidth]{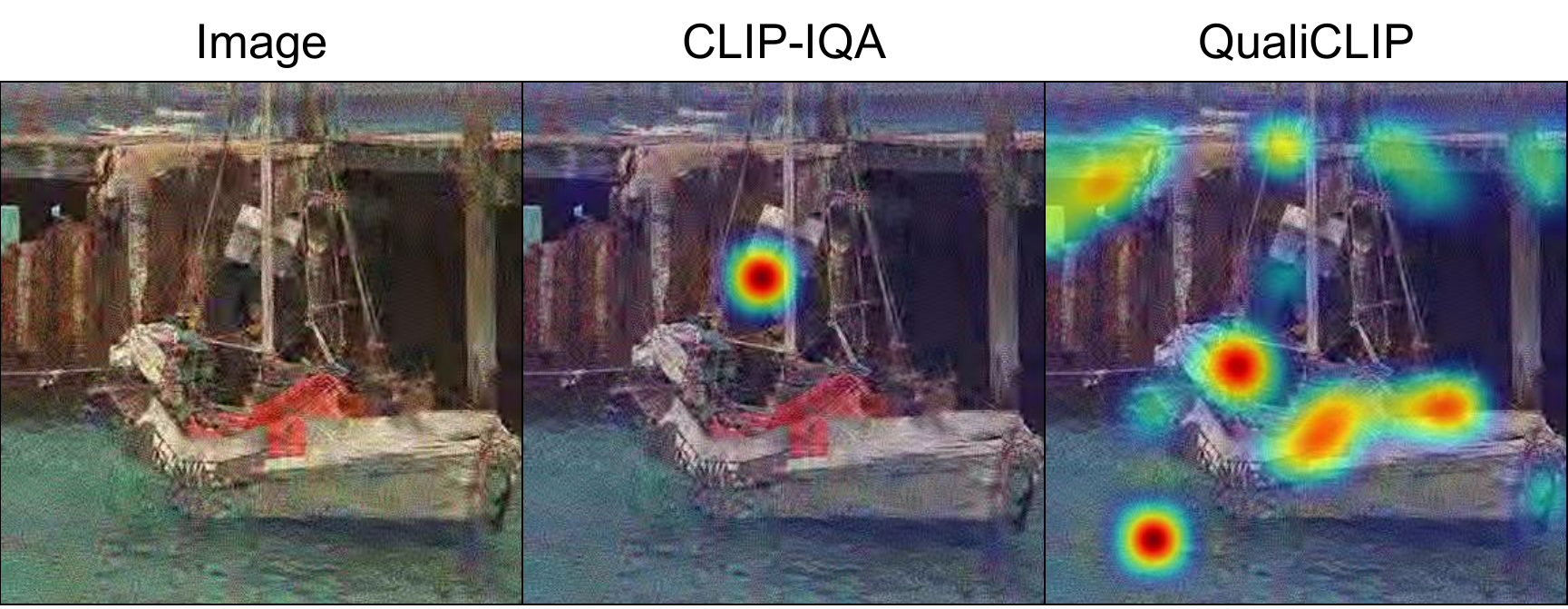}
        \caption{Negative prompt}
        \label{fig:gradcam_negative}
    \end{subfigure}
    \vspace{-15pt}
    \caption{gradCAM visualization of the most important regions of the input image for each of the antonym prompts.}
    \vspace{-10pt}
    \label{fig:gradcam}
\end{figure}

\subsection{gradCAM visualization}
We evaluate the explainability of our model and CLIP-IQA via a gradCAM \cite{selvaraju2017grad} visualization. gradCAM is a visualization technique aimed at understanding which regions of an input image are most influential for a model's decision by studying the gradients of a given layer. We employ gradCAM to produce a heatmap of the regions of the image that activate the most for each of the antonym prompts. We employ ``\textit{Good photo}'' and ``\textit{Bad photo}'' as the positive and negative prompts, respectively. Following \cite{selvaraju2017grad}, we consider the last convolutional layer of the ResNet50 backbone. \cref{fig:gradcam_positive} shows the result for the positive prompt. We observe that, compared to CLIP-IQA, our model leads to a better alignment with high-quality areas of the image, such as the head of the horse. Similarly, \cref{fig:gradcam_negative} illustrates that \method focuses on the most degraded parts of the images when considering the negative prompt, in contrast with CLIP-IQA. 
The improved alignment between the antonym prompts and the corresponding regions of the images makes \method more easily explainable than CLIP-IQA.

\subsection{t-SNE Visualization}

\begin{figure}[t]
    \centering
    \begin{subfigure}{0.875\linewidth}
        \includegraphics[width=\linewidth]{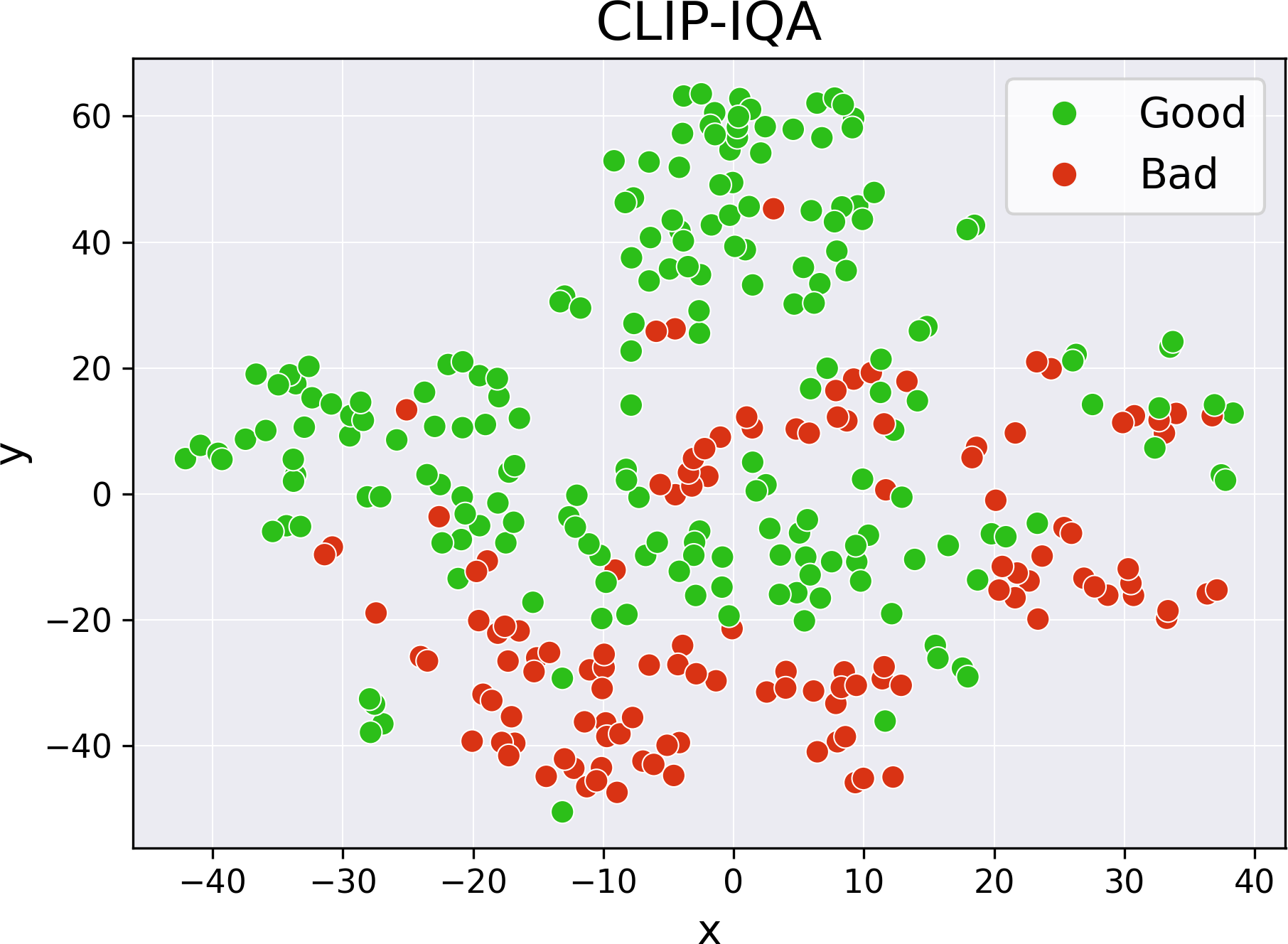}
        \caption{\vspace{5pt}}
        \label{fig:tsne_clipiqa}
    \end{subfigure}
    \begin{subfigure}{0.875\linewidth}
        \includegraphics[width=\linewidth]{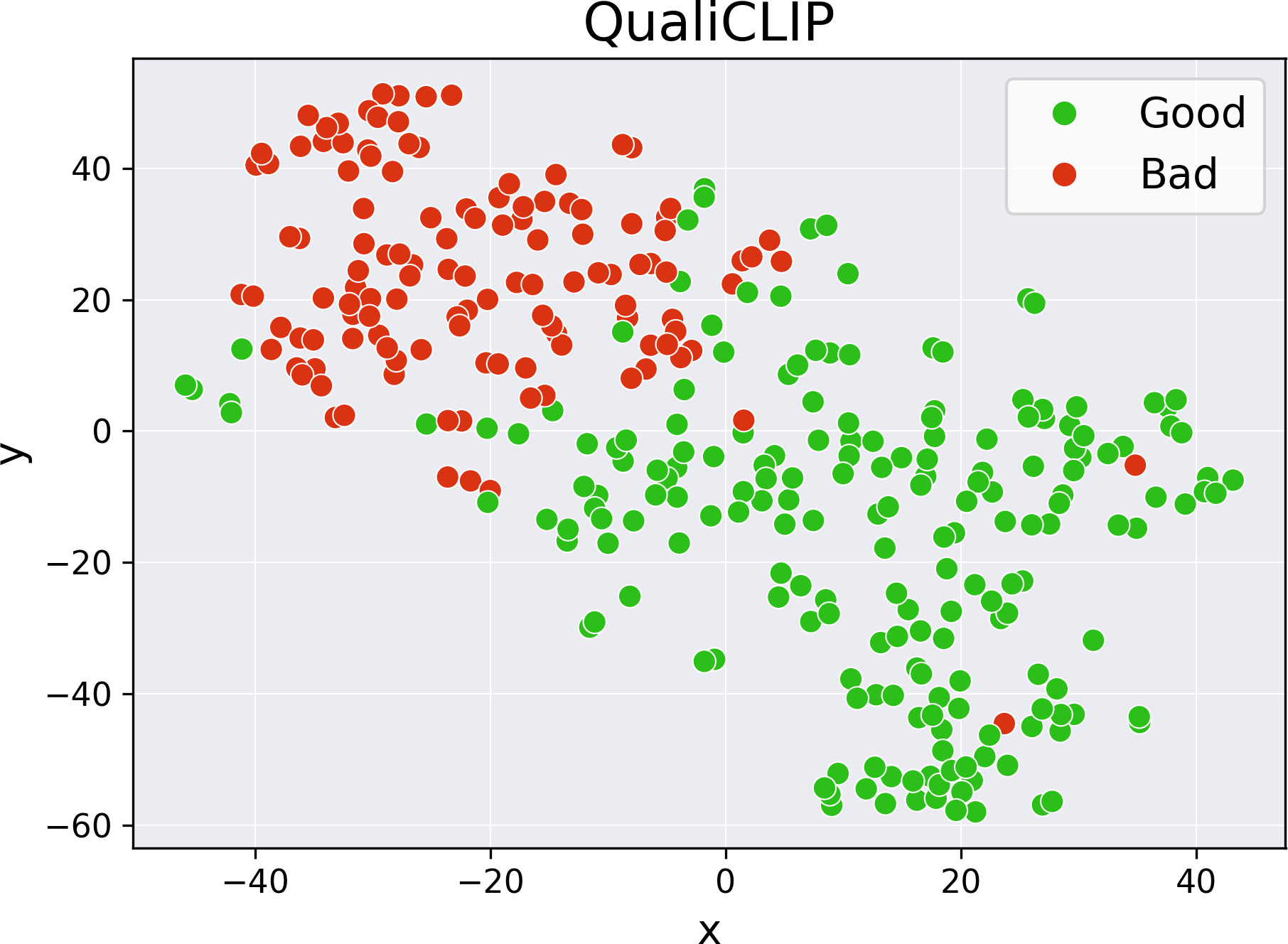}
        \caption{}
        \label{fig:tsne_qualiclip}
    \end{subfigure}
    \vspace{-5pt}
    \caption{Comparison of the t-SNE visualizations related to the image representations of the CLIVE dataset generated by CLIP-IQA (a) and \method (b), respectively. Good and bad points refer to images with a MOS greater than 75 and lower than 25, respectively.}
    \label{fig:tsne}
\end{figure}

We compare the image representations generated by \method and CLIP-IQA via a t-SNE \cite{van2008visualizing} visualization. Following \cite{srinath2024learning}, we consider images from the CLIVE dataset with very high or very low quality. In particular, we take into account images with a labeled MOS greater than 75 and lower than 25, respectively. \cref{fig:tsne} shows the results. We observe that the representations of high- and low-quality images obtained by the proposed approach (\cref{fig:tsne_qualiclip}) correspond to more easily separable clusters compared to those of CLIP-IQA (\cref{fig:tsne_clipiqa}), which are more intertwined. This result confirms that \method generates more accurate quality-aware representations.

\subsection{Supervised \method}
Although our approach is designed to remove the requirement for human annotations, it can easily be extended to leverage ground-truth labels. Similar to CLIP-IQA$^{+}$ \cite{wang2023exploring}, we exploit the MOS to fine-tune the antonym prompts with an MSE loss via standard backpropagation, while keeping the network weights fixed. We refer to this supervised opinion-aware variant of our approach as \methodplus. We train \methodplus on the training split of each of the testing datasets and report the performance in \cref{tab:supervised_qualiclip}. For a fair comparison, we only consider supervised baselines that do not fine-tune the network weights but instead train a smaller set of parameters, such as a linear regressor or the antonym prompts. We observe that \methodplus achieves competitive performance also in this evaluation setting, outperforming the baselines on most metrics. This outcome shows that the proposed method can also be applied in scenarios where human annotations are available.

\begin{table*}[t]
  \centering
  \setlength{\tabcolsep}{4.25pt}
  \resizebox{\linewidth}{!}{ 
  \begin{tabular}{lccccccccccccccccc} 
  \toprule
  \multicolumn{1}{c}{} & \multicolumn{1}{c}{} & \multicolumn{8}{c}{\textit{Authentic Datasets}} & \multicolumn{4}{c}{\textit{Image Restoration Datasets}} & \multicolumn{4}{c}{\textit{AIGC Datasets}} \\ 
  \cmidrule(lr){3-10} \cmidrule(lr){11-14} \cmidrule(lr){15-18}
  \multicolumn{1}{c}{} & \multicolumn{1}{c}{} & \multicolumn{2}{c}{\koniq} & \multicolumn{2}{c}{\clive} & \multicolumn{2}{c}{\flive} & \multicolumn{2}{c}{\spaq} & \multicolumn{2}{c}{\cviu} & \multicolumn{2}{c}{\pipal} & \multicolumn{2}{c}{\agiqaonek} & \multicolumn{2}{c}{\agiqathreek}\\
  \multicolumn{1}{l}{Method} & OU & \srcc & \plcc & \srcc & \plcc & \srcc & \plcc & \srcc & \plcc & \srcc & \plcc & \srcc & \plcc & \srcc & \plcc & \srcc & \plcc \\ 
  \midrule
  CONTRIQUE \cite{madhusudana2022image} & \hred{\xmark} & 0.874 & 0.882 & 0.806 & 0.838 & 0.596 & 0.648 & 0.910 & 0.912 & 0.912 & 0.916 & 0.581 & \underline{0.611} & 0.799 & \underline{0.855} & 0.817 & 0.879 \\
  Re-IQA \cite{saha2023re} & \hred{\xmark} & \underline{0.883} & 0.887 & 0.783 & 0.797 & \textbf{0.623} & \textbf{0.685} & \underline{0.909} & \underline{0.912} & 0.887 & 0.895 & 0.568 & 0.587 & 0.783 & 0.840 & 0.811 & 0.874 \\
  ARNIQA \cite{agnolucci2024arniqa} & \hred{\xmark} & 0.869 & 0.883 & 0.797 & 0.823 & 0.595 & \underline{0.670} & 0.904 & 0.909 & 0.919 & \underline{0.926} & \textbf{0.634} & \textbf{0.666} & 0.768 & 0.849 & 0.803 & 0.881 \\
  CLIP-IQA$^{+}$ \cite{wang2023exploring} & \hred{\xmark} & 0.873 & \underline{0.890} & \underline{0.815} & \underline{0.849} & 0.602 & 0.662 & 0.901 & 0.904 & \underline{0.925} & 0.918 & 0.552 & 0.558 & \underline{0.817} & 0.854 & \underline{0.844} & \underline{0.894} \\ 
  GRepQ \cite{srinath2024learning} & \hred{\xmark} & 0.882 & 0.883 & 0.793 & 0.813 & 0.576 & 0.611 & 0.902 & 0.903 & 0.882 & 0.857 & 0.554 & 0.568 & 0.740 & 0.797 & 0.807 & 0.858 \\ \midrule
  \rowcolor{LightCyan} \textbf{\methodplus} & \hred{\xmark} & \textbf{0.889} & \textbf{0.898} & \textbf{0.821} & \textbf{0.867} & \underline{0.618} & 0.665 & \textbf{0.911} & \textbf{0.914} & \textbf{0.936} & \textbf{0.936} & \underline{0.586} & 0.590 & \textbf{0.826} & \textbf{0.871} & \textbf{0.860} & \textbf{0.903} \\
  \bottomrule
  \end{tabular}}
  \vspace{-5pt}
  \caption{Comparison between \methodplus and supervised approaches. OU stands for Opinion-Unaware. Best and second-best scores methods are highlighted in bold and underlined, respectively.}
  \label{tab:supervised_qualiclip}
\end{table*}

\subsection{Inference Time}
As detailed in Sec. \textcolor{iccvblue}{3.3}, we do not finetune CLIP's text encoder. Consequently, we need to compute the text features of the antonym prompts only once, and we can employ them both for training and inference. Therefore, at inference time, \method has the same computational cost as an image-encoder-only model with a ResNet50 backbone.

To validate this, we compare the average inference time of our model with that of the supervised baselines on the \koniq dataset \cite{hosu2020koniq10k}, which comprises 10K $1024\times768$px images. We conduct the experiments on an NVIDIA RTX 2080Ti GPU. We report the results in \cref{tab:inference_time}. As expected, \method has a similar inference time to models based solely on ResNet50, such as CLIP-IQA$^{+}$, LIQE, CONTRIQUE, and ARNIQA. Note that CONTRIQUE and ARNIQA employ the full- and half-scale versions of the input image to compute the final quality score, and thus are more computationally expensive. Moreover, our model is faster than methods based on two encoders, namely GRepQ and Re-IQA. This outcome demonstrates the efficiency and applicability of \method in real-world scenarios.

\begin{table}
    \centering
    \Large
    \resizebox{0.8\columnwidth}{!}{ 
    \begin{tabular}{lc}
    \toprule
    Method & Inference Time (ms) \\ \midrule
    HyperIQA \cite{su2020blindly} & 16 $\pm$ 3 \\
    DBCNN \cite{zhang2018blind} & 25 $\pm$ 3 \\
    TReS \cite{golestaneh2022no} & 103 $\pm$ 5 \\
    LIQE \cite{zhang2023blind} & 13 $\pm$ 2 \\
    QCN \cite{shin2024blind} & 101 $\pm$ 6 \\
    CONTRIQUE \cite{madhusudana2022image} & 15 $\pm$ 3 \\
    Re-IQA \cite{saha2023re} & 42 $\pm$ 4 \\
    ARNIQA \cite{agnolucci2024arniqa} & 15 $\pm$ 3 \\
    CLIP-IQA+ \cite{wang2023exploring} & 13 $\pm$ 2 \\
    GRepQ \cite{srinath2024learning} & 27 $\pm$ 3 \\ \midrule
    \rowcolor{LightCyan} \textbf{\method} & 13 $\pm$ 2 \\ \bottomrule
    \end{tabular}
    }
    \caption{Comparison of the average inference time between \method and supervised methods on the \koniq \cite{hosu2020koniq10k} dataset.}
    \label{tab:inference_time}
\end{table}

\section{Implementation Details}

\tit{Training details}
We rely on a ResNet50 \cite{he2016deep} as the backbone for CLIP's image encoder. Similar to \cite{wang2023exploring}, we remove the positional embedding from the encoder to make our model capable of taking images of any resolution as input. The dimension $d$ of CLIP's embedding space is 1024. Differently from \cite{srinath2024learning}, we do not train a projector head on top of CLIP's image encoder. We keep CLIP's text encoder frozen.
We train our model for 10 epochs. We employ an AdamW \cite{loshchilov2017decoupled} optimizer with a weight decay and a learning rate of $1e\!-\!2$ and $1e\!-\!9$, respectively. During training, we use a patch size of 224 and a batch size of 16. We set the margins $\marginone$ in Eq. \textcolor{iccvblue}{1} and $\margintwo$ in Eq. \textcolor{iccvblue}{2} and \textcolor{iccvblue}{3} to $2.5e\!-\!3$ and $6.75e\!-\!2$, respectively. The loss weights $\lambdaone$, $\lambdatwo$ and $\lambdathree$ in Eq. \textcolor{iccvblue}{4} are all equal to 1. We set the temperature hyperparameter $\tau$ in Eq. \textcolor{iccvblue}{5} to 2. At inference time, our model takes the whole image as input.

\tit{Prompts} \label{sec:prompts}
Following \cite{wu2023exploring, wu2023towardsrobust}, we employ multiple pairs of antonym prompts during training and inference. In particular, we use: 1) \dquote{Good/Bad photo}; 2) \dquote{Good/Bad picture}; 3) \dquote{High-resolution/Low-resolution image}; 4) \dquote{High-quality/Low-quality image}; 5) \dquote{Sharp/Blurry image}; 6) \dquote{Sharp/Blurry edges}; 7) \dquote{Noise-free/Noisy image}. We average the similarities between the images and the pairs of prompts.

\begin{figure*}
    \centering
    \setlength{\tabcolsep}{0.9pt}
    \resizebox{\textwidth}{!}{ 
\begin{tabular}{C{6em}ccccc}
         & Level 1 & Level 2 & Level 3 & Level 4 & Level 5 \\
 Brighten & \includegraphics[width=\gridimagewidth,valign=m]{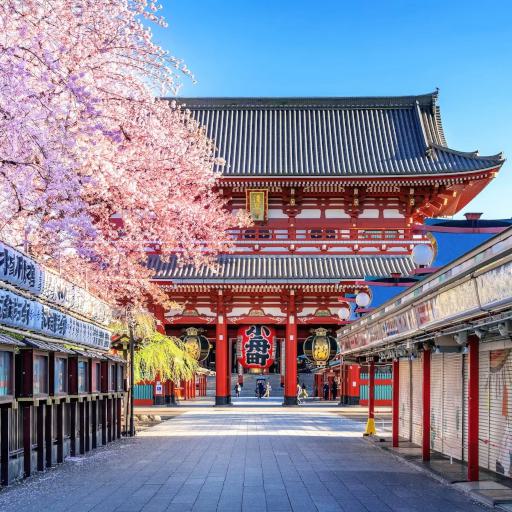} & \includegraphics[width=\gridimagewidth,valign=m]{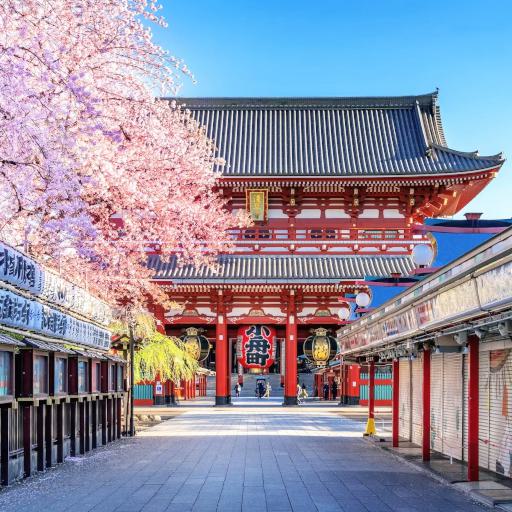} & \includegraphics[width=\gridimagewidth,valign=m]{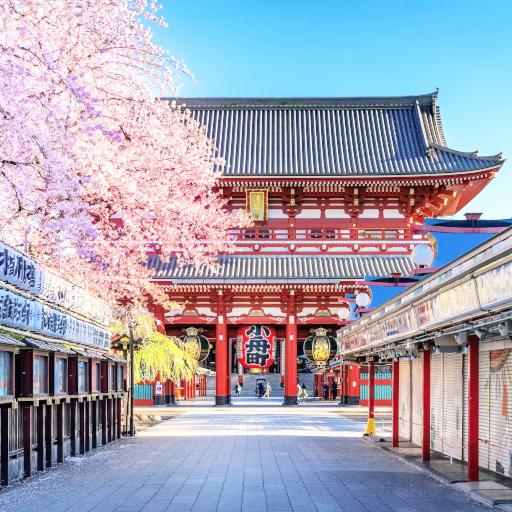} & \includegraphics[width=\gridimagewidth,valign=m]{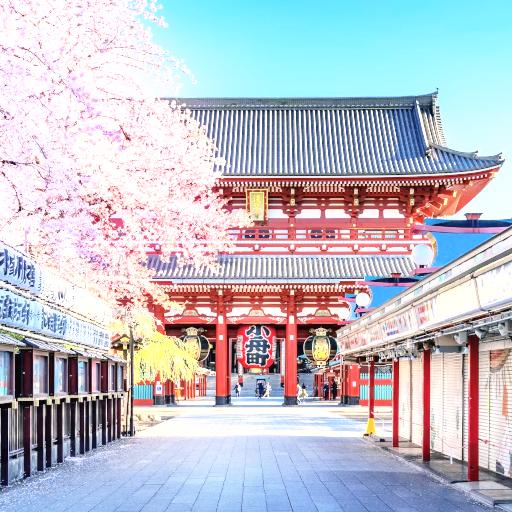} & \includegraphics[width=\gridimagewidth,valign=m]{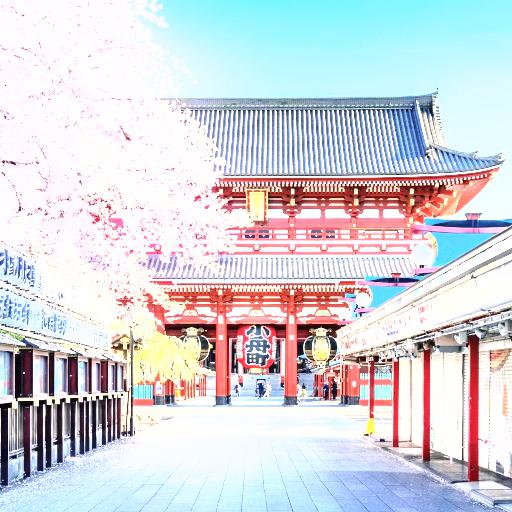} \\ [29pt]
 Darken & \includegraphics[width=\gridimagewidth,valign=m]{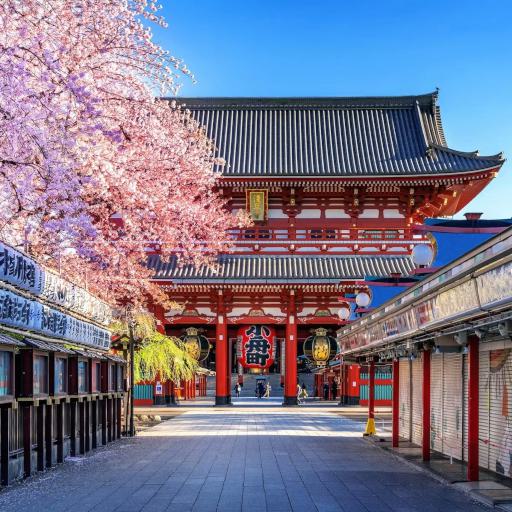} & \includegraphics[width=\gridimagewidth,valign=m]{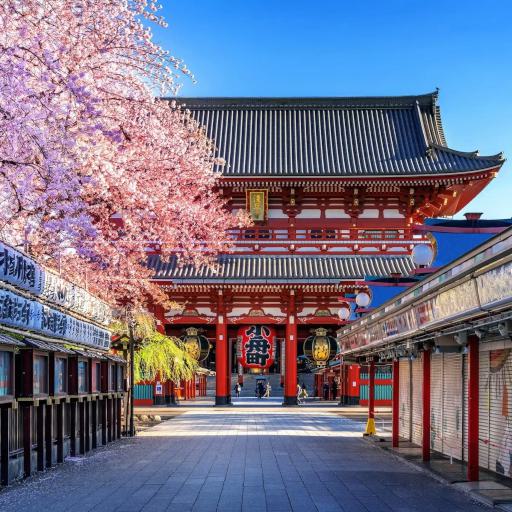} & \includegraphics[width=\gridimagewidth,valign=m]{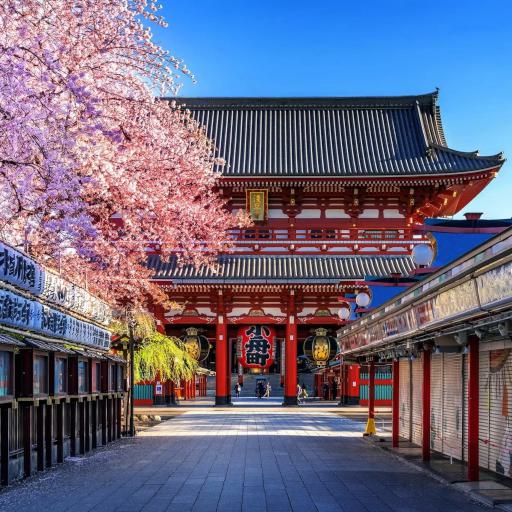} & \includegraphics[width=\gridimagewidth,valign=m]{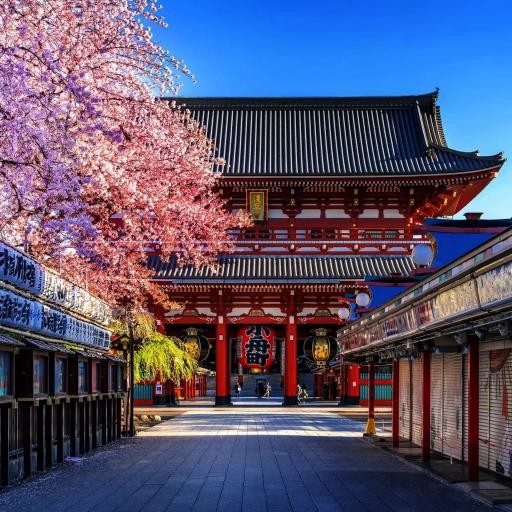} & \includegraphics[width=\gridimagewidth,valign=m]{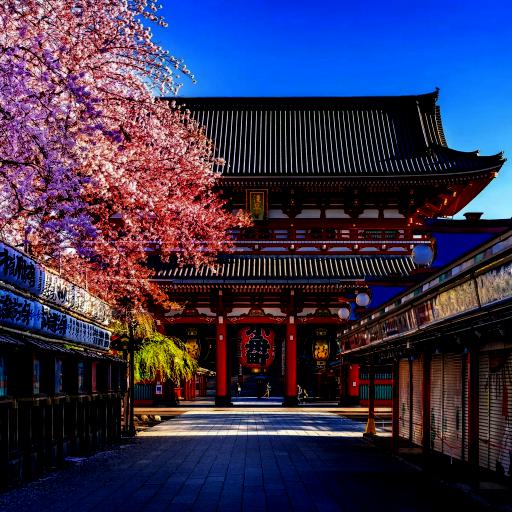} \\ [29pt]
 Mean shift & \includegraphics[width=\gridimagewidth,valign=m]{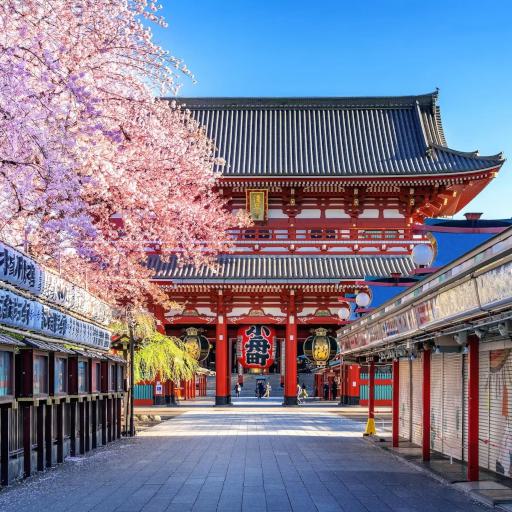} & 
 \includegraphics[width=\gridimagewidth,valign=m]{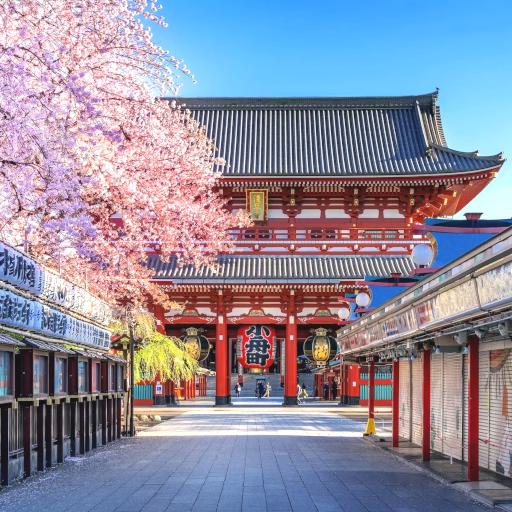} & \includegraphics[width=\gridimagewidth,valign=m]{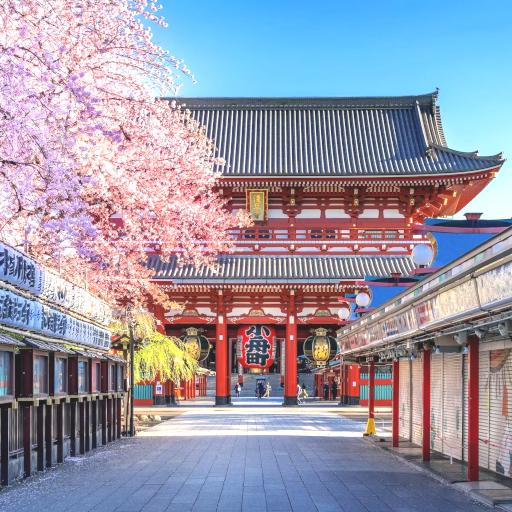} & \includegraphics[width=\gridimagewidth,valign=m]{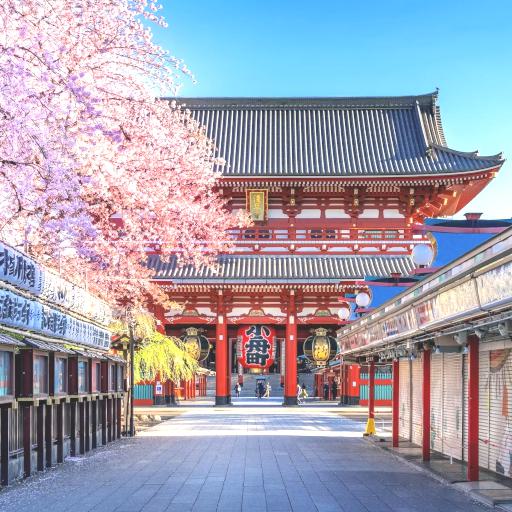} & \includegraphics[width=\gridimagewidth,valign=m]{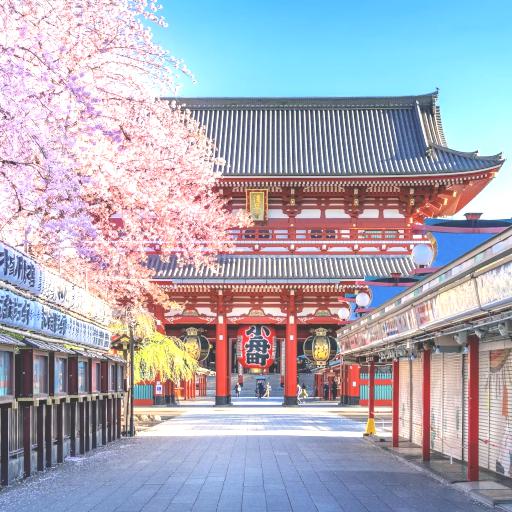} \\
 
\end{tabular}
}
\caption{Visualization of the distortion types belonging to the \textit{Brightness change} group for increasing intensity levels.}
\label{fig:brightness_change}
\end{figure*}

\begin{figure*}
    \centering
    \setlength{\tabcolsep}{0.9pt}
    \resizebox{\textwidth}{!}{ 
\begin{tabular}{C{6em}ccccc}
         & Level 1 & Level 2 & Level 3 & Level 4 & Level 5 \\
 Gaussian blur & \includegraphics[width=\gridimagewidth,valign=m]{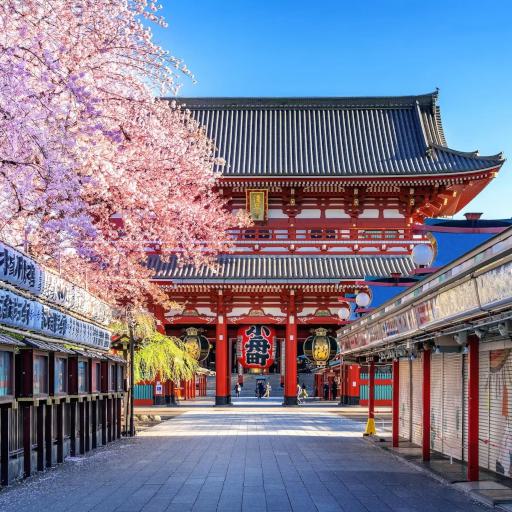} & \includegraphics[width=\gridimagewidth,valign=m]{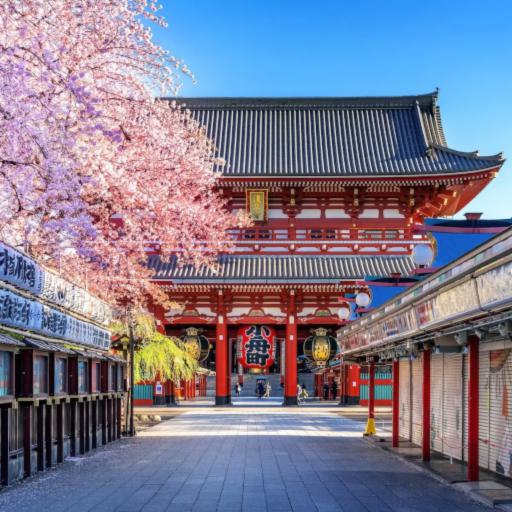} & \includegraphics[width=\gridimagewidth,valign=m]{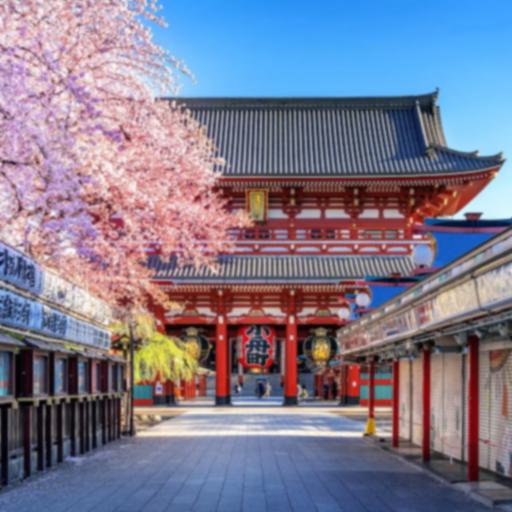} & \includegraphics[width=\gridimagewidth,valign=m]{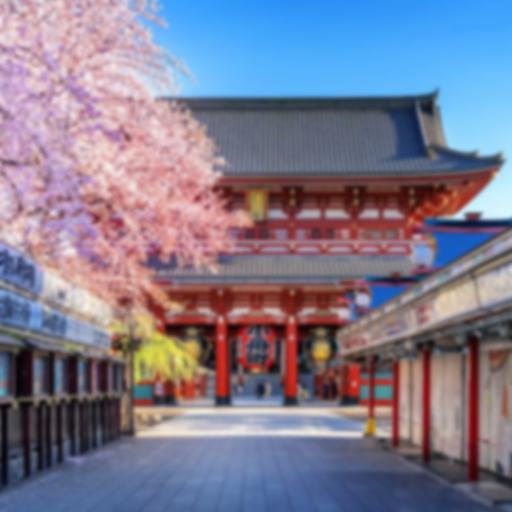} & \includegraphics[width=\gridimagewidth,valign=m]{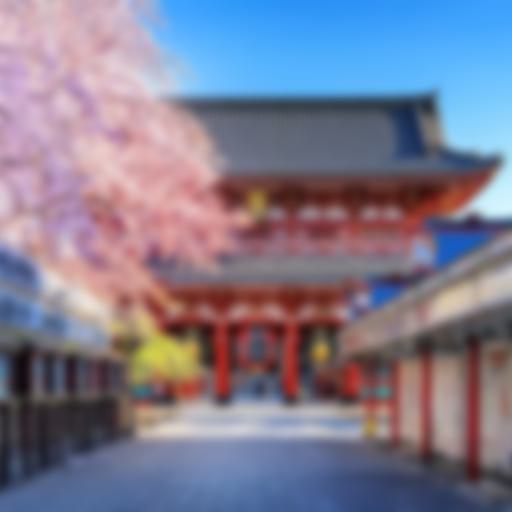} \\ [29pt]
 Lens blur & \includegraphics[width=\gridimagewidth,valign=m]{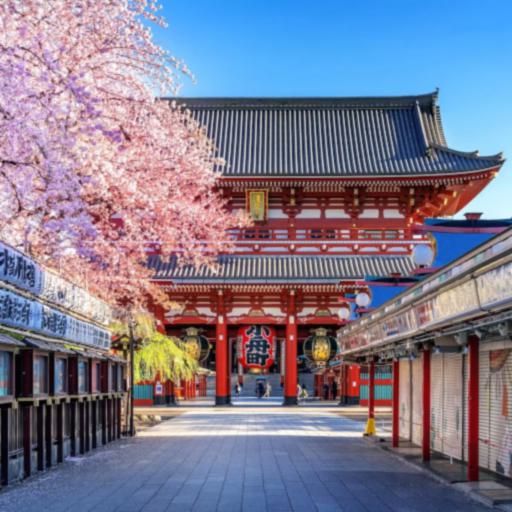} & \includegraphics[width=\gridimagewidth,valign=m]{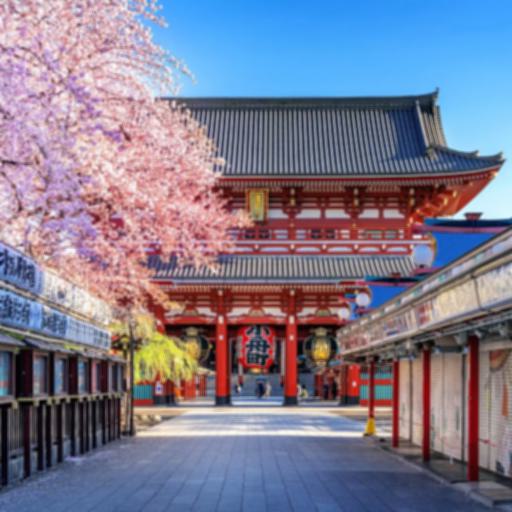} & \includegraphics[width=\gridimagewidth,valign=m]{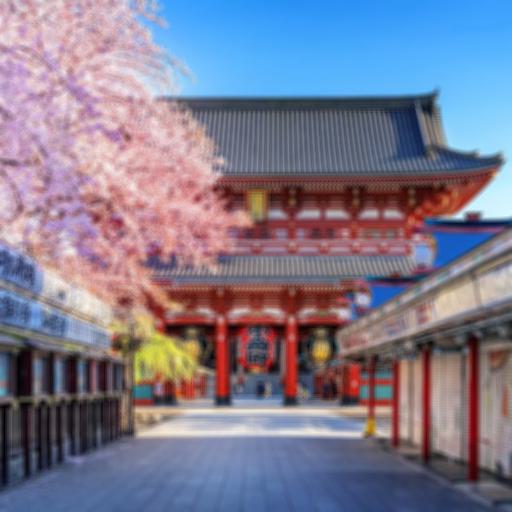} & \includegraphics[width=\gridimagewidth,valign=m]{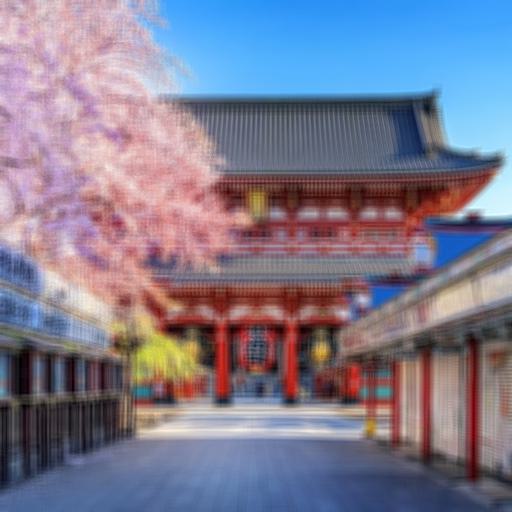} & \includegraphics[width=\gridimagewidth,valign=m]{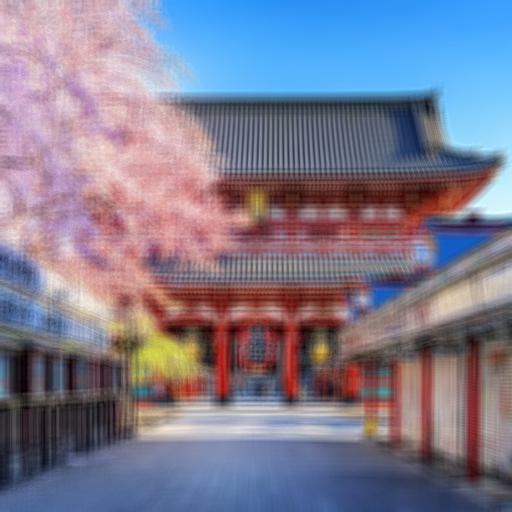} \\ [29pt]
 Motion blur & \includegraphics[width=\gridimagewidth,valign=m]{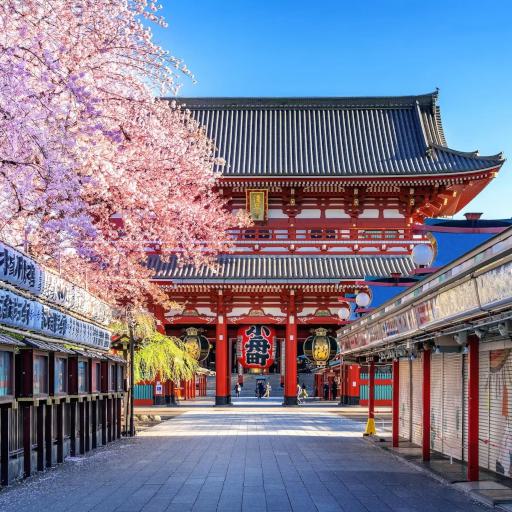} & \includegraphics[width=\gridimagewidth,valign=m]{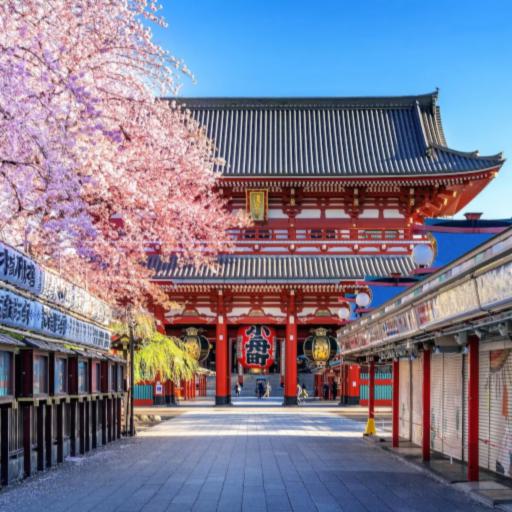} & \includegraphics[width=\gridimagewidth,valign=m]{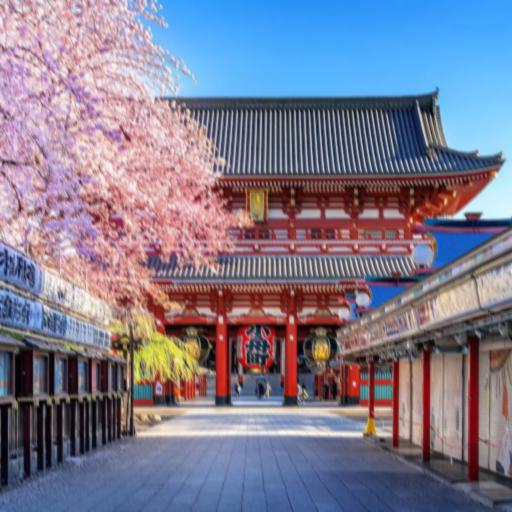} & \includegraphics[width=\gridimagewidth,valign=m]{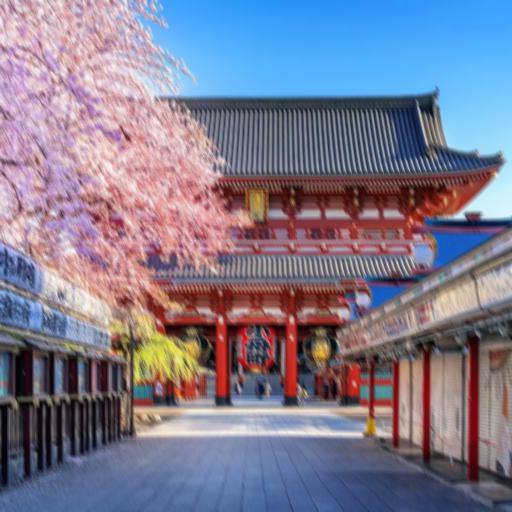} & \includegraphics[width=\gridimagewidth,valign=m]{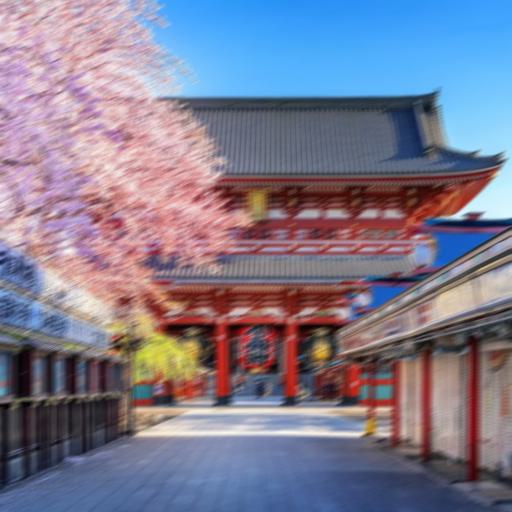} \\ 
\end{tabular}
}
\caption{Visualization of the distortion types belonging to the \textit{Blur} group for increasing intensity levels.}
\label{fig:blur}
\end{figure*}

\begin{figure*}
    \centering
    \setlength{\tabcolsep}{0.9pt}
    \resizebox{\textwidth}{!}{ 
\begin{tabular}{C{6em}ccccc}
         & Level 1 & Level 2 & Level 3 & Level 4 & Level 5 \\
 Jitter & \includegraphics[width=\gridimagewidth,valign=m]{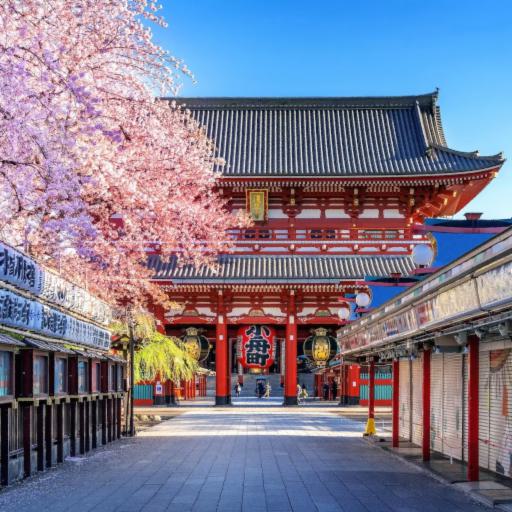} & \includegraphics[width=\gridimagewidth,valign=m]{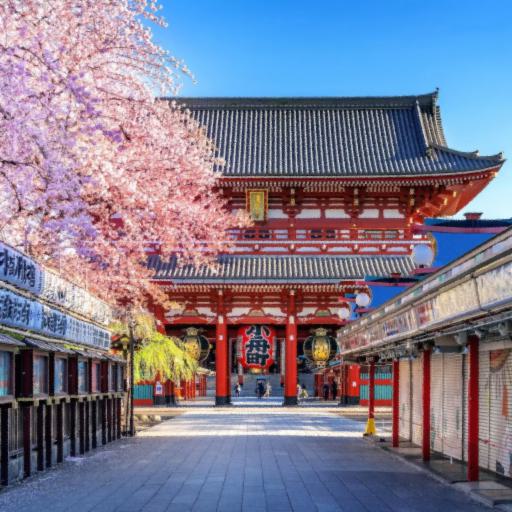} & \includegraphics[width=\gridimagewidth,valign=m]{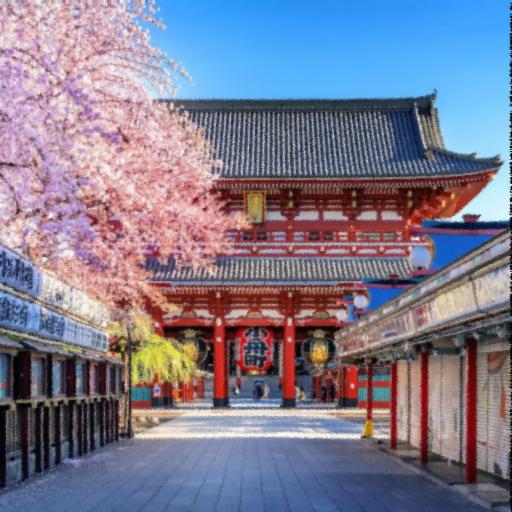} & \includegraphics[width=\gridimagewidth,valign=m]{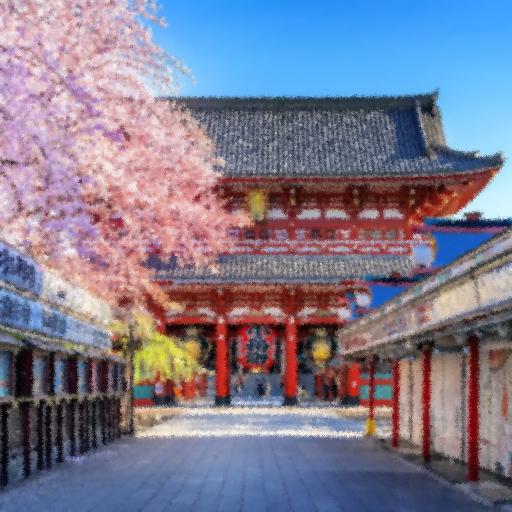} & \includegraphics[width=\gridimagewidth,valign=m]{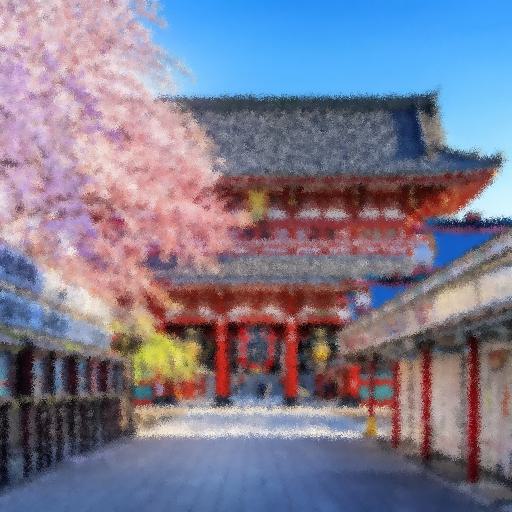} \\ [29pt]
 Non-eccentricity patch & \includegraphics[width=\gridimagewidth,valign=m]{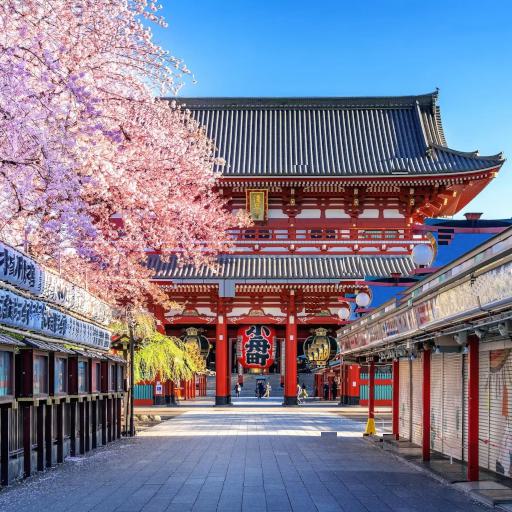} & \includegraphics[width=\gridimagewidth,valign=m]{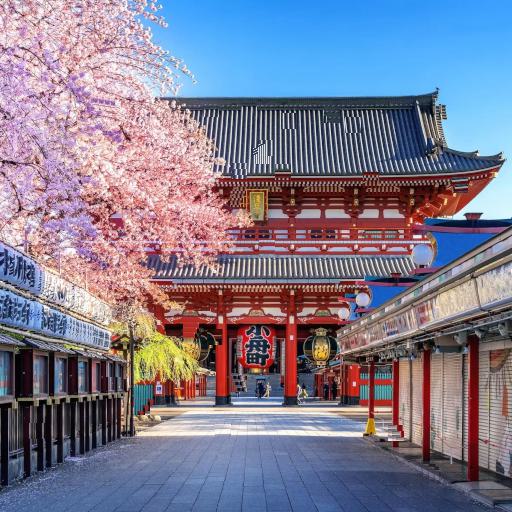} & \includegraphics[width=\gridimagewidth,valign=m]{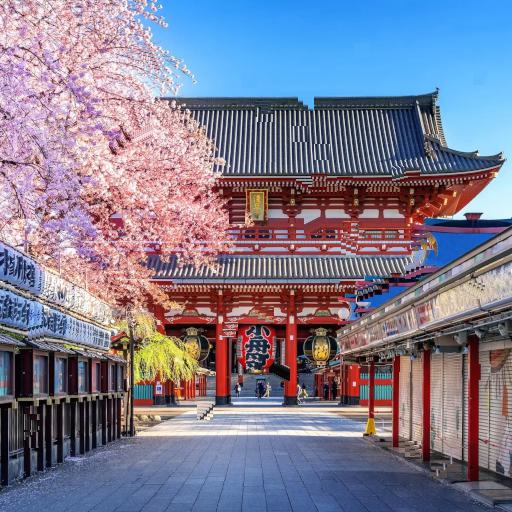} & \includegraphics[width=\gridimagewidth,valign=m]{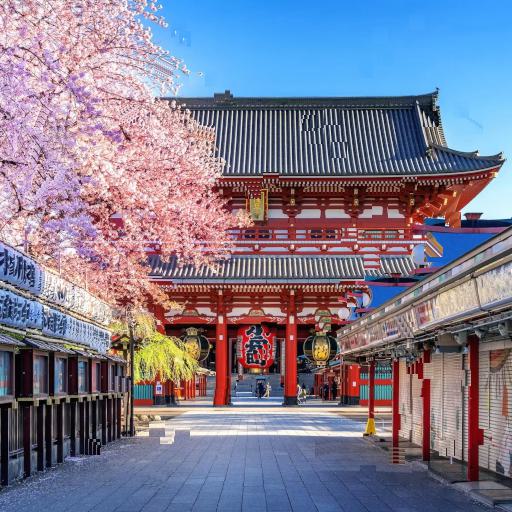} & \includegraphics[width=\gridimagewidth,valign=m]{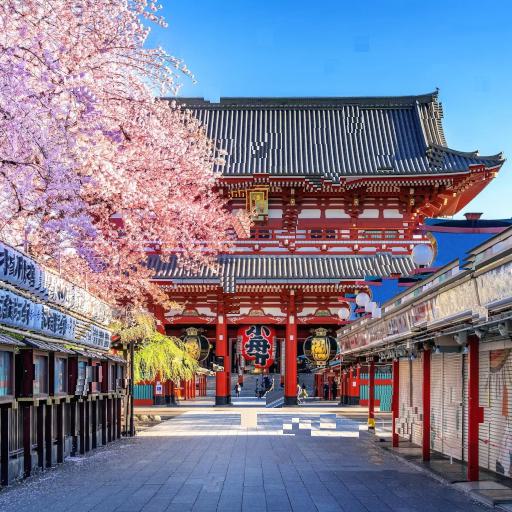} \\ [29pt]
 Pixelate & \includegraphics[width=\gridimagewidth,valign=m]{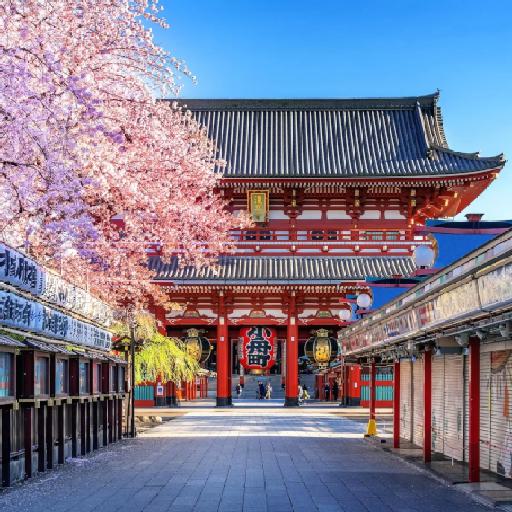} & \includegraphics[width=\gridimagewidth,valign=m]{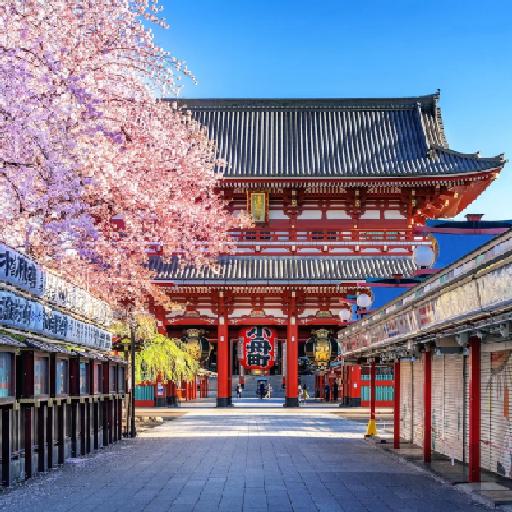} & \includegraphics[width=\gridimagewidth,valign=m]{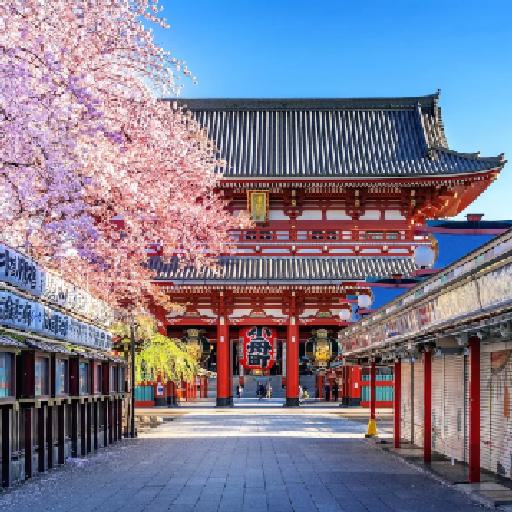} & \includegraphics[width=\gridimagewidth,valign=m]{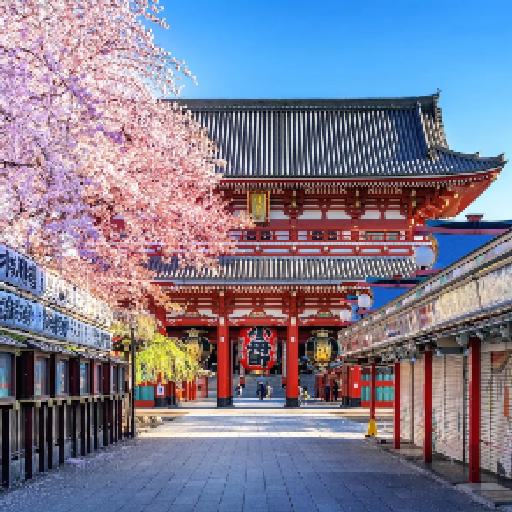} & \includegraphics[width=\gridimagewidth,valign=m]{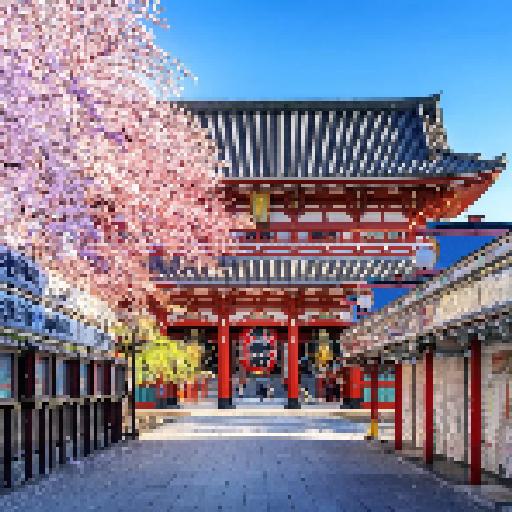} \\ [29pt]
 Quantization & \includegraphics[width=\gridimagewidth,valign=m]{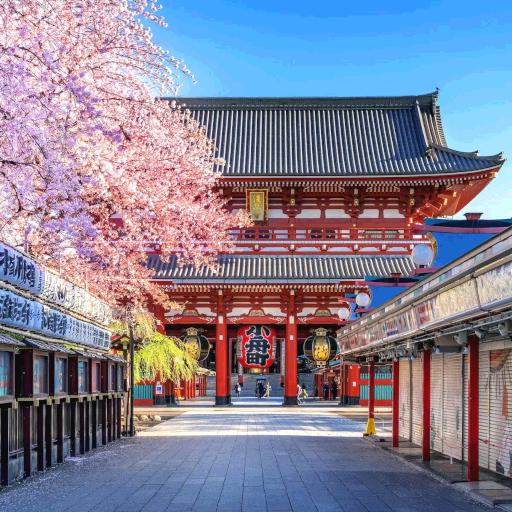} & \includegraphics[width=\gridimagewidth,valign=m]{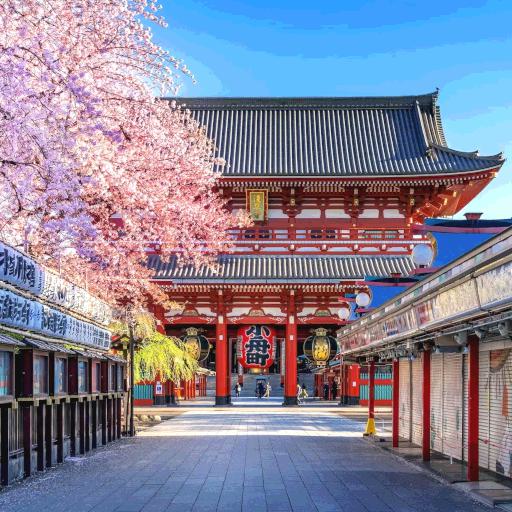} & \includegraphics[width=\gridimagewidth,valign=m]{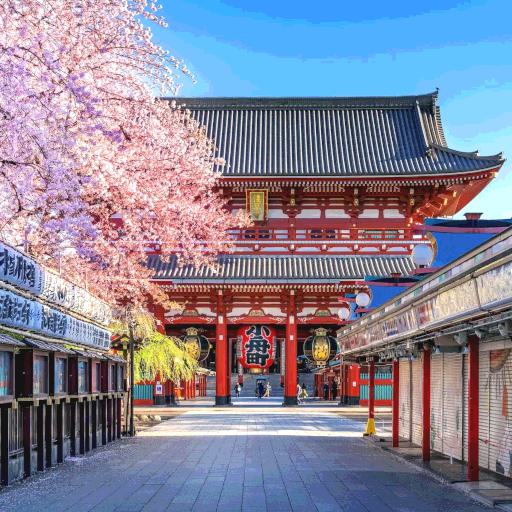} & \includegraphics[width=\gridimagewidth,valign=m]{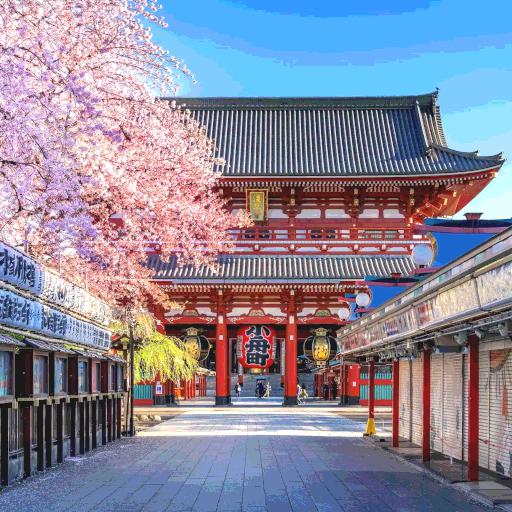} & \includegraphics[width=\gridimagewidth,valign=m]{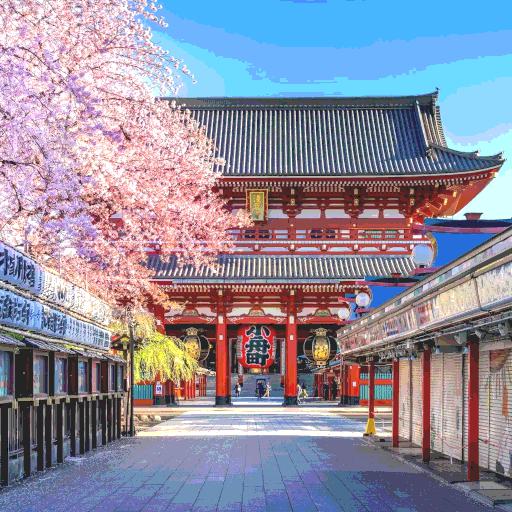} \\ [29pt]
 Color block & \includegraphics[width=\gridimagewidth,valign=m]{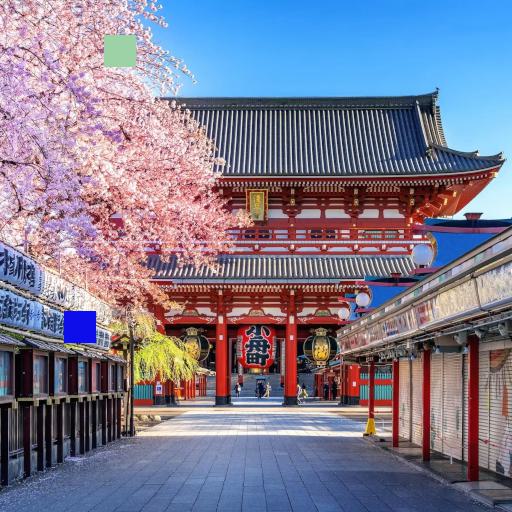} & \includegraphics[width=\gridimagewidth,valign=m]{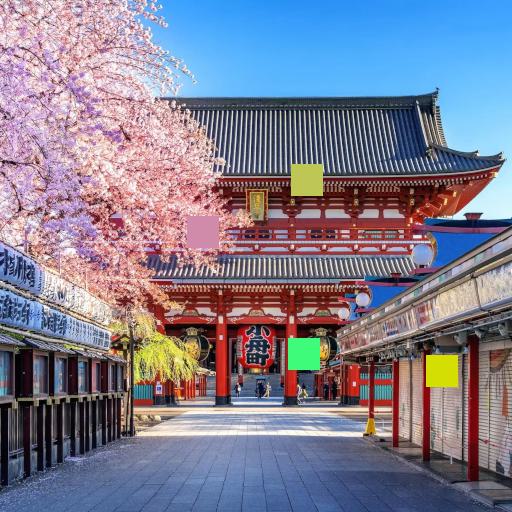} & \includegraphics[width=\gridimagewidth,valign=m]{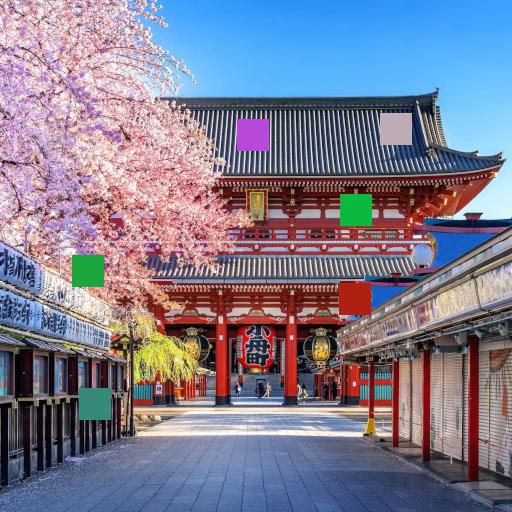} & \includegraphics[width=\gridimagewidth,valign=m]{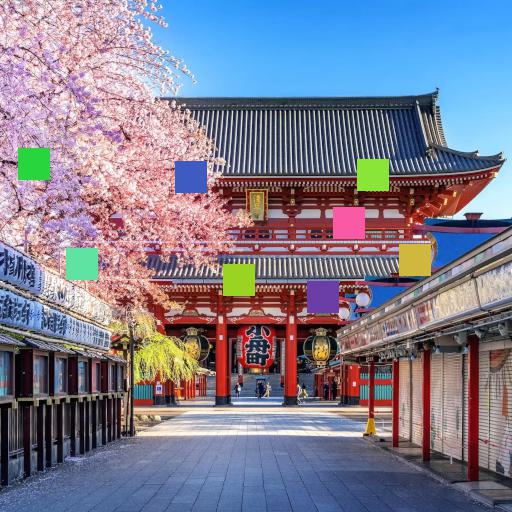} & \includegraphics[width=\gridimagewidth,valign=m]{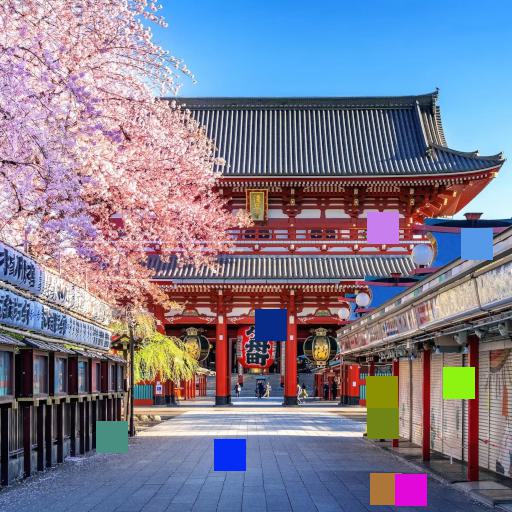} \\
\end{tabular}
}
\caption{Visualization of the distortion types belonging to the \textit{Spatial distortions} group for increasing intensity levels.}
\label{fig:spatial_distortions}
\end{figure*}

\begin{figure*}
    \centering
    \setlength{\tabcolsep}{0.9pt}
    \resizebox{\textwidth}{!}{ 
\begin{tabular}{C{6em}ccccc}
         & Level 1 & Level 2 & Level 3 & Level 4 & Level 5 \\
 White noise & \includegraphics[width=\gridimagewidth,valign=m]{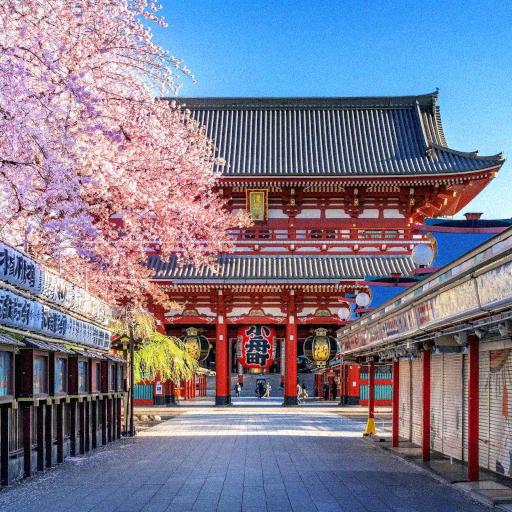} & \includegraphics[width=\gridimagewidth,valign=m]{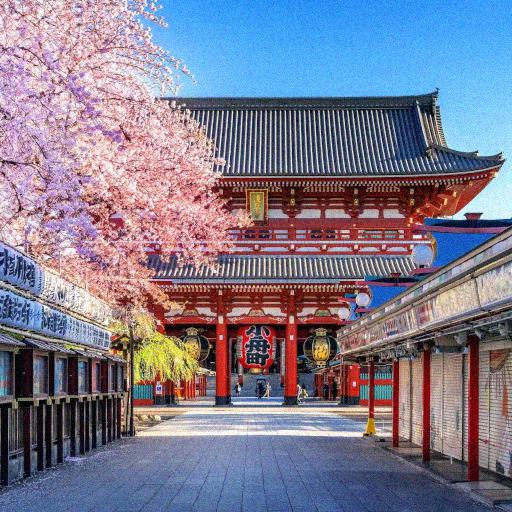} & \includegraphics[width=\gridimagewidth,valign=m]{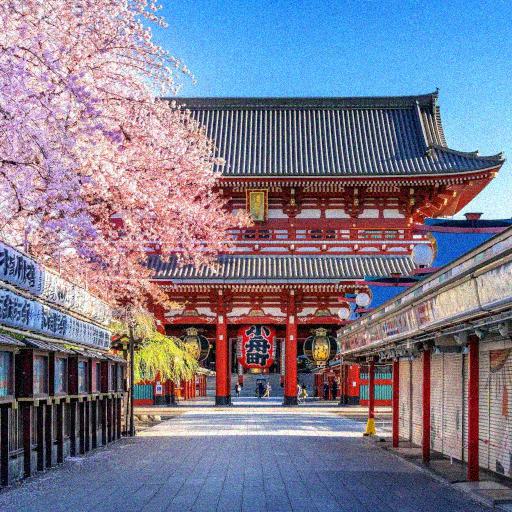} & \includegraphics[width=\gridimagewidth,valign=m]{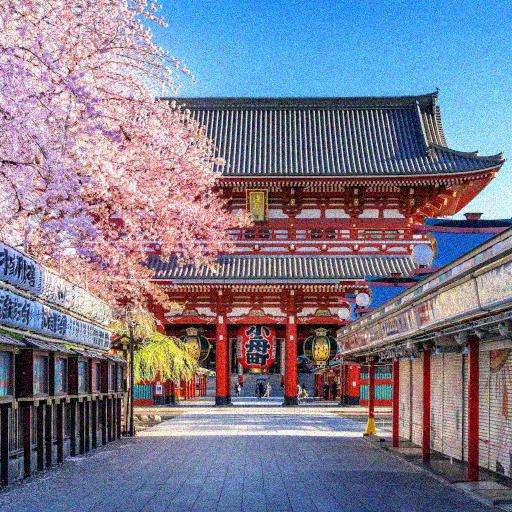} & \includegraphics[width=\gridimagewidth,valign=m]{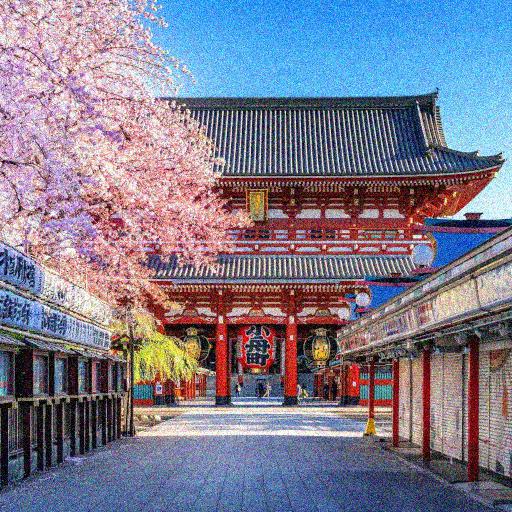} \\ [29pt]
 White noise \newline color \newline component & \includegraphics[width=\gridimagewidth,valign=m]{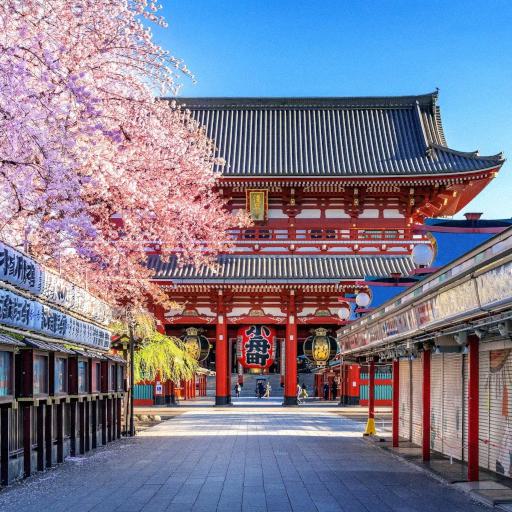} & \includegraphics[width=\gridimagewidth,valign=m]{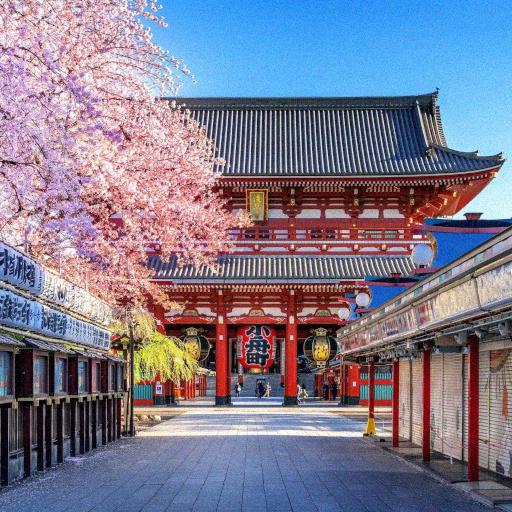} & \includegraphics[width=\gridimagewidth,valign=m]{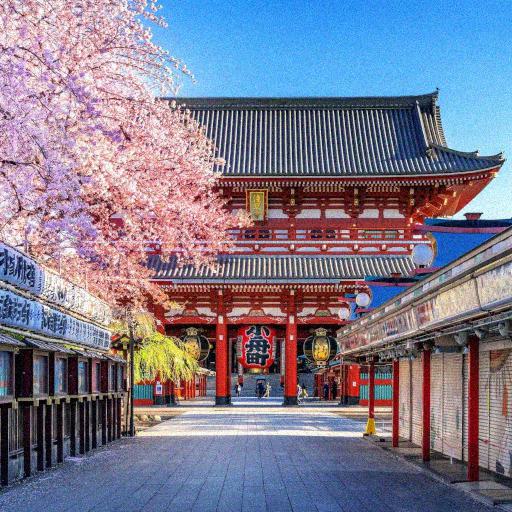} & \includegraphics[width=\gridimagewidth,valign=m]{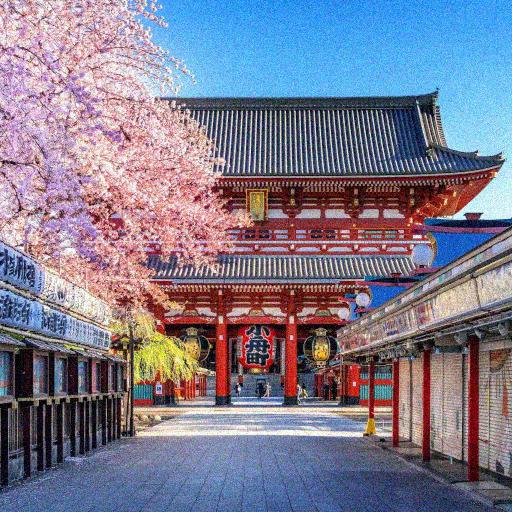} & \includegraphics[width=\gridimagewidth,valign=m]{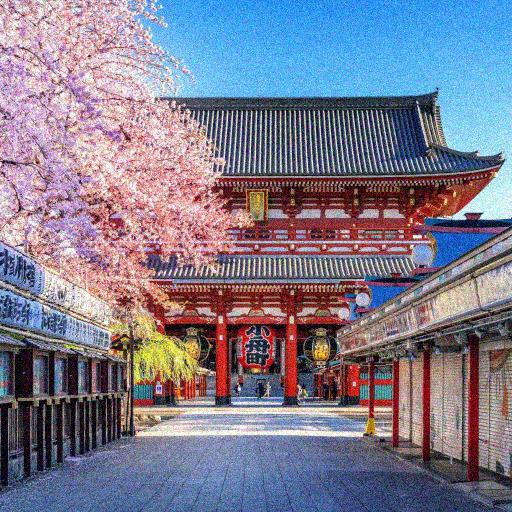} \\ [29pt]
 Impulse \newline noise & \includegraphics[width=\gridimagewidth,valign=m]{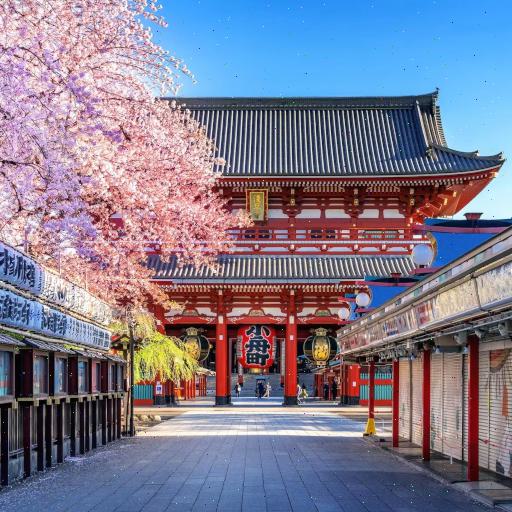} & \includegraphics[width=\gridimagewidth,valign=m]{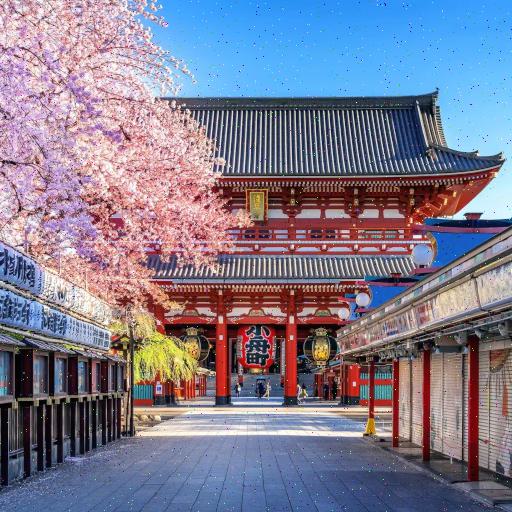} & \includegraphics[width=\gridimagewidth,valign=m]{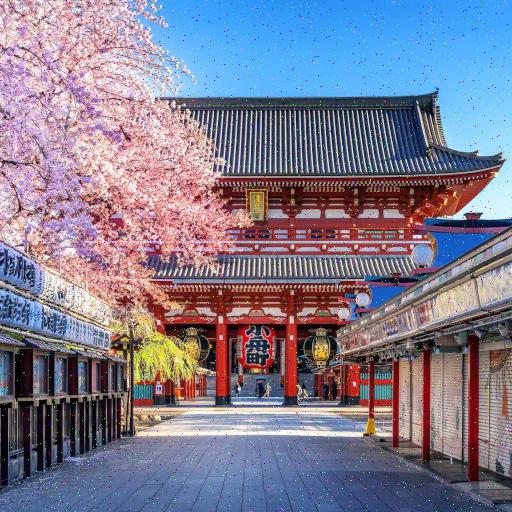} & \includegraphics[width=\gridimagewidth,valign=m]{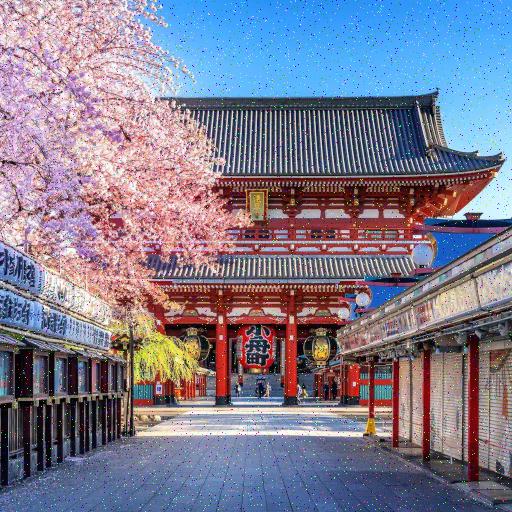} & \includegraphics[width=\gridimagewidth,valign=m]{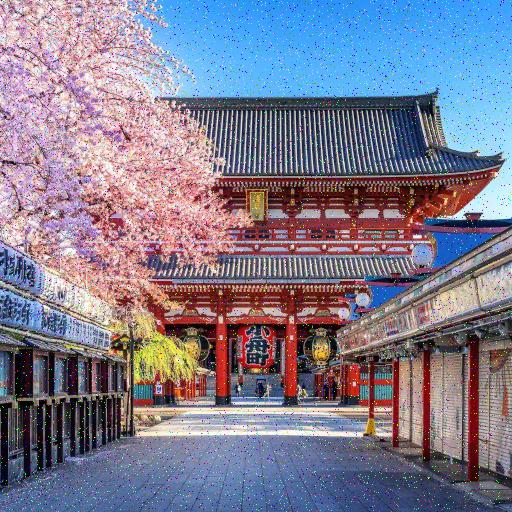} \\ [29pt]
 Multiplicative noise & \includegraphics[width=\gridimagewidth,valign=m]{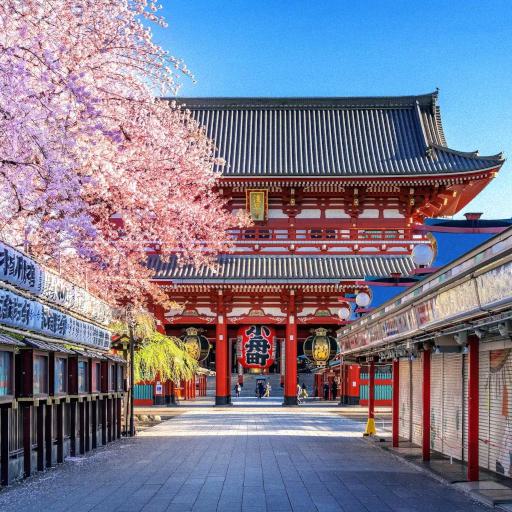} & \includegraphics[width=\gridimagewidth,valign=m]{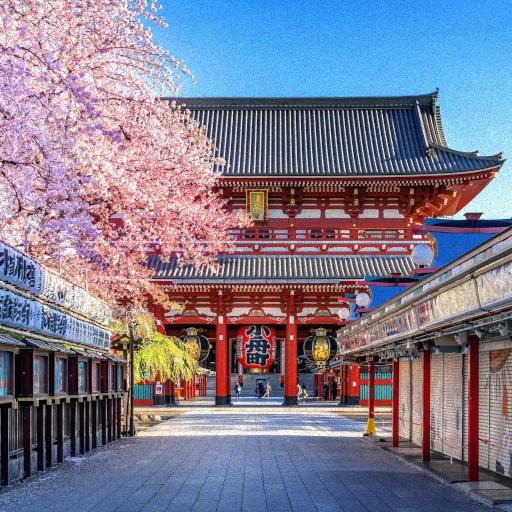} & \includegraphics[width=\gridimagewidth,valign=m]{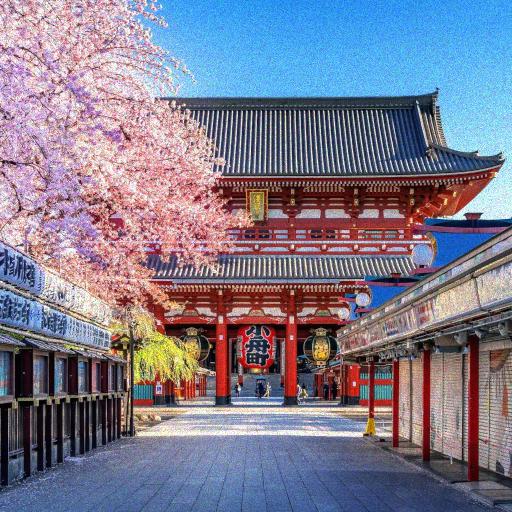} & \includegraphics[width=\gridimagewidth,valign=m]{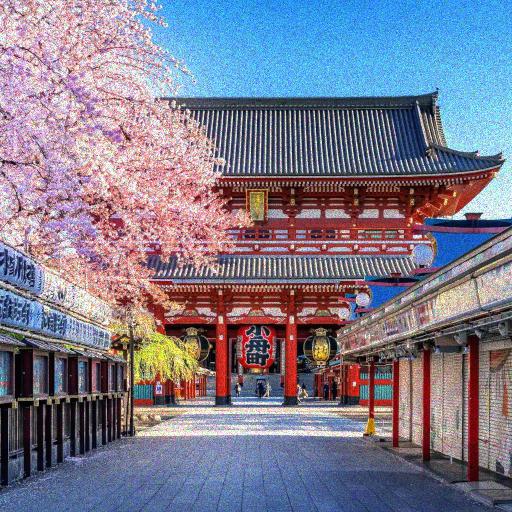} & \includegraphics[width=\gridimagewidth,valign=m]{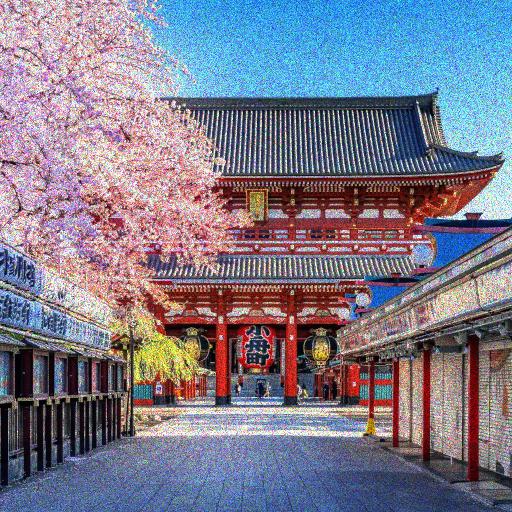} \\
\end{tabular}
}
\caption{Visualization of the distortion types belonging to the \textit{Noise} group for increasing intensity levels.}
\label{fig:noise}
\end{figure*}

\begin{figure*}
    \centering
    \setlength{\tabcolsep}{0.9pt}
    \resizebox{\textwidth}{!}{ 
\begin{tabular}{C{6em}ccccc}
         & Level 1 & Level 2 & Level 3 & Level 4 & Level 5 \\
 Color diffusion & \includegraphics[width=\gridimagewidth,valign=m]{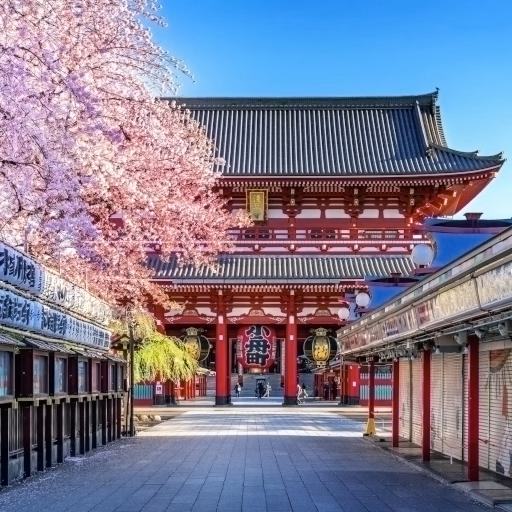} & \includegraphics[width=\gridimagewidth,valign=m]{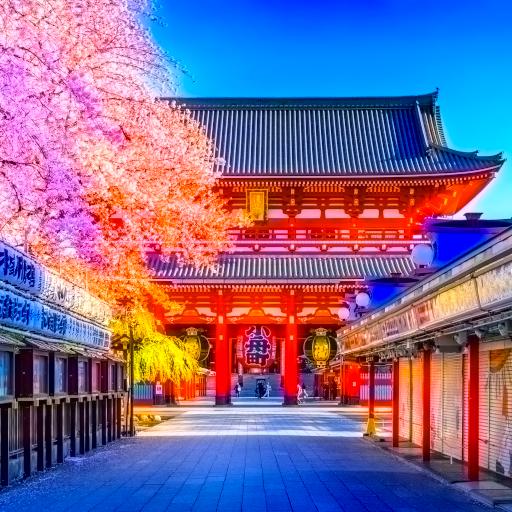} & \includegraphics[width=\gridimagewidth,valign=m]{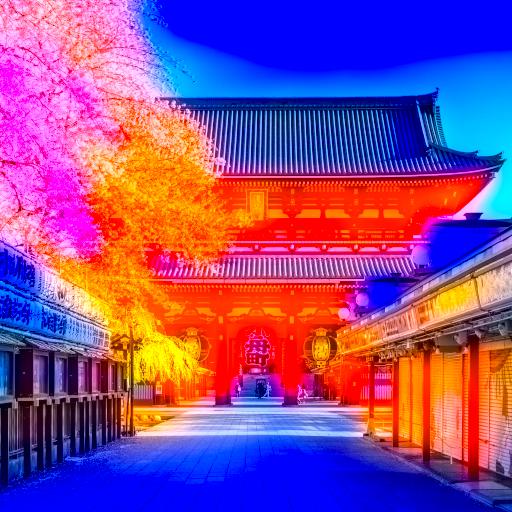} & \includegraphics[width=\gridimagewidth,valign=m]{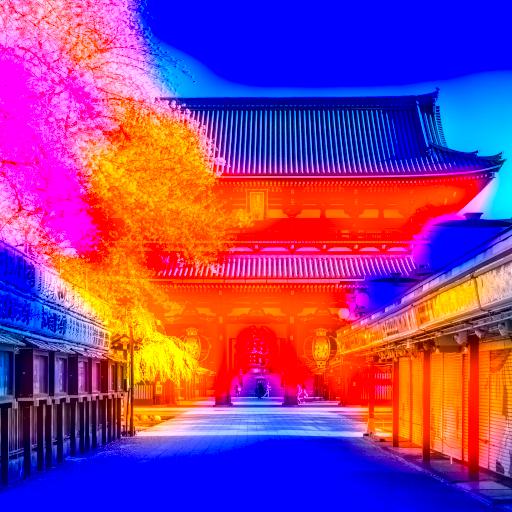} & \includegraphics[width=\gridimagewidth,valign=m]{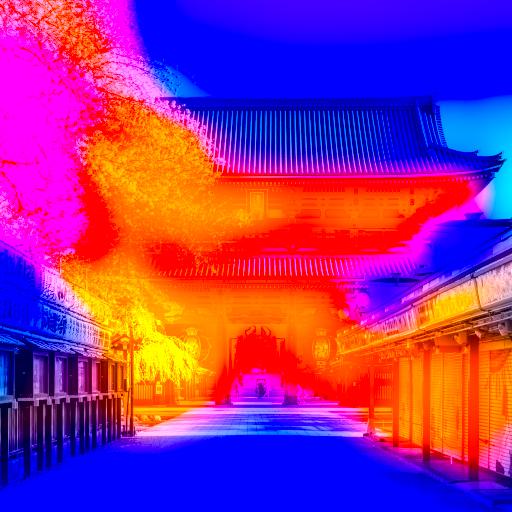} \\ [29pt]
 Color shift & \includegraphics[width=\gridimagewidth,valign=m]{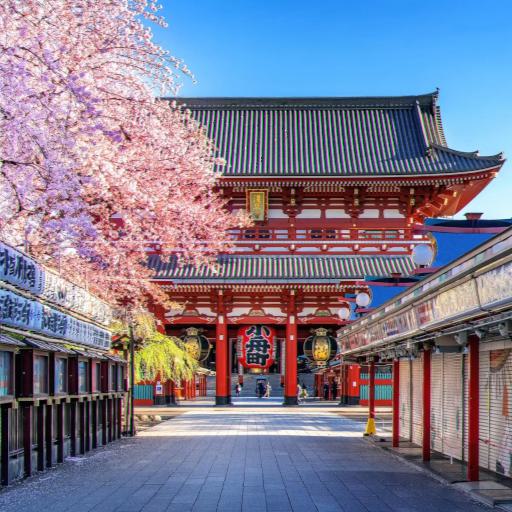} & \includegraphics[width=\gridimagewidth,valign=m]{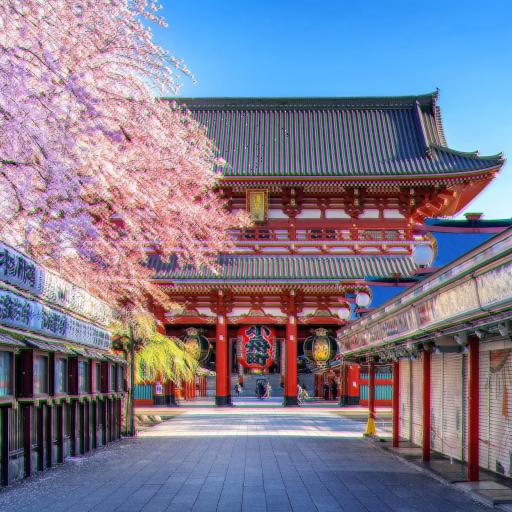} & \includegraphics[width=\gridimagewidth,valign=m]{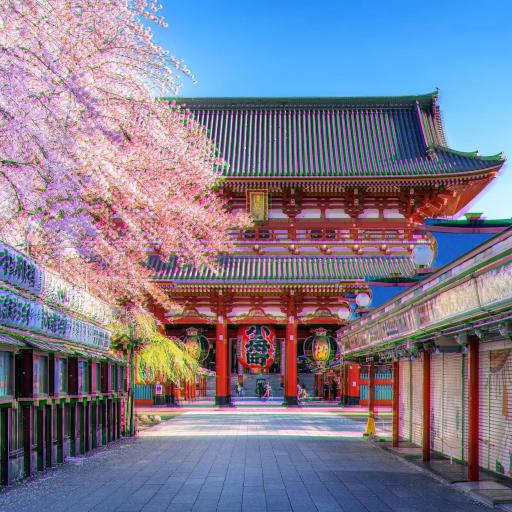} & \includegraphics[width=\gridimagewidth,valign=m]{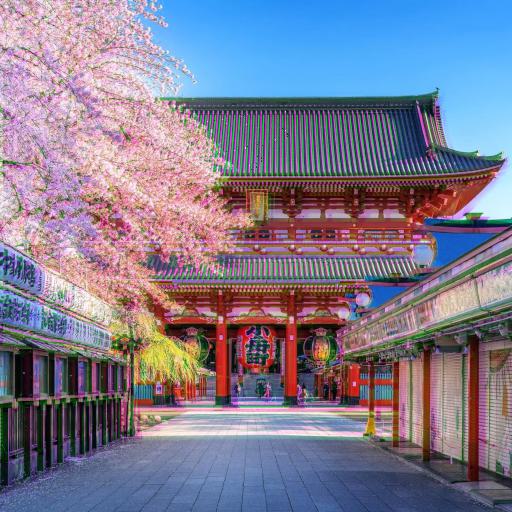} & \includegraphics[width=\gridimagewidth,valign=m]{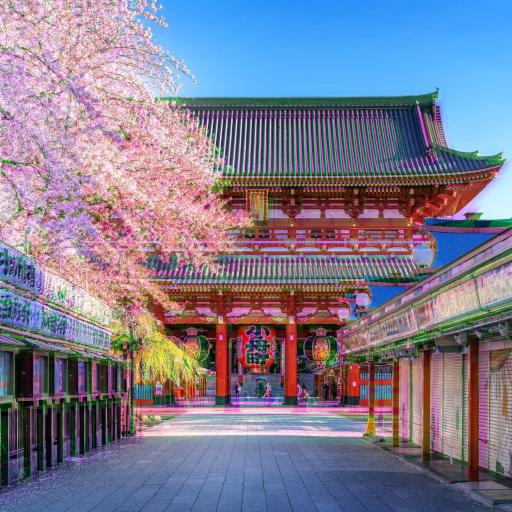} \\ [29pt]
 Color saturation 1 & \includegraphics[width=\gridimagewidth,valign=m]{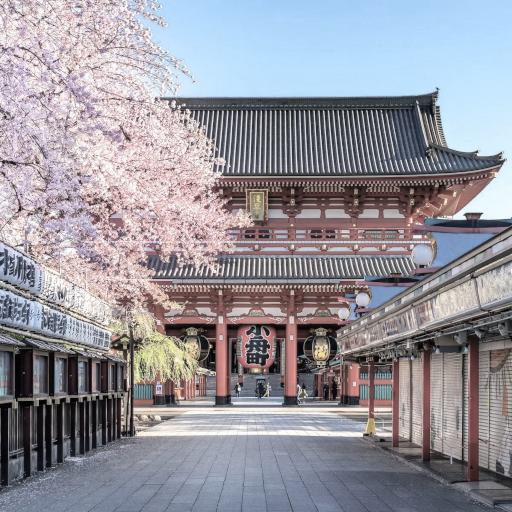} & \includegraphics[width=\gridimagewidth,valign=m]{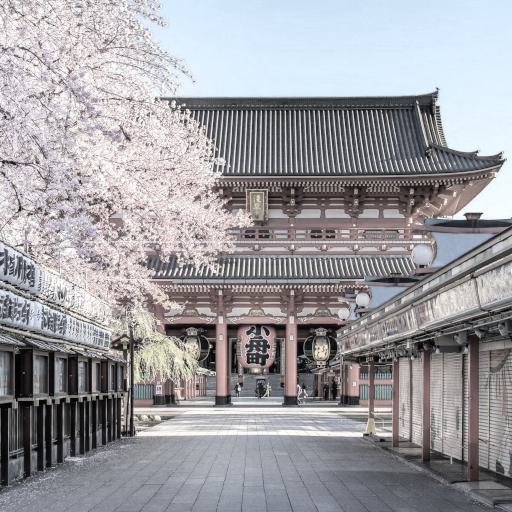} & \includegraphics[width=\gridimagewidth,valign=m]{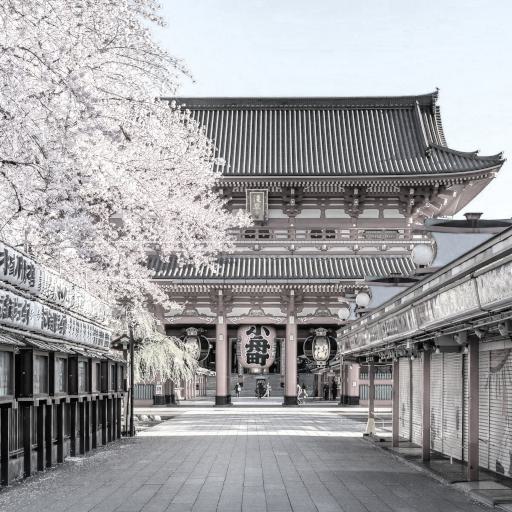} & \includegraphics[width=\gridimagewidth,valign=m]{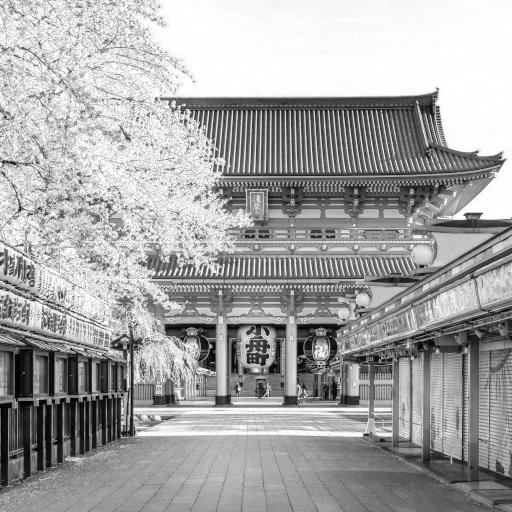} & \includegraphics[width=\gridimagewidth,valign=m]{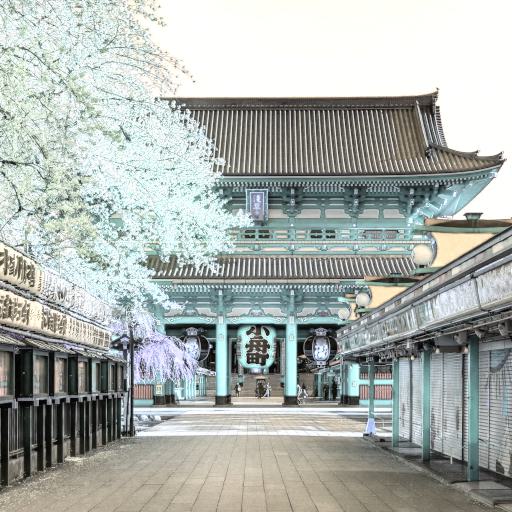} \\ [29pt]
 Color saturation 2 & \includegraphics[width=\gridimagewidth,valign=m]{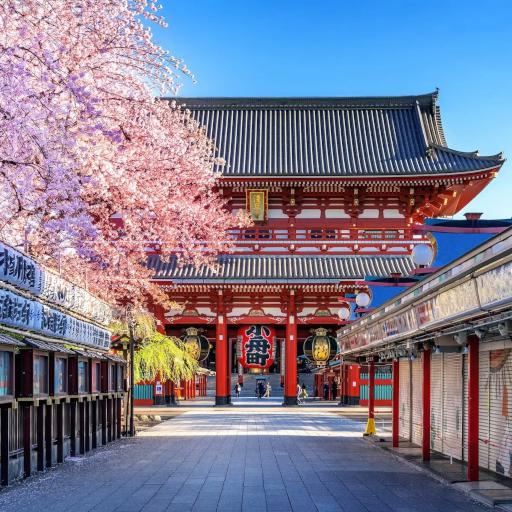} & \includegraphics[width=\gridimagewidth,valign=m]{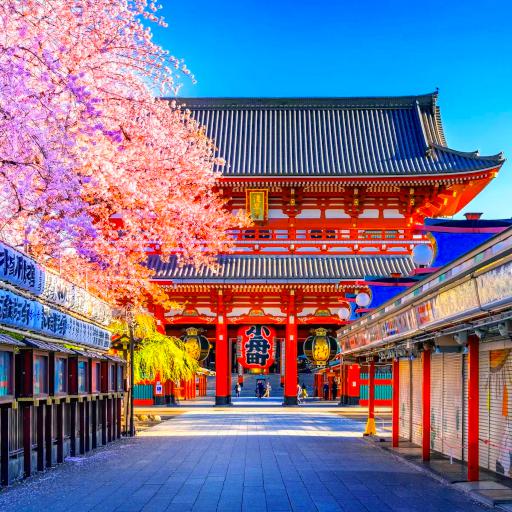} & \includegraphics[width=\gridimagewidth,valign=m]{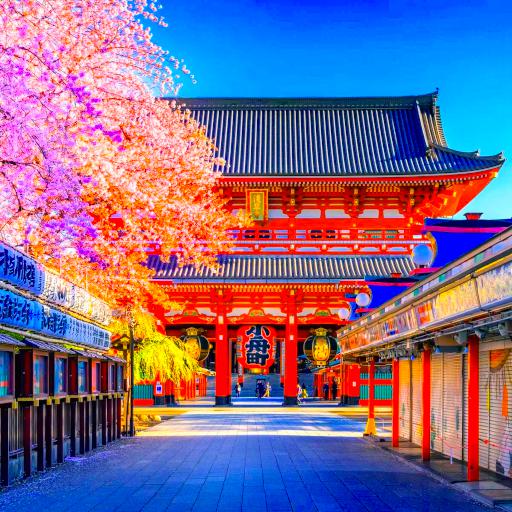} & \includegraphics[width=\gridimagewidth,valign=m]{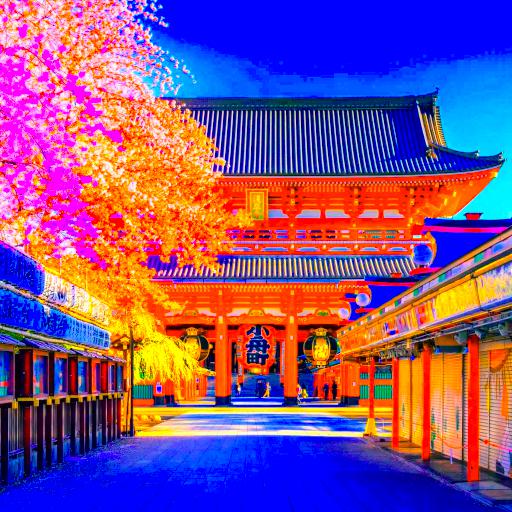} & \includegraphics[width=\gridimagewidth,valign=m]{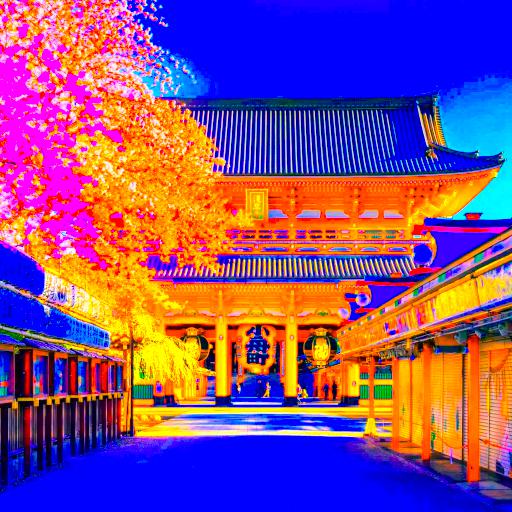} \\
\end{tabular}
}
\caption{Visualization of the distortion types belonging to the \textit{Color distortions} group for increasing intensity levels.}
\label{fig:color_distortions}
\end{figure*}

\begin{figure*}
    \centering
    \setlength{\tabcolsep}{0.9pt}
    \resizebox{\textwidth}{!}{ 
\begin{tabular}{C{6em}ccccc}
         & Level 1 & Level 2 & Level 3 & Level 4 & Level 5 \\
 JPEG2000 & \includegraphics[width=\gridimagewidth,valign=m]{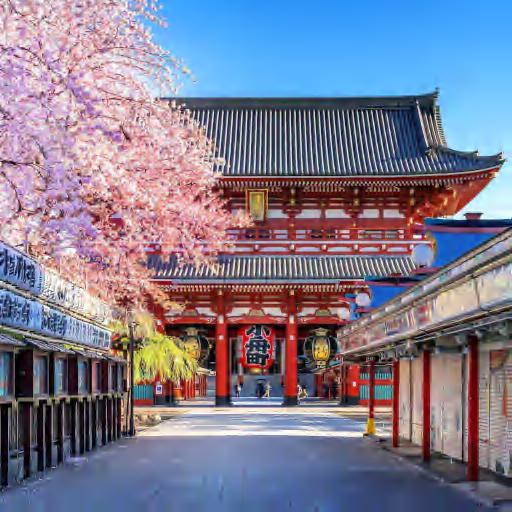} & \includegraphics[width=\gridimagewidth,valign=m]{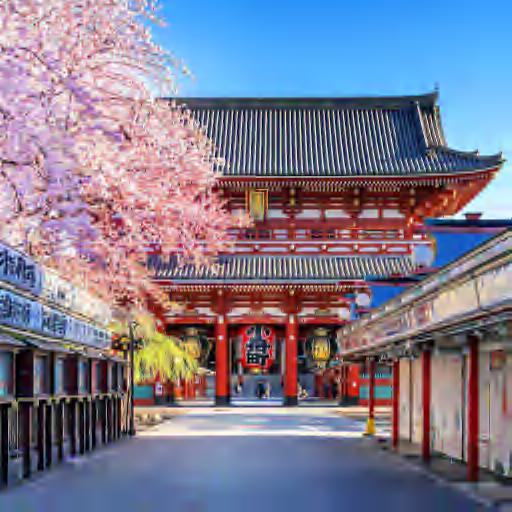} & \includegraphics[width=\gridimagewidth,valign=m]{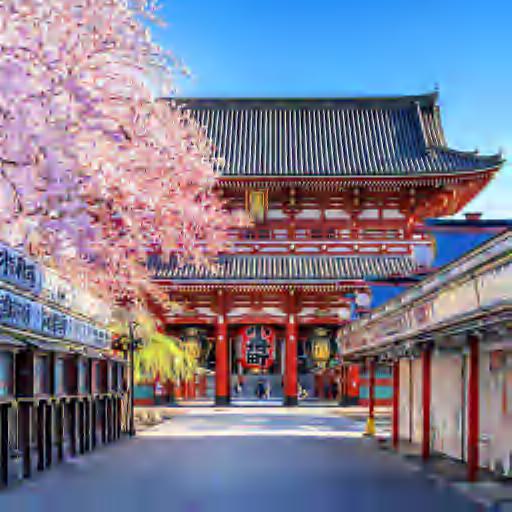} & \includegraphics[width=\gridimagewidth,valign=m]{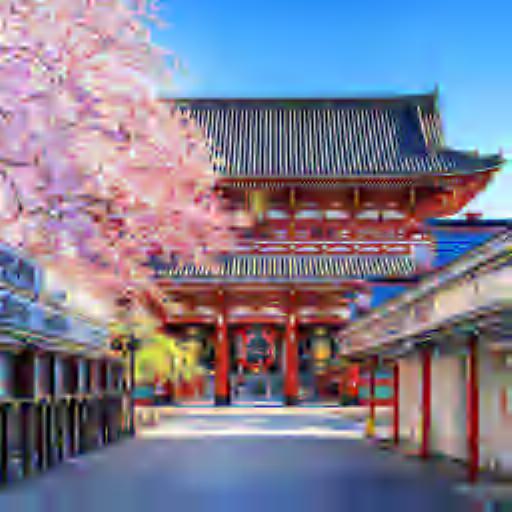} & \includegraphics[width=\gridimagewidth,valign=m]{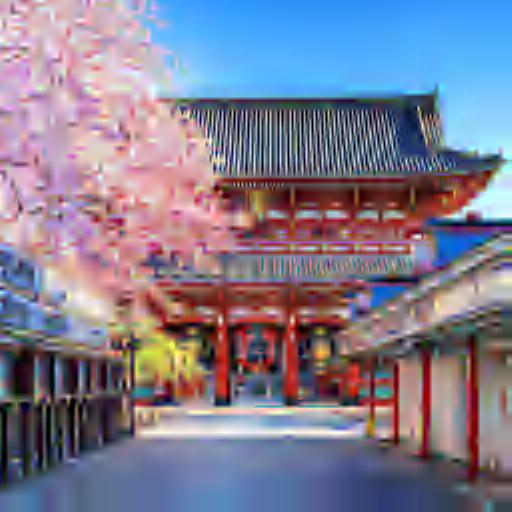} \\ [29pt]
 JPEG & \includegraphics[width=\gridimagewidth,valign=m]{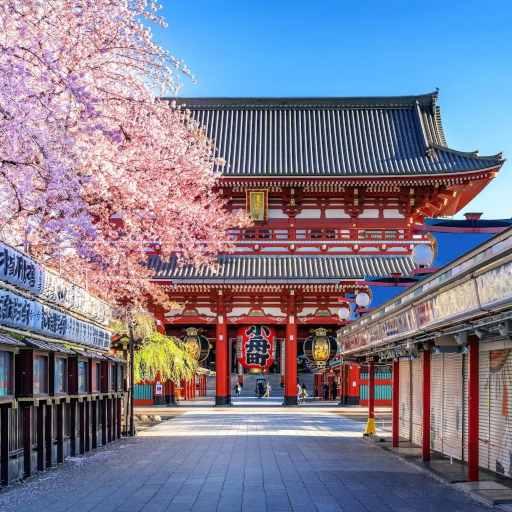} & \includegraphics[width=\gridimagewidth,valign=m]{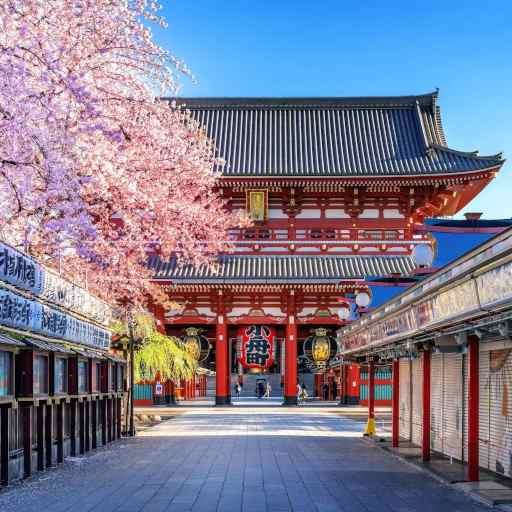} & \includegraphics[width=\gridimagewidth,valign=m]{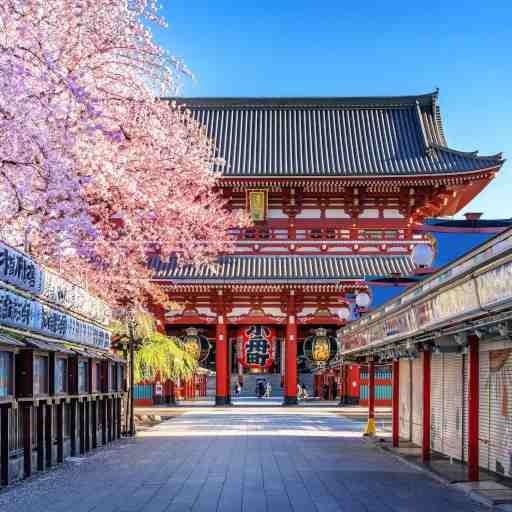} & \includegraphics[width=\gridimagewidth,valign=m]{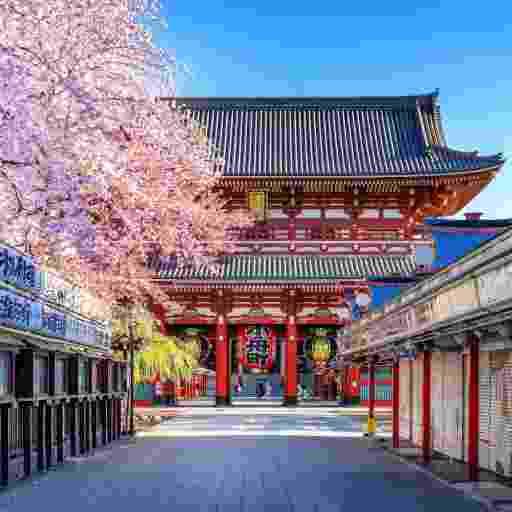} & \includegraphics[width=\gridimagewidth,valign=m]{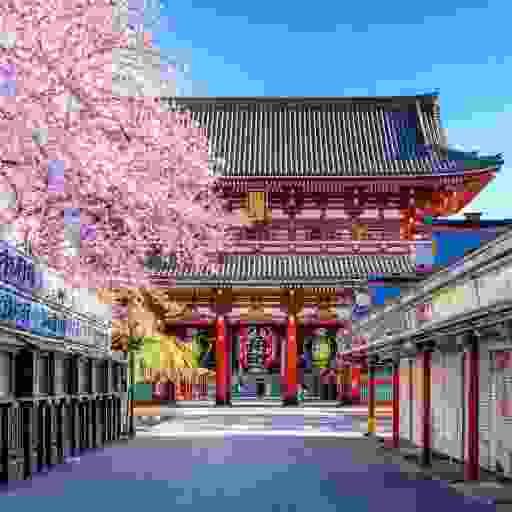} \\
\end{tabular}
}
\caption{Visualization of the distortion types belonging to the \textit{Compression} group for increasing intensity levels.}
\label{fig:compression}
\end{figure*}

\begin{figure*}
    \centering
    \setlength{\tabcolsep}{0.9pt}
    \resizebox{\textwidth}{!}{ 
\begin{tabular}{C{6em}ccccc}
         & Level 1 & Level 2 & Level 3 & Level 4 & Level 5 \\
 High sharpen & \includegraphics[width=\gridimagewidth,valign=m]{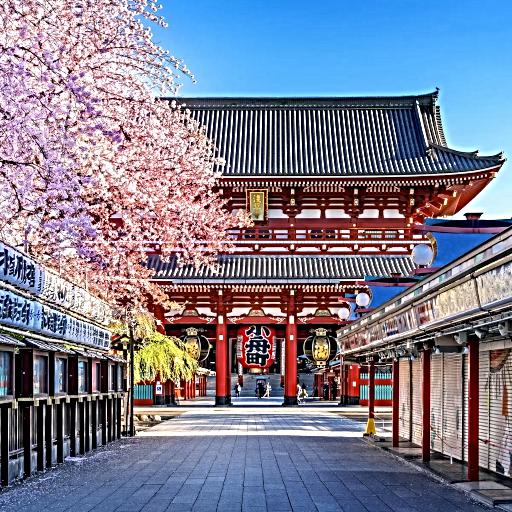} & \includegraphics[width=\gridimagewidth,valign=m]{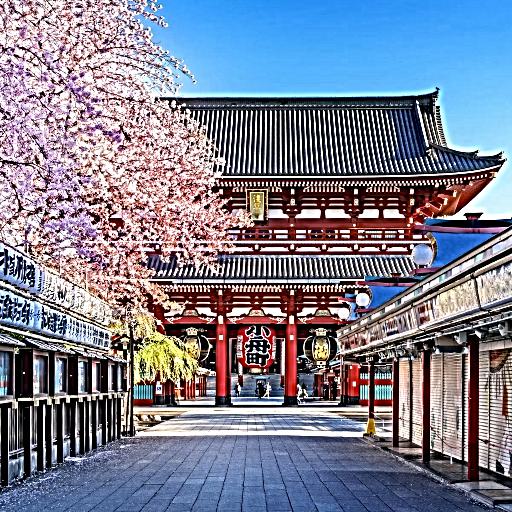} & \includegraphics[width=\gridimagewidth,valign=m]{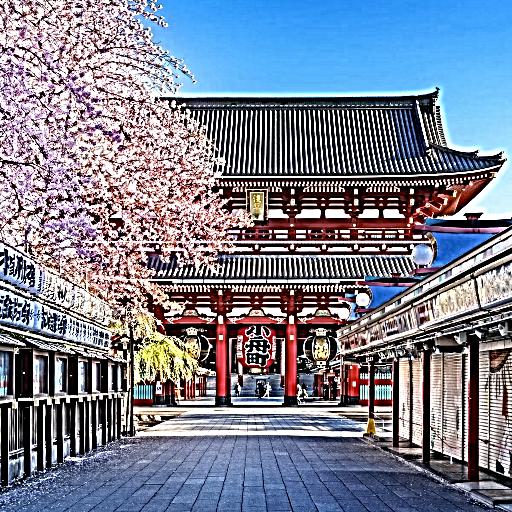} & \includegraphics[width=\gridimagewidth,valign=m]{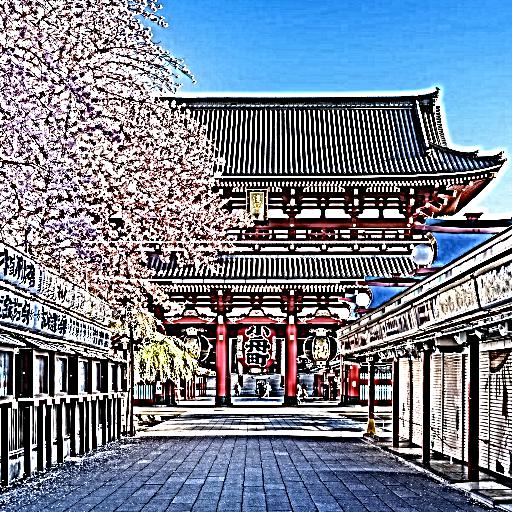} & \includegraphics[width=\gridimagewidth,valign=m]{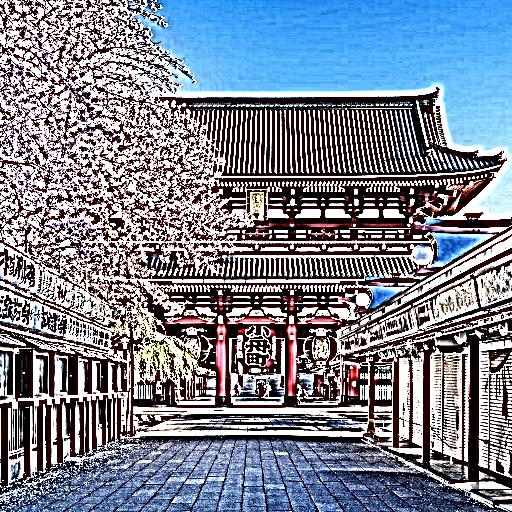} \\ [29pt]
 Nonlinear contrast change & \includegraphics[width=\gridimagewidth,valign=m]{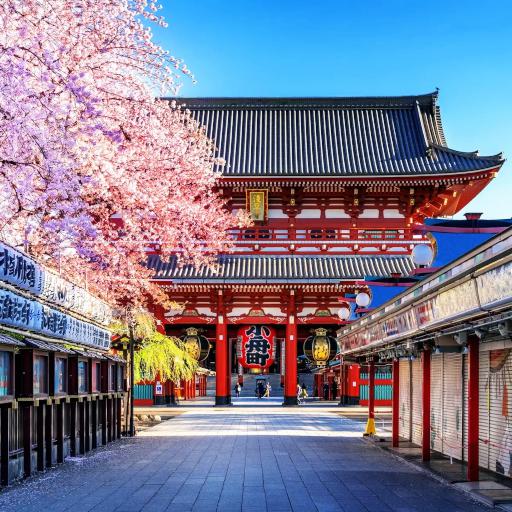} & \includegraphics[width=\gridimagewidth,valign=m]{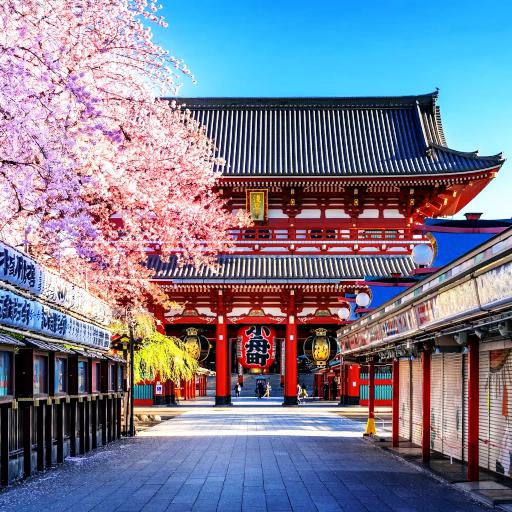} & \includegraphics[width=\gridimagewidth,valign=m]{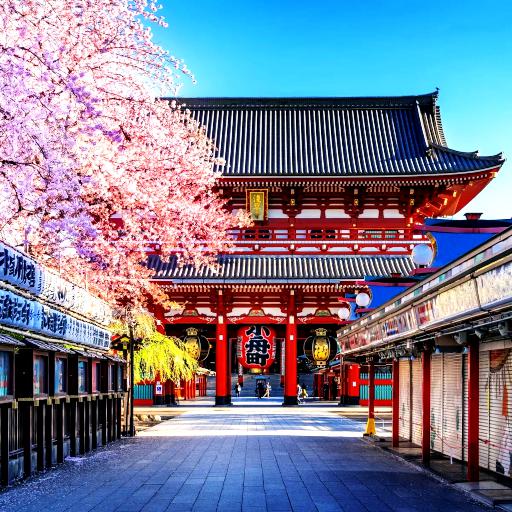} & \includegraphics[width=\gridimagewidth,valign=m]{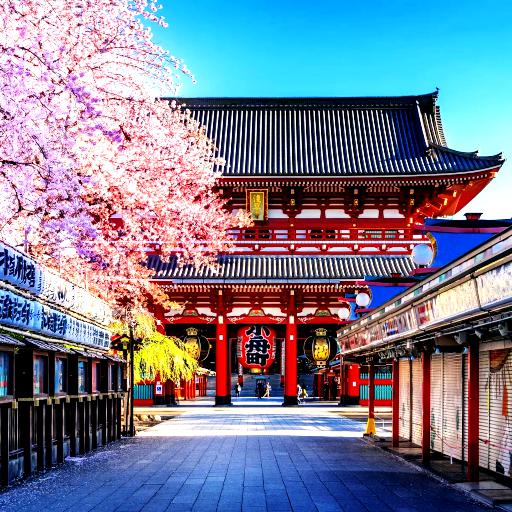} & \includegraphics[width=\gridimagewidth,valign=m]{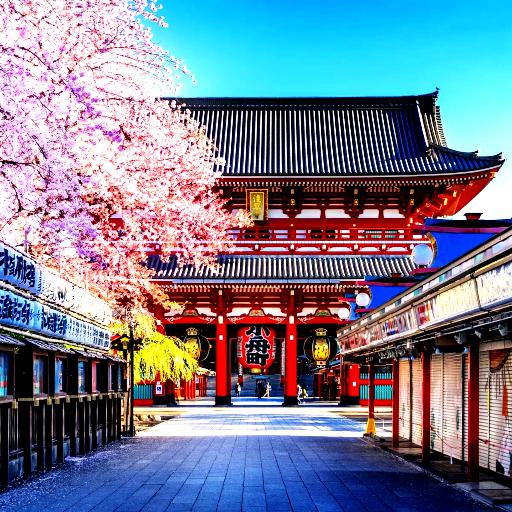} \\ [29pt]
 Linear contrast change & \includegraphics[width=\gridimagewidth,valign=m]{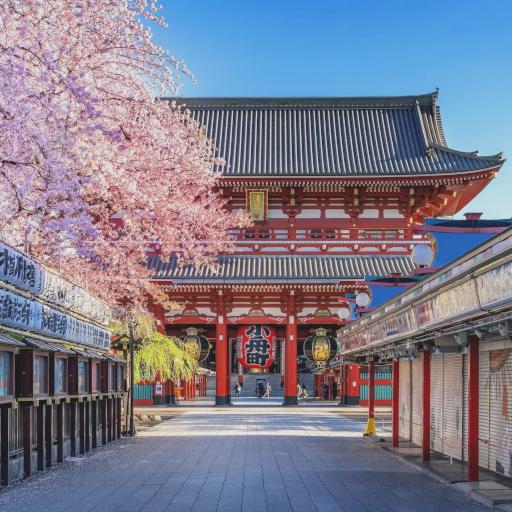} & \includegraphics[width=\gridimagewidth,valign=m]{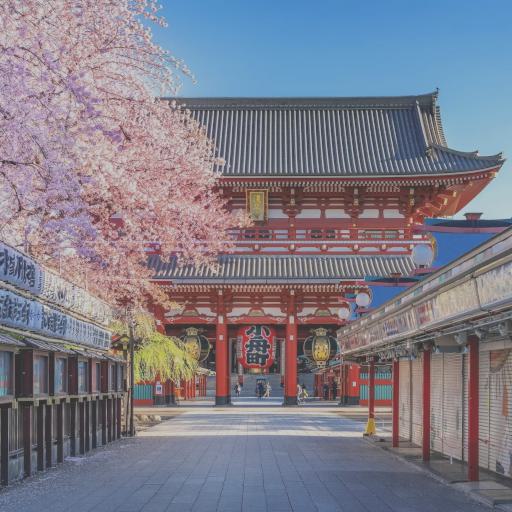} & \includegraphics[width=\gridimagewidth,valign=m]{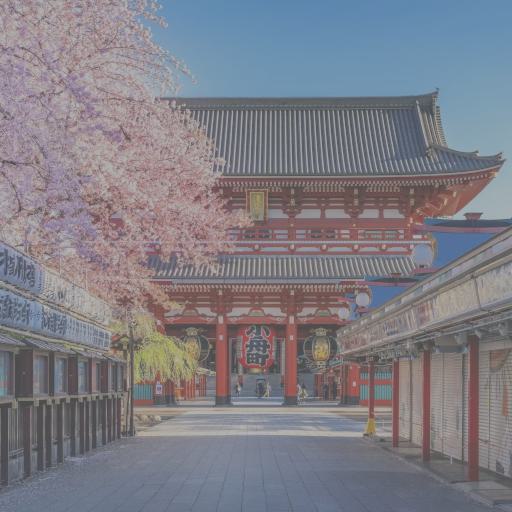} & \includegraphics[width=\gridimagewidth,valign=m]{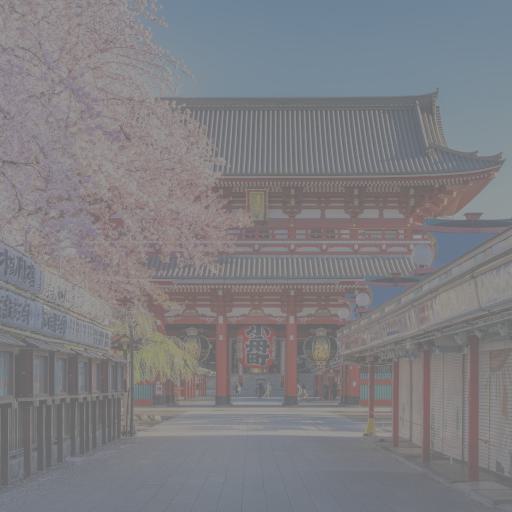} & \includegraphics[width=\gridimagewidth,valign=m]{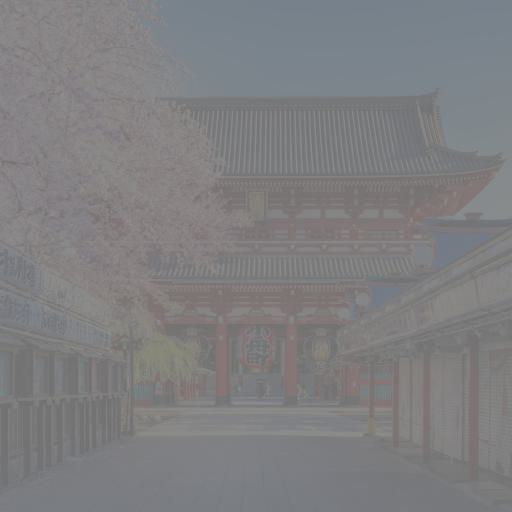} \\
\end{tabular}
}
\caption{Visualization of the distortion types belonging to the \textit{Sharpness \& contrast} group for increasing intensity levels.}
\label{fig:sharpness_contrast}
\end{figure*}

\tit{Synthetic distortions} \label{sec:distortions}
As detailed in Sec. \textcolor{iccvblue}{3.2}, during training we synthetically degrade pristine images with increasing intensity levels. Specifically, similar to \cite{agnolucci2024arniqa}, we consider 24 distinct distortion types divided into the 7 degradation groups defined by the \kadid \cite{kadid10k} dataset. Each degradation has 5 degrees of progressively higher intensity. We report an example for all the intensity levels of each type of degradation in \cref{fig:brightness_change,fig:blur,fig:spatial_distortions,fig:noise,fig:color_distortions,fig:compression,fig:sharpness_contrast}. Each distortion is described as follows:
\begin{enumerate}
    \item Brightness change:
        \begin{itemize}
            \item \textit{Brighten}: applies a sequence of color space transformations, curve adjustments, and blending operations to increase the brightness of the image;
            \item \textit{Darken}: similar to the brighten operation, but reduces the brightness instead of increasing it;
            \item \textit{Mean shift}: adjusts the average intensity of image pixels by adding a constant value to all pixel values. Then, it constrains the resulting values to stay within the original image range;
        \end{itemize}
    \item Blur:
        \begin{itemize}
            \item \textit{Gaussian blur}: applies a Gaussian kernel filter to each image pixel;
            \item \textit{Lens blur}: applies a circular kernel filter to each image pixel;
            \item \textit{Motion blur}: applies a linear motion blur kernel to each image pixel, simulating the effect of either a moving camera or a moving object in the scene. This results in the image appearing blurred in the direction of the motion;
        \end{itemize}
    \item Spatial distortions:
        \begin{itemize}
            \item \textit{Jitter}: randomly displaces image data by applying small offsets to warp each pixel;
            \item \textit{Non-eccentricity patch}: randomly selects patches from the image and places them in random neighboring positions;
            \item \textit{Pixelate}: employs a combination of downscaling and upscaling operations using nearest-neighbor interpolation;
            \item \textit{Quantization}: quantizes the image into $N$ uniform levels. The quantization thresholds are dynamically computed using Multi-Otsu’s method \cite{liao2001fast};
            \item \textit{Color block}: randomly superimposes uniformly colored square patches onto the image;
        \end{itemize}
    \item Noise:
        \begin{itemize}
            \item \textit{White noise}: adds Gaussian white noise to the image;
            \item \textit{White noise in color component}: transforms the image to the YCbCr color space and then adds Gaussian white noise to each channel;
            \item \textit{Impulse noise}: adds salt and pepper noise to the image;
            \item \textit{Multiplicative noise}: adds speckle noise to the image;
        \end{itemize}
    \item Color distortions:
        \begin{itemize}
            \item \textit{Color diffusion}: transforms the image to the LAB color space and then applies Gaussian blur to each channel;
            \item \textit{Color shift}: randomly shifts the green channel and then blends it into the original image, masking it with the normalized gradient magnitude of the original image;
            \item \textit{Color saturation 1}: transforms the image to the HSV color space and then scales the saturation channel by a factor;
            \item \textit{Color saturation 2}: transforms the image to the LAB color space and then scales each color channel by a factor;
        \end{itemize}
    \item Compression:
        \begin{itemize}
            \item \textit{JPEG2000}: applies the standard JPEG2000 compression to the image;
            \item \textit{JPEG}: applies the standard JPEG compression to the image;
        \end{itemize}
    \item Sharpness \& contrast:
        \begin{itemize}
            \item \textit{High sharpen}: applies unsharp masking to sharpen the image in the LAB color space;
            \item \textit{Nonlinear contrast change}: applies a nonlinear tone mapping operation to adjust the contrast of the image;
            \item \textit{Linear contrast change}: applies a linear tone mapping operation to adjust the contrast of the image;
        \end{itemize}
\end{enumerate}

\section{Limitations}
The proposed approach leverages the intrinsic quality ranking of progressively degraded images as supervision to train a model in a self-supervised way. This involves defining a method to synthetically generate an inherent ranking from unlabeled images, which, in our work, is represented by the application of synthetic degradations. While we find this strategy beneficial for assessing technical image quality, it is not directly applicable to abstract perception (\eg happiness or naturalness) \cite{wang2023exploring} or aesthetic quality assessment. Indeed, generating an inherent image ranking for such abstract or aesthetic quantities without employing annotations requires more sophisticated strategies compared to the straightforward yet effective application of low-level degradations. Future work could focus on developing such strategies, for example by exploiting the capabilities of text-to-image generation models \cite{rombach2022high, zhang2023adding} to directly synthesize intrinsically ranked images.

\end{document}